\pdfoutput=1
\documentclass[11pt]{article}

\usepackage[final]{acl}

\usepackage{times}
\usepackage{adjustbox}
\usepackage{makecell}
\usepackage{multirow}
\usepackage{tabularx}
\usepackage{amsmath}
\usepackage{latexsym}
\usepackage{arydshln}
\usepackage{amsthm}
\usepackage{multirow}
\usepackage{subcaption}
\usepackage{soul}
\usepackage{arydshln}

\definecolor{delectricblue}{RGB}{93, 117, 131}
\definecolor{limegreen}{RGB}{50, 205, 50}
\colorlet{lightdelectricblue}{delectricblue!30}

\definecolor{lightblue}{rgb}{.90,.95,1}
\sethlcolor{lightblue}
\usepackage{xcolor}
\usepackage{makecell} % Add this in your preamble
\usepackage{booktabs}
\usepackage{tcolorbox}

\usepackage{arydshln}
\usepackage{color,soul}

\usepackage{graphicx}
\usepackage[T1]{fontenc}
\usepackage[utf8]{inputenc}
\usepackage{multirow}
\usepackage{microtype}

\usepackage{inconsolata}

\usepackage{graphicx}
\newtcolorbox{promptbox}{
  colback=white!20,
  colframe=gray!50!black,
  boxsep=10pt,
  left=1pt, right=1pt, top=1pt, bottom=1pt, boxrule=0.5pt
}
\definecolor{beaublue}{rgb}{0.74, 0.83, 0.9}

\definecolor{lighterpastelgreen}{rgb}{0.9, 1.0, 0.9}

\definecolor{pastelorange}{rgb}{1.0, 0.95, 0.9}

\definecolor{midorange}{rgb}{1.0, 0.75, 0.4}

\definecolor{lightgray}{RGB}{220,220,220} % Lighter gray than default

\DeclareRobustCommand{\hlgray}[1]{{\sethlcolor{lightgray}\hl{#1}}}

\newtcolorbox{mybox}[1]{colback=white!20,colframe=gray!50!black,fonttitle=\bfseries,title=#1}
\title{\emph{Reviewing the Reviewer:} \\
LLM-Assisted Reviewer Feedback Generation for Guideline Compliance}
\author{ Sukannya Purkayastha$^1$\thanks{Currently at CAIDAS, University of W\"urzburg}, Qile Wan$^2$, Anne Lauscher$^3$, Lizhen Qu$^2$, Iryna Gurevych$^1$ \\
         $^1$ Ubiquitous Knowledge Processing Lab (UKP Lab),\\ Department of Computer Science, Technical University of Darmstadt and \\ National Research Center for Applied Cybersecurity ATHENE, Germany \\
         \url{www.ukp.tu-darmstadt.de} \\
         $^2$ Department of Data Science \& AI, Monash University, Australia \\
         $^3$ Trustworthy AI Lab, University of Hamburg  
         }%  

\begin{document}
\maketitle
\begin{abstract}
Peer review is central to scientific quality, yet reliance on simple heuristics, namely \textit{lazy thinking} and non-\textit{specific} critiques, has threatened review quality. Prior work frames \textit{lazy thinking} detection as single-label classification and stops at detection, yet review segments often exhibit multiple co-occurring issues, and reviewers benefit more from actionable, guideline-aware feedback than from labels alone. We further show that off-the-shelf LLMs prompted for feedback frequently rewrite the entire review or address the authors rather than the reviewer, motivating an inference-time approach. We introduce an LLM-driven framework that decomposes reviews into argumentative segments, identifies issues violating ACL Rolling Review (ARR) guidelines, and generates targeted feedback using issue-specific templates refined by a novel iterative, reranking-based generation algorithm. In a controlled rewriting study, our feedback reduces guideline violations by up to 92.4\%. We also release \textsc{LazyReviewPlus}, the \textit{first} multi-label dataset of 1,309 sentences annotated for detecting \textit{lazy thinking} and lack of \textit{specificity}.\footnote{Code and Data: \url{https://github.com/UKPLab/emnlp2026-reviewfeedbackagent}}

\end{abstract}

\section{Introduction}
Peer review is the cornerstone of scientific quality control, ensuring rigorous evaluation of research~\cite{ware2015stm}. However, the system faces growing strain, particularly in AI, where paper submissions have surged from 1,678 at NeurIPS 2014 to 17,491 in 2024, a $10.4\times$ increase~\cite{wei2025ai}. This growth, driven by large language models (LLMs)~\cite{liang2024mapping} and the \textit{publish-or-perish} culture~\cite{https://doi.org/10.1002/asi.22636}
, has far outpaced the supply of qualified reviewers, even with mandatory reviewing policies.\footnote{\url{https://aclrollingreview.org/reviewing-workload-adjustment/}}
 The result is an unsustainable global workload of over 130 million hours annually that threatens review quality~\cite{aczel2021billion}.
%Peer review serves as the primary mechanism for maintaining scientific quality control, relying on expert evaluation of manuscripts to refine and validate scholarly contributions~\cite{ware2015stm}. Despite its central role in lending credibility to research, the system is increasingly strained by systemic pressures. In the domain of AI, particularly, research output has accelerated at a dramatic pace. For instance, submissions to NeurIPS, the top-most AI conference, grew from 1,678 in 2014 to 17,491 in 2024, showing a $10.4\times$ increase~\cite{wei2025ai}. As per~\citet{kim2025position}, this surge is mainly driven by large language models (LLMs), which have lowered barriers to paper writing~\cite{liang2024mapping}, and by academia’s persistent \textit{publish-or-perish} culture~\cite{https://doi.org/10.1002/asi.22636}. Yet the supply of qualified reviewers has not kept pace, even with measures such as mandatory reviewing load for authors at major AI/NLP conferences.\footnote{\url{https://aclrollingreview.org/reviewing-workload-adjustment/}} The result is an unsustainable workload—estimated at 15 million hours annually worldwide—that undermines review quality~\cite{aczel2021billion}.
\begin{figure}[!t]
    \centering
    \includegraphics[width=0.7\linewidth]{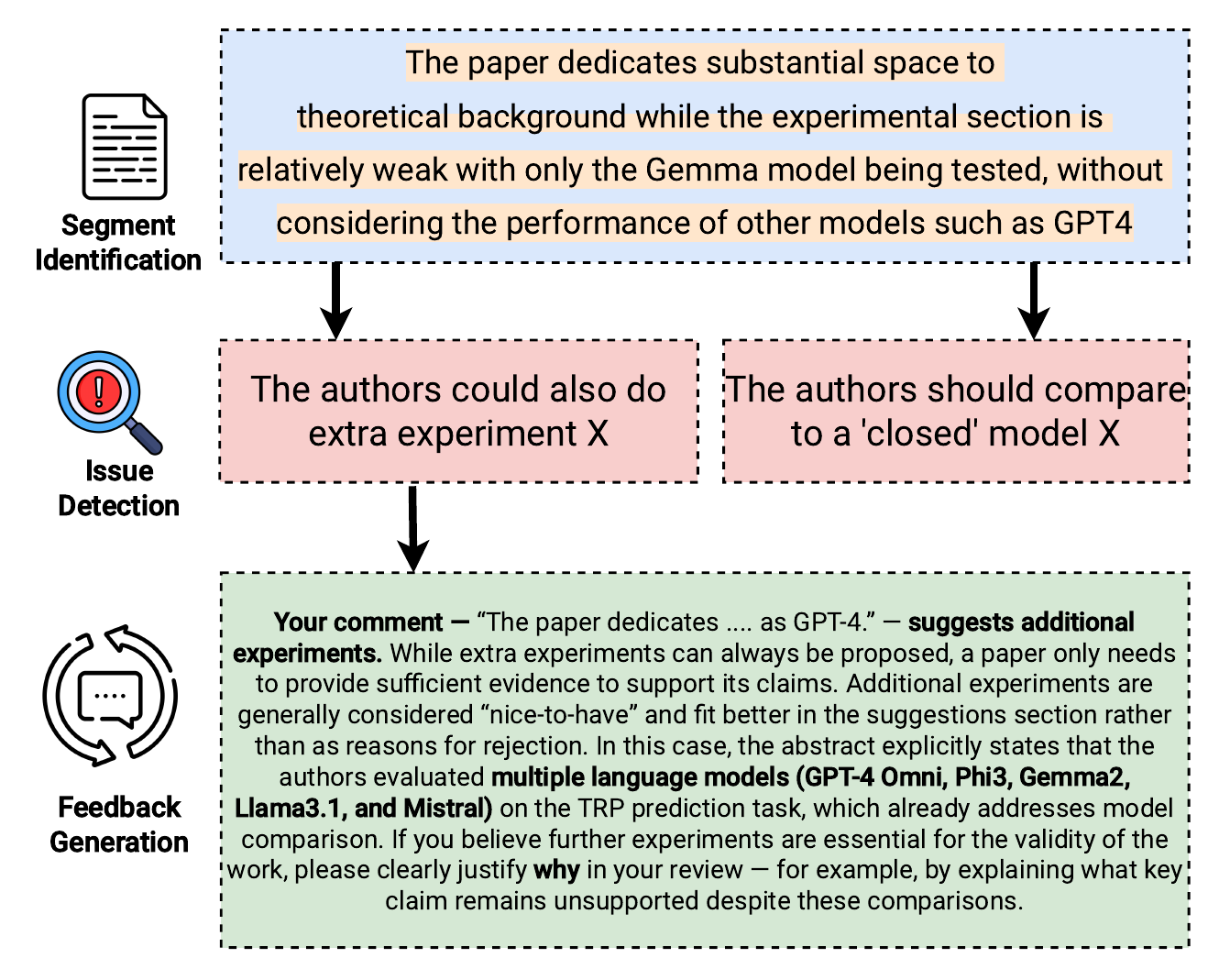}
    \caption{Overall pipeline of our method. We first \textbf{identify segments} within a review, then \textbf{detect issues}, and finally \textbf{generate feedback} to improve each segment.}
    \label{fig:lazy_thinking}
\end{figure}

In NLP research, declining review quality is often attributed to two failure modes flagged by the ACL Rolling Review (ARR) guidelines: \textit{lazy thinking}, where oversimplified heuristics lead reviewers to dismiss submissions~\cite{Rogers_Augenstein_2021}, and \textit{specificity issues}, where vague phrasing such as \textit{``X is not clear''} replaces precise feedback such as \textit{``X is not clear because of Y''}~\cite{sadallah2025goodbadconstructiveautomatically}.\footnote{\url{https://aclrollingreview.org/reviewerguidelines}} \textit{Lazy thinking} alone accounts for 24.3\% of author-reported issues at ACL 2023~\cite{rogers-etal-2023-report}. For example, the \textit{review segment} (an argumentative unit consisting of one or more sentences) in Fig~\ref{fig:lazy_thinking} criticizes a paper for using only GPT-4 and requests additional ``nice-to-have'' experiments, an example of the ARR guideline issue ``authors could also do extra experiment X''. We focus on these two issue families because the ARR guidelines define them explicitly and at sufficient granularity to support annotation, detection, and targeted feedback. Our framework is not tied to ARR, as it applies to any venue whose reviewer guidelines specify identifiable issue categories.

 Recent efforts to address these issues fall short in two ways. \citet{purkayastha-etal-2025-lazyreview} show that lazy thinking annotations can improve review quality but stop at detection, providing no \textit{verbal feedback} to guide reviewer revision, which is vital for actionable improvement~\cite{jansen2025constructive}. \citet{thakkar2025llmfeedbackenhancereview} report improved specificity and professionalism in a large-scale ICLR 2024 trial, but their feedback is not aligned with any reviewer guidelines and depends on a closed-source model. A natural alternative is to prompt open-weight LLMs to generate feedback directly. In our pilot study, however, 41\% of such outputs extended reviews (`review extension') and 23\% misaddressed authors instead of the reviewer (`role separation') (cf.~\S\ref{sec:feedback_gen}). These failures underscore the need for inference-time methods that produce specific, targeted, and guideline-aware feedback.

% While addressing \textit{lazy thinking} and \textit{specificity} issues is essential for improving review quality, progress has been limited by the lack of suitable datasets for training and evaluation. The most closely related work~\citep{purkayastha-etal-2025-lazyreview} annotates a single lazy thinking issue per review segment \textit{without addressing specificity issues}, focusing primarily on the ``weakness'' sections of reviews. However, such patterns can also occur in other parts, such as ``comments or suggestions'', and may involve multiple issue types simultaneously (cf. Figure~\ref{fig:lazy_thinking}). This motivates building a new dataset enabling \textit{multi-label} detection of \textit{both} types of issues across diverse review sections.
 
%Addressing \textit{lazy thinking} and \textit{specificity} is crucial for review quality, but the progress is limited by dataset availability. Prior work~\citep{purkayastha-etal-2025-lazyreview} annotates only a single lazy thinking issue per segment and focuses on ``weakness'' sections, ignoring specificity and other review sections. However, multiple issue types can co-occur across sections (cf. Figure~\ref{fig:lazy_thinking}), motivating a new dataset for \textit{multi-label} detection of both issues.
Progress on this problem is further limited by dataset availability. Existing resources annotate only a \textit{single} lazy thinking issue per segment and cover only the ``weaknesses'' section~\citep{purkayastha-etal-2025-lazyreview}, even though multiple issue types can co-occur across sections (cf.~Figure~\ref{fig:lazy_thinking}), and \textit{specificity} is absent entirely.

%To address these gaps, we present a framework for issue-specific feedback generation. Our framework uses open-weight LLMs to segment reviews, detects issues using a \textbf{neurosymbolic precision-oriented} module that transforms text into structured features through a set of yes/no QA subtasks, classified by an Extra-Trees classifier~\cite{geurts2006extremely}. It then generates actionable feedback for identified issues through \textit{LLM-driven templates} refined with a novel \textit{genetic algorithm}~\cite{lee2025evolvingdeeperllmthinking} to balance specificity and relevance to the guidelines. Additionally, we introduce \textsc{LazyReviewPlus}, a dataset of \textbf{1,309} expert-annotated review sentences from ARR 2022 and EMNLP 2024~\cite{dycke-etal-2023-nlpeer}, with multi-label annotations for \textit{lazy thinking} and \textit{specificity} issues. 

To address these gaps, we present a framework that generates actionable, guideline-aligned feedback for reviewers using open-weight LLMs. Rather than relying on generic suggestions, the framework identifies specific weaknesses in a review and produces targeted recommendations grounded in ARR reviewing guidelines. To achieve this, it segments reviews into argumentative units, detects guideline violations, and generates \textit{issue-specific} feedback using templates optimized through a customised iterative reranking approach inspired by the genetic algorithm~\cite{lee2025evolvingdeeperllmthinking} with verifiable composite rewards tailored to reviewer feedback. Alongside the framework, we release \textsc{LazyReviewPlus}, a dataset of \textbf{1,309} expert-annotated sentences from ARR 2022 and EMNLP 2024~\cite{dycke-etal-2023-nlpeer} with multi-label annotations for \textit{lazy thinking} and \textit{specificity}.

In summary, our contributions are three-fold: 
   (1) an inference-time \textit{feedback generation} framework combining issue-specific templates and a dedicated \textit{iterative, reranking-based generation algorithm} with verifiable rewards, improving constructiveness by about 26\% and relevance by about 23\% relative to the strongest baseline (cf. Table~\ref{tab:feedback_models_metrics}) and mitigating \textit{review extension} and \textit{role separation} failures of zero-shot LLM feedback;
   (2) a controlled review-rewriting study with senior PhD reviewers, showing that feedback from our framework reduces guideline violations by up to \textbf{92.4}\% relative to original reviews and outperforms rewrites guided by issue labels alone;
   (3) \textsc{LazyReviewPlus}, the \textit{first} multi-label review dataset annotated for the presence of \textit{lazy thinking} and lack of \textit{specificity}, spanning \textit{weaknesses} and \textit{comments/suggestions} sections across two venues.
%\textbf{Contributions.} 
%Our contributions are three-fold:
%\begin{itemize}
%    \item an LLM-driven \textbf{feedback generation} module that integrates issue-specific templates and a genetic algorithm, improving constructiveness and relevance by about 20\% over the baselines;
%    \item controlled review-rewriting studies showing that feedback from our model reduces review issues by up to \textbf{92.4}\% relative to original reviews;
%    \item \textsc{LazyReviewPlus}, the \textit{first} multi-label review dataset annotated with \textit{lazy thinking} and \textit{specificity} for issue identification and feedback generation.
%\end{itemize}
%Our contributions are: (1) a neurosymbolic \textbf{issue detection} approach that doubles the F0.5 performance of the strongest LLM-only model; (2) an LLM-driven \textbf{feedback generation} module that integrates issue-specific templates and a genetic algorithm, improving constructiveness and relevance by about 20\% over the baselines; and (3) controlled review-rewriting studies showing that feedback from our model reduces review issues by up to \textbf{92.4}\% relative to original reviews; (4) \textsc{LazyReviewPlus}, the \textit{first} multi-label review dataset annotated with \textit{lazy thinking} and \textit{specificity} for issue identification and feedback generation.

\section{Construction of \textsc{LazyReviewPlus}} \label{sec:background}

Building on top of the \textsc{LazyReview} dataset~\cite{purkayastha-etal-2025-lazyreview}, which annotates \textit{500} segments with \textit{16} \textit{lazy thinking} classes in single-label form on the \textit{weaknesses} section only, we extend annotations to a single-segment, multi-label form covering both \textit{lazy thinking} and \textit{specificity} (\textit{18} and \textit{7} ARR-defined issues, respectively), spanning both \textit{weaknesses} and \textit{comments/suggestions} sections.\footnote{Of the 18 \textit{lazy thinking} issues, 16 are from \textsc{LazyReview}, while 2 were recently added to the ARR guidelines. Full issue lists in Tables~\ref{tab:heuristics_clean} and~\ref{tab:issues_rewrite}; examples in Table~\ref{tab:issues}.}

\begin{figure}
    \centering
    % Row 1
    \begin{subfigure}[b]{0.4\textwidth}
        \centering
        \includegraphics[width=0.8\textwidth]{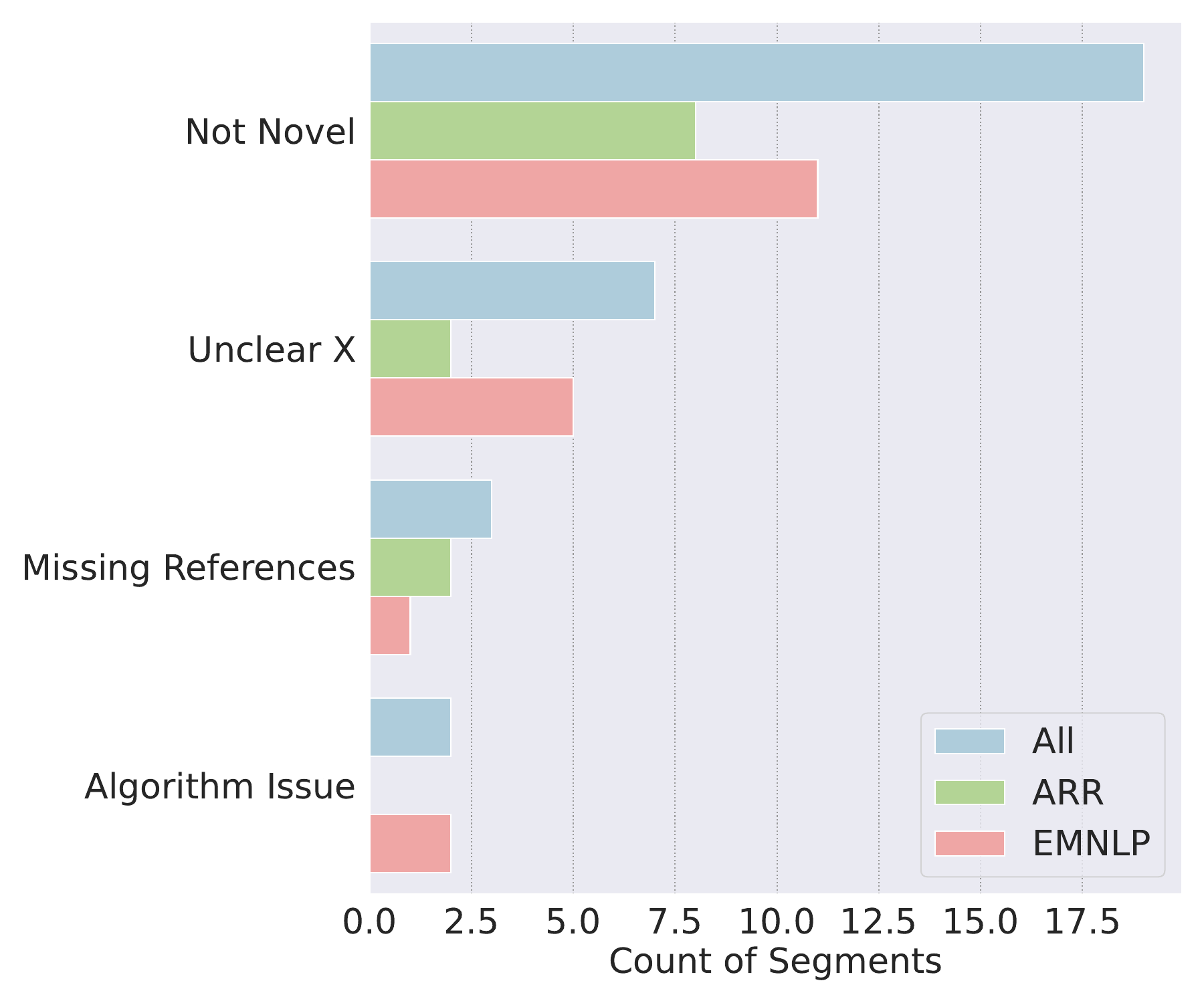}
        \caption{\textit{Specificity} Issues}
        \label{fig:specificity_iss}
    \end{subfigure}
    \hfill
    \begin{subfigure}[b]{0.4\textwidth}
        \centering
        \includegraphics[width=0.8\textwidth]{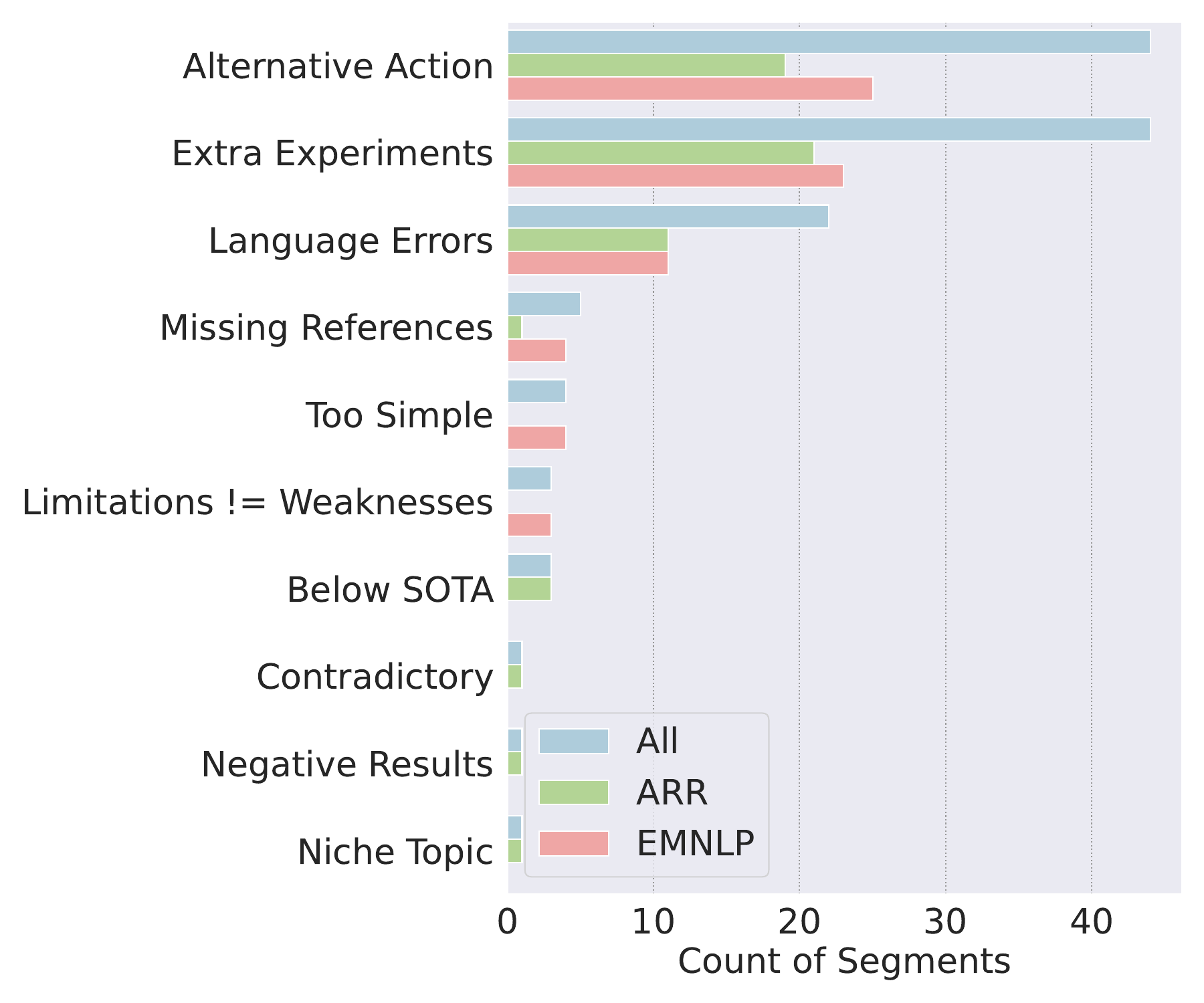}
        \caption{\textit{Lazy Thinking} Issues}
        \label{fig:issue_distribution}
    \end{subfigure}

    \caption{\textit{Specificity} issues and overall distribution of \textit{lazy thinking} issues in the \textsc{LazyReviewPlus} dataset.}
\end{figure}

\subsection{Data Procurement and Annotation}
\noindent \textbf{Underlying Data.} To construct \textsc{LazyReviewPlus}, we sample 100 reviews (50 each) from ARR 2022 and EMNLP 2024 within \textsc{NLPeer}~\cite{dycke-etal-2023-nlpeer}, a licensed dataset of ACL and journal reviews~\cite{dycke-etal-2022-yes}. We sample from two different timestamps to avoid temporal biases in the dataset creation. Our 100 sampled reviews span 50 papers from ARR 2022 and 38 papers from EMNLP 2024. Reviews were sampled uniformly at random within each venue, subject to the exclusion criteria applied prior to sampling: the review must contain both a non-empty summary of weaknesses and a non-empty comments/suggestions section. Reviews are segmented at the sentence level using SpaCy, covering both \textit{weaknesses} and \textit{comments/suggestions}, unlike \textsc{LazyReview} which considers only \textit{weaknesses}.\footnote{\url{https://spacy.io/api/sentencizer}} This produces 616 and 693 sentences, totaling \textbf{1,309} sentences.

\noindent \textbf{Task and Guidelines.} We identify review segments with \textit{lazy thinking} and \textit{specificity}. Annotators view the full review along with sentences from the \textit{summary of weaknesses} and \textit{comments, suggestions, and typos} sections. Sentences are labeled using the \textbf{B-I-O} scheme~\cite{Ramshaw1999}, with each span assigned an issue type without access to the paper, as paper-level verification would substantially increase annotation cost. Building on \textsc{LazyReview}~\cite{purkayastha-etal-2025-lazyreview}, we add two categories—\textit{None} (not an issue) and \textit{Not Enough Info} (cases requiring paper-level verification, e.g., novelty or SOTA claims)—resulting in \textbf{27} issue types (25 from ARR guidelines plus 2 additional classes).\footnote{Final annotation instructions in Appendix~\S\ref{sec:annotation}.}\looseness=-1

\noindent \textbf{Quality Control.}
Four Ph.D. students with extensive reviewing experience served as annotators, trained on the finalized guidelines. Disagreements were resolved by a senior student. To ensure reliability, 8\% of the dataset (8 reviews; 102 sentences inline with doubly annotating 99 peer review sentences in \citet{purkayastha-etal-2023-exploring}) was multiply-annotated to compute agreement, while the remaining 92\% (92 reviews; 1,207 sentences) was evenly split among annotators. Annotating each review took \(\sim\)35 minutes. This amounted to 124 annotation passes in total: 92 passes for the singly annotated reviews, plus 32 passes (8 reviews \(\times\) 4 annotators) for the reviews annotated by all four annotators. These passes took \(\sim\)72 hours combined. Adding 3 hours for guideline development and adjudication brought the total to \textbf{\(\sim\)75 hours} of work, or \(\sim\)19 hours per annotator. The task requires prior *ACL reviewing experience to make fine-grained judgments, imposes a substantial time commitment on the senior researchers and NLP PhD students who alone qualify. Pool sizes here accordingly match prior expert studies of peer review~\cite{kuznetsov-etal-2022-revise, purkayastha-etal-2023-exploring, bharti2024politepeer}. Inter-Annotator Agreement (Krippendorff's alpha~\cite{krippendorff2004reliability}) was \textbf{0.78} for \textbf{segment identification} (pairwise 0.74–0.82) and \textbf{0.53} for \textbf{issue detection} (pairwise 0.51–0.55) consistent with prior peer-review studies~(e.g., 0.479 in \citet{kennard-etal-2022-disapere} and 0.52 in \citet{purkayastha-etal-2025-lazyreview}) and reflecting the difficulty of identifying \textit{lazy thinking} and \textit{specificity} issues.
\looseness=-1

\subsection{Dataset Analysis}
Our dataset, \textsc{LazyReviewPlus}, contains \textbf{1,309} sentences for \textit{lazy thinking} and \textit{specificity} issues (Table~\ref{tab:new_dataset}) corresponding to \textbf{518} segments, comparable to prior human-annotated datasets~(e.g., 500 review segments in \citet{purkayastha-etal-2025-lazyreview}). Segment lengths range from 1--20 sentences (mostly single sentences; Fig.~\ref{fig:seg_length_distribution}) with 1--2 labels per segment (Fig.~\ref{fig:label_to_segment}). For \textit{lazy thinking}, `Alternative Action' (corresponding to `authors should do X') and `Extra Experiments' dominate (Fig.~\ref{fig:issue_distribution}), consistent with \citet{purkayastha-etal-2025-lazyreview}, while \textit{specificity} issues mostly manifest as clarity problems such as `Not Novel' and `Unclear X' (Fig.~\ref{fig:specificity_iss}). The `Not Novel' concern likely reflects the fast pace of advancements in ML, which quickly render state-of-the-art methods obsolete~\cite{9891914}, while `Unclear X' reflects common non-constructive feedback patterns~\cite{sadallah2025goodbadconstructiveautomatically}. These issues of \textit{lazy thinking} and lack of \textit{specificity} persist across both the ARR 2022 and EMNLP 2024 sections (cf. Figures~\ref{fig:specificity_iss} and \ref{fig:issue_distribution}), suggesting that reliance on such heuristics remains a recurring trend.

\section{Methodology}
Our pipeline comprises three stages. First, each review is split into sentences and grouped into argumentative segments via \textbf{B-I-O} tagging as described in \S\ref{sec:segment_id}.\footnote{Sentence split is performed using Spacy} Second, each segment undergoes multi-label \textbf{issue detection}, permitting co-occurring labels (e.g., \textit{Extra Experiments} and \textit{Unclear X}) (\S\ref{sec:issue_detect}). Third, each detected label triggers an issue-specific template in the \textbf{feedback generation} module, enriched by a planner and refined via a genetic algorithm, producing multiple targeted feedback pieces for segments with multiple labels (\S\ref{sec:feedback_gen}). The full pipeline is illustrated in Figure~\ref{fig:lazy_thinking}. We briefly describe the segment identification and issue detection approaches, and primarily focus on our main contribution of \textbf{targeted feedback generation}.
%Since we adopt a standard zero-shot LLM tagging approach for segment identification and  , we focus the remainder of this section on our novel contributions on \textbf{feedback generation}.

%This section formalizes LLM-driven feedback generation and its subtasks: \textbf{issue detection}, which assigns lazy thinking and specificity labels to segments (Sec.~\S\ref{sec:issue_detect}), and \textbf{feedback generation}, which produces targeted feedback for problematic segments (Sec.~\S\ref{sec:feedback_gen}).

%This section formalizes the overall task of LLM-driven \textbf{feedback generation} and outlines its decomposition into multiple subtasks. First, \hlcyan{segment identification} divides reviews into argumentative units (Sec.~\S\ref{sec:segment_id}). Next, \textbf{issue detection} maps each segment to potential \textit{lazy thinking} and \textit{specificity} issues (Sec.~\S\ref{sec:issue_detect}). Finally, the \textbf{feedback generation} module produces feedback for segments exhibiting these issues (Sec.~\S\ref{sec:feedback_gen}).

%\iffalse
\subsection{Segment Identification} \label{sec:segment_id}

\noindent \textbf{Task definition.} For a review \( R = (sent_1, \dots, sent_n) \), each sentence \( sent_i \) is tagged \( y_i \in \{B, I, O\} \), where \( B \) marks a segment beginning, \( I \) an inside sentence, and \( O \) one outside any segment. The tagging function \( f(sent_i) = y_i \) yields \( Y = (y_1, \dots, y_n) = f(R) \).

%For a review $R = (s_1, \dots, s_n)$, each sentence $s_i$ is assigned a tag $y_i \in \{B, I, O\}$, where $B$ marks the beginning of a segment, $I$ a sentence inside a segment, and $O$ a sentence outside any segment. Let $f(s_i) = y_i$ be the tagging function; then the full review is tagged as $Y = (y_1, \dots, y_n) = f(R)$, with $y_i = B$ if $s_i$ starts a segment, $y_i = I$ if it continues a segment, and $y_i = O$ otherwise.\vspace{-2mm}

%\noindent \textbf{Approach.} Following prior work~\cite{lan-etal-2024-multi, pichler-etal-2025-evaluating}, we prompt the LLMs with each review sentence and the full review, instructing them to tag every sentence with a `B/I/O' label, detailed in Appendix \ref{sec:segment_id_full}. The tag definitions match those provided to annotators during dataset construction.
\noindent \textbf{Approach.} A naive approach of prompting an LLM to directly segment a review tends to produce oversized segments that span multiple argumentative units, motivating a finer-grained formulation. Following prior work on LLM-based sequence labeling~\cite{lan-etal-2024-multi, pichler-etal-2025-evaluating}, we instead prompt the LLM with each review sentence together with the full review as context, and instruct it to assign a `B/I/O' tag per sentence, detailed in Appendix~\ref{sec:segment_id_full}. The tag definitions match those provided to annotators during dataset construction.\looseness=-1
%\fi
%Let $f_\theta$ denote the LLM mapping a sentence to a tag in $\{B, I, O\}$. Following prior work~\cite{lan-etal-2024-multi, pichler-etal-2025-evaluating}, for each sentence $s_i$ in review $R$, we predict 
%\(
%\hat{y}_i = f_\theta(s_i, R)
%\)
%to indicate beginning ($B$), inside ($I$), or outside ($O$) a segment. %This prediction relies solely on $s_i$ within the full review, treating each sentence independently.

 %This prediction relies solely on $s_i$ contextualized within the full review, without conditioning on prior predictions.

%Let $f_\theta$ denote the LLM mapping a sentence (optionally with prior information) to a tag in ${B, I, O}$. Following prior work~\cite{lan-etal-2024-multi, pichler-etal-2025-evaluating}, we perform segment detection using two approaches: \textbf{1. Standalone}: For each sentence $s_i$ in review $R$, predict $\hat{y}_i = f_\theta(s_i, R)$ to indicate beginning ($B$), inside ($I$), or outside ($O$) a segment. This relies solely on $s_i$ contextualized within the full review. \textbf{2. Sequential}: To capture inter-sentence dependencies, the LLM conditions on prior predictions $\hat{Y}_{<i} = (\hat{y}_1, \dots, \hat{y}_{i-1})$, predicting $\hat{y}_i = f_\theta(s_i, R, \hat{Y}_{<i})$. This provides temporary memory of past tagging decisions.

%\paragraph{Approach.}

\subsection{Issue Detection} \label{sec:issue_detect}
%LLMs with zero-shot classification showcase low precision predictions for hard classes, however to provide effective feedback to the reviewers we require correct classification of the issue types. With precision-oriented issue identification we increase the overall trustworthiness and faithfulness.
 %Unlike sentence-level tagging for segmentation, this is a multi-label classification problem at the segment level.

\noindent \textbf{Task definition.}
%We are working on a multi-label classification task where the objective is to determine whether a given review segment demonstrates signs of lazy thinking. If so, the segment is further categorized into one or more specific classes of lazy thinking.
%We placed strong emphasis on precision-oriented classification. The reasoning is that false positives—incorrectly labeling a well-argued, high-quality review as lazy thinking—would be more harmful than false negatives. Such misclassification could undermine user trust and penalize thoughtful contributions. We wish to highlight which dimension of lazy thinking it falls under and suggest tailored improvements in the feedback.
Given a predefined label set $\mathcal{L} = \{\ell_1, \dots, \ell_m\}$, the goal is to map each review segment $s_i$, consisting of multiple sentences $s_i = (sent_1, \dots, sent_k)$, to a subset of labels, where a review $R$ consists of multiple segments, i.e., $R = \{s_1, s_2, \dots, s_n\}$. Formally, an issue detection function $g(s_i) \subseteq \mathcal{L}$ produces label sets for all segments in a review, yielding $G(R) = (g(s_1), \dots, g(s_n))$.

%

%\noindent \textbf{Approach.} %\lizhen{rewrite this subsection to focus on justifying key design choices and reproducibility. At most half page.}
%Following prior work on issue identification~\cite{purkayastha-etal-2025-lazyreview}, we evaluated zero-shot classification with open-source LLMs and found that even the best model misses at least 56\% of frequent issues, with or without fine-tuning.\footnote{See Figure \ref{fig:issue_identification_LLM_challenge}.} Since issue detection is inherently a reasoning task—requiring intermediate judgments (e.g., whether a request is \textit{nice-to-have} or \textit{must-have}) rather than surface-level matching—and the input space vastly exceeds the annotated data, mainstream approaches like fine-tuning are insufficient to instill this specialized reasoning capability.

%\noindent \textbf{Approach.} Issue detection is inherently a reasoning task—requiring intermediate judgments (e.g., whether a request is \textit{nice-to-have} or \textit{must-have}) rather than surface-level matching—and the input space vastly exceeds the annotated data, mainstream approaches like fine-tuning are insufficient to instill this specialized reasoning capability. To address those challenges, we propose a \textbf{neurosymbolic} method to decompose the issue detection task into two stages: \textit{(1) Abstract Feature Extraction}  and \textit{(2) Precision-oriented Classification}. 

\noindent \textbf{Approach.} Following prior work on issue identification~\cite{purkayastha-etal-2025-lazyreview}, we first evaluated zero-shot and fine-tuned LLM classification, both of which miss at least 45\% of frequent issues on our dataset (cf.~Fig.~\ref{fig:issue_identification_LLM_challenge}). We attribute this to two factors: issue detection requires intermediate judgments (e.g., whether an experiment request is \textit{nice-to-have} or \textit{must-have}) rather than surface matching, and the input space vastly exceeds the amount of available annotated training data. We therefore decompose the task into two stages: \textit{(1) Abstract Feature Extraction} and \textit{(2) Precision-Oriented Classification}.

\noindent \textit{\underline{1. Abstract Feature Extraction.}} For each issue class $\ell \in \mathcal{L}$, we prompt an LLM to generate a fixed set of $F_c$ issue-specific Yes/No questions $Q^{\ell} = \{q_1, \dots, q_{F_c}\}$ by comparing segments labeled with $\ell$ and identifying recurring patterns (cf.~Fig.~\ref{fig:abstract_q}). For a given segment $S_i$, the LLM then answers each question with $A_i^{\ell} \in \{\text{Yes}, \text{No}, \text{Not Relevant}\}$. Responses across all issue classes are aggregated and encoded as a fixed-length feature vector $\mathbf{v}_i \in \{-1, 0, 1\}^{|\mathcal{L}| \times F_c}$, mapping Yes/No/Not Relevant to $1$/$-1$/$0$. This converts the original reasoning task into pattern recognition over structured features, following prior work on LLM-based feature extraction~\cite{manikandan2023language, jeong2025llmselect}.

\begin{figure}[!t]
\begin{tcolorbox}[
  colback=gray!5, colframe=black,
  fontupper=\scriptsize,
    fonttitle=\small\bfseries,
  title={Abstract Feature Questions for Label `Extra Experiments'},
  boxrule=0.5mm, arc=1mm, fontupper=\small
]

\textbf{Comment:} \textit{``The paper does not explore more complex causal graphs
involving potential confounders and mediators.''}

\smallskip
\textbf{Abstract Feature Question (1):} Does the reviewer suggest further tests or missing results without
justifying why they are necessary?

\end{tcolorbox}
\caption{Example of abstract feature question generated by the LLM for a comment corresponding to the label `Extra Experiment'.}
\label{fig:abstract_q}
\vspace{-2mm}
\end{figure}
\noindent \textit{\underline{2. Precision-Oriented Classification}} Training classical machine learning models on these structured feature vectors captures non-deterministic correlations with issue labels while remaining effective in low-resource settings without large-scale finetuning~\cite{taye2023understanding}. To reduce false alarms and improve reviewer trust, we optimize precision-oriented issue detection using the \textbf{F\textsubscript{0.5}} score by tuning the hyperparameters on the validation set.
%\noindent \textbf{\textbf{2. Precision-Oriented Classification}} Training classical machine learning models on those structured feature vectors is necessary because the extracted features are correlated but not deterministically linked to issue labels. Additionally, these classifiers can easily be trained on low-resource setting circumventing the need for large-scale datasets as in other finetuning approaches~\cite{taye2023understanding}. Furthermore, to build trust among reviewers by minimizing false alarms, we prioritize precision over recall in issue detection. Accordingly, we optimize for the precision-oriented \textbf{F\textsubscript{0.5}} metric by selecting models maximizing F\textsubscript{0.5} on the validation set. 

Similar to prior LLM-assisted machine learning approaches~\cite{manikandan2023language, jeong2025llmselect}, our method converts raw text into structured representations, which captures intermediate reasoning steps explicitly and supports precision-oriented classification under limited supervision. Further details are provided in Appendix~\ref{sec:Issue_additional}.%, explicitly emphasizing precision in line with prior work~\cite{rei-yannakoudakis-2016-compositional}.  
\subsection{Feedback Generation} \label{sec:feedback_gen}
\noindent \textbf{Task definition and challenges.}
Let each segment be $s_i$ with label $l_i$ and optional context $C$ (e.g., abstract $A$). The goal is to produce feedback $F_i = g(s_i, l_i \mid C)$; without context, $F_i = g(s_i, l_i)$. For a review $R = (s_1, \dots, s_n)$ with labels $L = (l_1, \dots, l_n)$, the feedback sequence is $F = (F_1, \dots, F_n)$, supporting segment-only and context-informed generation.

Building on prior work on feedback generation~\cite{thakkar2025llmfeedbackenhancereview}, we evaluated zero-shot generation with open-weight LLMs. Many outputs either rewrote/extended entire reviews (“review extension”) or incorrectly addressed authors instead of reviewers (``role separation"). In our \textsc{LazyReviewPlus} dataset, at least 41\% of outputs were review extensions and 23\% showed role separation. These results motivate an inference-time approach that explicitly mitigates these issues.\footnote{See \S\ref{sec:exp_role} for details and examples.}

\noindent \textbf{Approach.}
To ensure targeted and diverse reviewer feedback, we adopt an iterative, reranking-based generation algorithm inspired by the genetic evolution approach of \citet{lee2025evolvingdeeperllmthinking}. Unlike standard zero-shot generation, which frequently produces review extensions or violates reviewer--author role separation, evolutionary search enables controlled exploration and refinement of candidate feedback through selection. This inference-time optimization encourages issue-focused, reviewer-aligned comments while reducing repetitive or off-target generations. Each piece of feedback is tailored to its corresponding issue, making it more actionable and constructive. The algorithm operates in six stages: \textit{(1) Template Construction}, \textit{(2) Plan Generation}, \textit{(3) Population Initialization}, \textit{(4) Fitness Evaluation}, \textit{(5) Parent Selection and Crossover}, and \textit{(6) Final Candidate Selection}. Stages (4)-(6) are repeated till the maximum number of iterations are reached within the genetic algorithm. Further details are provided in Appendix \ref{sec:details_feedback_generation}.

%\noindent \textit{\underline{1. Template Construction}}
%LLMs often produce less diverse or hallucinated outputs when unconstrained. Templates offer scaffolds that reduce hallucinations while maintaining control over diversity, effectively mitigating \emph{review extension}~\cite{kang2025template, xu2024jamplate}. We therefore design \textit{25} author-crafted issue-specific templates \( T = \{t_1, \dots, t_{25}\} \), each aligned with ACL ARR guideline issue types~\cite{Rogers_Augenstein_2021}.\footnote{Full list in Table~\ref{tab:templates}} All templates are verified and refined by senior authors for consistency and fit. Those templates can be viewed as issue-specific prompts, which can be adapted easily for new guidelines.

\noindent \textit{\underline{1. Template Construction}}
Unconstrained LLM generation often drifts off-task, producing \emph{review extensions} rather than targeted feedback. Templates provide structural scaffolds that constrain format while leaving the content open to adaptation, mitigating this drift~\cite{kang2025template, xu2024jamplate}. We design \textbf{25} author-crafted issue-specific templates $T = \{t_1, \dots, t_{25}\}$, one per ARR guideline category~\cite{Rogers_Augenstein_2021}, and refine them with senior authors for consistency.\footnote{Full list in Table~\ref{tab:templates}.} Because each template corresponds to a single issue, the set can be re-instantiated for new guidelines or other venues by curating templates against their issue taxonomy. In our setting, drafting all 25 templates took approximately 250 minutes.

%LLMs often generate less diverse or hallucinated outputs when unconstrained. Templates provide useful scaffolds for guiding less diverse but less hallucinated generations providing an effective way to mitigate `review extension'~\cite{kang2025template, xu2024jamplate}. We therefore design \textbf{25} author-crafted templates, $T = \{t_1, \dots, t_{25}\}$, each aligned with ACL ARR guideline issue types~\cite{Rogers_Augenstein_2021}.\footnote{Full list in Table~\ref{tab:templates}} The templates are double checked by senior authors in the project and are refined until deemed fit for each issue type.

\noindent \textit{\underline{2. Plan Generation}} 
To promote diversity in generated feedback, prior work uses model-generated plans to guide unconstrained generation and reduce uncertainty~\cite{narayan-etal-2023-conditional, huot-etal-2023-text}. For each review segment \(s_i\) and issue type \(l_i\), we prompt the LLM to produce a plan enriching the template using the paper’s abstract, reviewer summary, and noted strengths—mitigating the \emph{role separation} bias. The planner selects relevant knowledge \(k_i\) and justifies it with an explanation \(e_i\), forming the plan \(\mathcal{P}(s_i, l_i) = \{(k_i, e_i) \mid i = 1,\dots,N\}\).

%To ensure diversity in generated feedback, prior work leverages model-generated plans for unconstrained generation to reduce uncertainty~\cite{narayan-etal-2023-conditional, huot-etal-2023-text}. For each review segment $s_i$ and issue type $l_i$, we therefore, prompt the LLM to generate a plan to enrich the template based on the identified issue, using knowledge from the paper’s abstract, the reviewer’s summary, and their noted strengths. This, in turn, provides a way to mitigate the `role separation' issue in LLM-generated feedback.
%The planner selects relevant knowledge $k_i$ and justifies it with a explanation $e_i$, producing a plan \(
%\mathcal{P}(r_i, l_i) = \{(k_i, e_i) \mid i = 1,\dots,N\}.
%\)

\noindent \textit{\underline{3. Population Initialization}} The LLM is then prompted with the review segment, $s_i$, the plan, $P$ and the template, $t_i$ to generate $n$ sets of candidates. The initial population set, $M_0$ is thus given as, $F_0=\{fb_1, fb_2, \dots, fb_n \}$.

\noindent \textit{\underline{4. Fitness Evaluation}} Since we lack ground-truth feedback, we employ composite, verifiable rewards that combine positive scores and penalties to balance generated responses~\citep{hu2020makes}. We define a fitness function based on intrinsic text properties. Let $n_\text{sent}$ and $n_\text{words}$ denote the number of sentences and words. Following~\citep{yaacoub2025analyzingfeedbackmechanismsaigenerated}, we reward conciseness: $sc_{len} = \frac{\min(n_\text{sent}, n_\text{max})}{n_\text{max}}$, template adherence via n-gram overlap: $sc_{temp} = \frac{|\text{n-grams}_\text{fb} \cap \text{n-grams}_\text{temp}|}{|\text{n-grams}_\text{temp}|}$, readability via Flesch score~\cite{flesch1948new}: $sc_{read} = \frac{\text{Flesch score}}{100}$, and penalize off-task expressions: $pen_{forb} = \frac{\# \text{forbidden terms}}{n_\text{words}}$. The overall fitness is 
\(
fit(F_0) = - sc_{len} + sc_{temp} + sc_{read} - pen_{forb},
\)
enabling self-contained evaluation in LLM-driven \textit{feedback generation}.

\noindent \textit{\underline{5. Parent Selection and Crossover}} 
To \textit{select parents}, we use Boltzmann Tournament Selection~\cite{goldberg1990note}, converting fitness scores into probabilities via softmax:
\(
P({fb}_i) = \frac{\exp(fit({fb}_i)/\tau)}{\sum_{j=1}^{m} \exp(fit({fb}_j)/\tau)},
\)
where $\tau$ controls distribution sharpness. Candidates are sampled according to $P({fb}_i)$, favoring high-quality feedback while maintaining diversity, selecting exactly $n_{parents}$ for crossover. The \textit{crossover function} combines parent candidates, e.g., ${fb}_1$ and ${fb}_2$, to generate new offspring:
\(
Cross({fb}_1, {fb}_2) \rightarrow \text{new candidates},
\)
with the LLM prompted to integrate features from the parents, producing novel and diverse feedback.\footnote{When we prompted the LLM to mutate the population by generating semantically similar candidates as in \citet{pham-etal-2025-surveypilot}, approximately 62\% of the resulting population exhibited review extension and 31\% exhibited role separation. We therefore restrict exploration to fitness-selected, template-and-plan-grounded crossover and omit the mutation operator.}

\noindent \textit{\underline{6. Final Candidate selection}}
After a fixed number of generations, $n_{gen}$, we select the candidate with the maximal fitness score, 
\(
fb^* = \arg\max_{fb_i \in \mathcal{M}} fit(fb_i),
\) 
where \(\mathcal{M}\) is the set of final candidates. If multiple candidates achieve the same maximal score, 
we perform a cross-over of these candidates to generate a new candidate, which is taken as the final solution.\looseness=-1

\section{Experimental Setup} \label{sec:exp}

\noindent \textbf{Models.}
We evaluate our pipeline using LLMs from diverse families: Qwen 2.5 7B Instruct~\cite{qwen2025qwen25technicalreport} (Qwen), Yi 1.5 9B Chat~\cite{ai2025yiopenfoundationmodels} (Yi), Deepseek LLM 7B Chat~\cite{deepseekai2024deepseekllmscalingopensource} (Deep.), Phi-4 14B~\cite{abdin2024phi4technicalreport} (Phi), and GPT OSS 20B~\cite{openai2025gptoss120bgptoss20bmodel} (Oss.). For issue detection, we test various classifiers: KNN, Logistic Regression (L1/L2), Random Forest, Decision Trees, SVM (Linear/RBF/Poly), Gradient Boost, AdaBoost, Extra Trees, and MLP.\vspace{1mm}

\noindent \textbf{Evaluation Metrics.}
For \textit{\underline{feedback generation}}, no ground-truth references exist, so we evaluate along four criteria defined in consultation with senior authors: Constructiveness (Fig.~\ref{fig:constructiveness}), Relevance (Fig.~\ref{fig:relevance}), Specificity (Fig.~\ref{fig:specificity}), and Conciseness (Fig.~\ref{fig:conciseness}). For automated evaluation, we use \texttt{Prometheus}-V2~\cite{kim-etal-2024-prometheus} to score responses on a 1--5 scale per criterion, following prior work~\cite{sahnan2025llmsautomatefactcheckingarticle, maurya-etal-2025-unifying}; we select it over \texttt{Prometheus}-V1 and GPT-4o~\cite{openai2024gpt4ocard} based on the alignment analysis in \S\ref{sec:alignment}. For human evaluation, following prior work on peer review~\cite{purkayastha-etal-2025-lazyreview, lu-etal-2025-identifying}, three Ph.D. students fluent in English and experienced in peer reviewing rated a total of 100 responses per issue type from our best-performing model (Phi), applying the same criteria as \texttt{Prometheus}-V2.

For \textit{\underline{segment identification}}, we report Precision, Recall, and F1. For \textit{\underline{issue detection}}, we report macro-averaged $F_{0.5}$ and Precision, in line with the precision-oriented formulation in \S\ref{sec:issue_detect}. Both are computed against ground-truth annotations in \textsc{LazyReviewPlus}.
\vspace{1mm}

\noindent \textbf{Data Split and Validation.} Our \textit{feedback generation} and \textit{segment identification} approaches do not require training, and therefore results are reported on the full dataset. For \textit{issue identification}, we use 90\% of the data for training and 10\% for hyper-parameter tuning, with a 5-fold cross-validation split at the review level.\footnote{Details about the split are provided in \S\ref{sec:data_split}}

\noindent \textbf{Baselines.} 
Given a review segment $s_i$ and the issue set $\mathcal{L}=\{l_1, \dots, l_n\}$, we evaluate the following baselines for \textit{feedback generation}: (i) 1-pass (\texttt{1-pass}): the LLM generates feedback from $s_i$ and $l_i$; (ii) Template-guided (\texttt{Temp}): the LLM incorporates a template $t_i$ with $s_i$ and $l_i$; (iii) Plan-then-generate (\texttt{Plan}): the LLM devises a plan $\mathcal{P}i$ for $l_i$ and then integrates it into $t_i$ to generate feedback; (iv) Best-of-N (\texttt{BoN})~\cite{brown2024largelanguagemonkeysscaling}: we sample $N=n\times n_{gen}$ candidates and select the best using our fitness function; (v) Self-refinement (\texttt{Self-Ref.})~\cite{madaan2023selfrefine}: we sample $n$ candidates and refine each over $n_{gen}$ iterations, retaining the final output. (vi) Template-Plan-Best-of-N~(Temp + Plan): the LLM generates a plan from the review segment ($s_i$), issue label ($l_i$), and template ($t_i$), produces $n$ candidates, and ranks them using our fitness function.
\footnote{Hyperparameters are listed in Table~\ref{tab:ga_hyperparameters}.}\looseness=-1

\section{Results and Discussion}
\label{sec:results}
%In this section, we present the results of different subtasks within our approach. Following prior work~\cite{lan-etal-2024-multi, pichler-etal-2025-evaluating}, for \textit{\underline{segment identification}}, we adapt a zero-shot approach of detecting segments within a review as tagging sentences with a \textbf{B-I-O} tag. We achieve an Precision, Recall and F1 score of \textbf{0.81}, \textbf{0.77} and \textbf{0.79} respectively using Phi.\footnote{Details about the segment identification experiment are in \S\ref{sec:segment_id}.} Following our setup for \textit{\underline{issue detection}} in Sec~\S\ref{sec:issue_detect}, we obtain best results with Extra Trees classifier augmented with Phi-generated abstract features yielding F0.5 score of \textbf{0.51} and Precision score of \textbf{0.75}. We therefore select Phi for \textit{\underline{segment identification}} and abstract feature generation in \textit{\underline{issue identification}}, combined with the Extra Trees classifier, for the controlled review rewriting study in \S\ref{sec:exp_rewrite}. \footnote{The detailed experimental setup and results for \textbf{issue detection} are presented in Sec~\S\ref{sec:Issue_additional}}  The results for \textbf{feedback generation} are outlined below.
\begin{table}[!t]
\centering
\resizebox{!}{0.3\columnwidth}{
\begin{tabular}{llccccc}
\toprule
 \textbf{Method} & \textbf{Yi} & \textbf{Phi} & \textbf{Qwen} & \textbf{Deep.} & \textbf{Oss.} \\
\midrule
 \texttt{LLM\textsubscript{Zero}} & 0.02 & 0.09 & 0.18 & 0.03 & 0.29 \\
     \texttt{LLM\textsubscript{Fine}} & 0.12 & 0.15 & 0.09 & 0.11 & 0.09 \\
     \texttt{LLM\textsubscript{QA}} & 0.13 & 0.13 & 0.12 & 0.20 & 0.14 \\
     \texttt{LLM\textsubscript{QAFine}} & 0.05 & 0.10 & 0.03 & 0.08 & 0.08 \\
\midrule
 \texttt{RoBERTa} &\multicolumn{5}{c}{0.58} \\ \midrule
 \texttt{KNN} &0.60 &0.65 &0.62 &0.65 &0.66 \\
 \texttt{LogReg-L2} &0.55 &0.58 &0.58 &0.55 &0.62 \\
 \texttt{LogReg-L1} &0.54 &0.59 &0.61 &0.55 &0.64 \\
 \texttt{RF} &0.66 &0.66 &0.68 &0.62 &0.67 \\
 \texttt{DT} &0.26 &0.27 &0.24 &0.47 &0.55 \\
 \texttt{SVM-RBF} &0.55 &0.60 &0.65 &0.66 &0.66 \\
 \texttt{SVM-Lin} & 0.60 & 0.53 & 0.55 & 0.57 & 0.62 \\
 %\texttt{SVM-Poly} & 0.01 & 0.01 & 0.12 & 0.02 & 0.03 \\
 %\texttt{GNB} & 0.26 & 0.31 & 0.26 & 0.26 & 0.26 \\
%& \texttt{SGD} & 0.45 & 0.44 & 0.42 & 0.37 & 0.38 \\
 \texttt{GBoost} & 0.60 & 0.60 & 0.64 & 0.59 & 0.66 \\
 \texttt{AdaB} & 0.07 & 0.08 & 0.16 & 0.27 & 0.10 \\
 \texttt{ExtraT} & \hlgray{\textbf{0.68}} & \textbf{0.65} & \textbf{0.67} & \textbf{0.61} & \hlgray{\textbf{0.68}} \\
 \texttt{MLP} & 0.61 & 0.62 & 0.61 & 0.62 & 0.67 \\ 
\bottomrule
\end{tabular}}
\caption{Cross-validation F$_{0.5}$ scores for issue detection across LLMs and the RoBERTa baseline. The best classifier is in bold, and the best LLM is \hlgray{highlighted}.}
\label{tab:methods_comparison}
\vspace{-4mm}
\end{table}
In this section, we present results across the three subtasks of our pipeline, which together establish the configuration used in the downstream review-rewriting study in \S\ref{sec:exp_rewrite}.

For \textit{\underline{segment identification}}, following prior work on LLM-based sequence labeling~\cite{lan-etal-2024-multi, pichler-etal-2025-evaluating}, we cast segment detection as sentence-level \textbf{B-I-O} tagging in a zero-shot setting (full prompt and tag definitions in \S\ref{sec:segment_id}). Among the five LLMs evaluated, Phi achieves the best performance with Precision, Recall, and F1 scores of \textbf{0.81}, \textbf{0.77}, and \textbf{0.79}, outperforming the next-best model by at least 0.06 F1.\footnote{The full experimental setup is provided in Sec~\S\ref{sec:segment_id_full}}

For \textit{\underline{issue detection}}, following the precision-oriented setup in \S\ref{sec:issue_detect}, the strongest configuration combines Yi or GPT Oss-generated abstract features with an Extra Trees classifier, yielding macro-averaged $F_{0.5}$ of \textbf{0.68} and Precision of \textbf{0.65} under 5-fold cross-validation. This outperforms the strongest baseline (\texttt{RoBERTa}) by 0.10 $F_{0.5}$ points. Per-fold variance is below 0.01 (cf.~Table~\ref{tab:methods_comparison} and Table ~\ref{tab:precision_recall}), and the full ablation, baseline comparisons, and experimental setup are reported in \S\ref{sec:Issue_additional}.

Based on these results, we adopt Phi for both segment identification and abstract feature generation, paired with the Extra Trees classifier for issue detection, throughout the remaining experiments. This is based on a single model for the pipeline. With the upstream components fixed, the following sections evaluate the central contribution of our framework for \textit{feedback generation}.

\begin{table}[!t]
\centering
\resizebox{0.8\columnwidth}{!}{%
\begin{tabular}{llcccccccc}
\toprule
\textbf{Metric} & \textbf{Model} & \texttt{1-pass} & \texttt{BoN} & \texttt{Self-Ref.} & \texttt{Temp} & \texttt{Plan} & \texttt{Temp+Plan} & \texttt{Ours} \\
\midrule
Const. & Yi       & 1.9 & 2.0 & 2.3 & 2.8 & 3.0 & 3.6 & 3.9 \\
Rel.    & Yi       & 2.0 & 2.2 & 2.4 & 2.9 & 3.1 & 3.5 & 3.8 \\
\midrule
Const. & Phi      & \underline{2.1} & \underline{2.2} & \underline{2.4} & \underline{3.1} & \underline{3.3} & 3.4 & \underline{\textbf{4.3}} \\
Rel.    & Phi      & \underline{2.2} & \underline{2.3} & \underline{2.6} & \underline{3.2} & \underline{3.4} & 3.5 & \underline{\textbf{4.3}} \\
\midrule
Const. & Qwen     & 1.8 & 2.0 & 2.2 & 2.9 & 3.1 & 3.4 & 3.8 \\
Rel.    & Qwen     & 1.9 & 2.1 & 2.4 & 3.0 & 3.2 & 3.4 & 3.8 \\
\midrule
Const. & Deep.    & 1.7 & 1.8 & 2.4 & 2.7 & 2.9 & 3.2 & 3.5 \\
Rel.    & Deep.    & 1.8 & 1.9 & 2.4 & 2.8 & 3.0 & 3.3 & 3.6 \\
\midrule
Const. & Oss.     & 1.9 & 2.0 & 3.0 & 2.8 & 3.0 & 3.2 & 4.0 \\
Rel.    & Oss.     & 2.0 & 2.1 & 3.1 & 2.9 & 3.1 & 3.4 & 4.1 \\
\bottomrule
\end{tabular}}
\caption{Model performance on \textbf{feedback generation} across six methods for \textbf{Constructiveness (Const.)} and \textbf{Relevance (Rel.)}. Overall best results are \textbf{bolded}, method-wise best are \underline{underlined}.}

\label{tab:feedback_models_metrics}
\end{table}

\begin{table}[!ht]
\centering
\resizebox{!}{0.1\textwidth}{
\begin{tabular}{l c c c c}
\toprule
\textbf{Method} & \textbf{Const.} & \textbf{Relev.} & \textbf{Spec.} & \textbf{Conc.} \\
\midrule
\texttt{1-pass} & 1.8 & 1.9 & 2.2 & 2.2 \\ \cmidrule(lr){2-5}
\texttt{BoN} & 2.4 & 2.4 & 2.5 & 2.5 \\ \cmidrule(lr){2-5}
\texttt{Self-Ref.} & 2.5 & 2.6 & 2.6 & 2.6 \\ \cmidrule(lr){2-5}
\texttt{Temp} & 2.8 & 2.7 & 2.8 & 2.8 \\ \cmidrule(lr){2-5}
\texttt{Plan} & 3.2 & 3.1 & 3.2 & 3.2 \\ \cmidrule(lr){2-5}
\texttt{Temp + Plan} & 3.4 & 3.4 & 3.6 & 3.6 \\ \cmidrule(lr){2-5}
\texttt{Ours} & \textbf{4.1} & \textbf{4.2} & \textbf{4.0} & \textbf{4.3} \\
\bottomrule
\end{tabular}}
\caption{Human evaluation of the best-performing model (Phi) on the \textbf{feedback generation} task, reporting Constructiveness (Const.), Relevance (Relev.), Specificity (Spec.) and Conciseness (Conc.)}
\label{tab:feedback_human}
\end{table}

\begin{table}[!t]
\centering
\resizebox{0.65\columnwidth}{!}{%
\begin{tabular}{lcc}
\toprule
\textbf{Variant} & \textbf{Const.} & \textbf{Rel.} \\
\midrule
Full Algorithm (Ours)              & \textbf{4.3} & \textbf{4.3} \\
\midrule
$\quad$ w/o. Template Construction          & 3.6 & 3.5 \\
$\quad$ w/o. Plan Generation                & 3.8 & 3.8 \\
$\quad$ w/o. Population Initialization      & 3.9 & 3.9 \\
$\quad$ w/o. Fitness Evaluation             & 3.7 & 3.7 \\
$\quad$ w/o. Parent Selection \& Crossover  & 3.9 & 3.8 \\
$\quad$ w/o. Final Candidate Selection      & 3.8 & 3.7 \\
\bottomrule
\end{tabular}}
\caption{Ablation study of the various components for the best performing model, Phi within the iterative, reranking-based generation algorithm focusing on the automated metrics.}
\label{tab:ablation_short}
\end{table}

\begin{table}[!t]
\centering
\resizebox{0.8\columnwidth}{!}{%
\begin{tabular}{l l c c}
\toprule
\textbf{Model} & \textbf{Variant} & \textbf{Const.} & \textbf{Relev.} \\
\midrule
Yi    & Full Fitness Function                   & \textbf{3.9} & \textbf{3.8} \\
\cmidrule(lr){2-4}
      & $\quad$ w/o Length Score ($sc_{len}$)  & 3.6 & 3.6 \\
      & $\quad$ w/o Template Adherence ($sc_{temp}$) & 3.7 & 3.7 \\
      & $\quad$ w/o Readability ($sc_{read}$)  & 3.8 & 3.8 \\
\midrule
Qwen  & Full Fitness Function                   & \textbf{3.8} & \textbf{3.8} \\
\cmidrule(lr){2-4}
      & $\quad$ w/o Length Score ($sc_{len}$)  & 3.5 & 3.5 \\
      & $\quad$ w/o Template Adherence ($sc_{temp}$) & 3.6 & 3.6 \\
      & $\quad$ w/o Readability ($sc_{read}$)  & 3.7 & 3.7 \\
\midrule
DeepSeek & Full Fitness Function                & \textbf{3.5} & \textbf{3.6} \\
\cmidrule(lr){2-4}
         & $\quad$ w/o Length Score ($sc_{len}$) & 3.2 & 3.3 \\
         & $\quad$ w/o Template Adherence ($sc_{temp}$) & 3.3 & 3.4 \\
         & $\quad$ w/o Readability ($sc_{read}$) & 3.4 & 3.5 \\
\midrule
Oss.   & Full Fitness Function                   & \textbf{4.0} & \textbf{4.1} \\
\cmidrule(lr){2-4}
      & $\quad$ w/o Length Score ($sc_{len}$)  & 3.7 & 3.6 \\
      & $\quad$ w/o Template Adherence ($sc_{temp}$) & 3.8 & 3.7 \\
      & $\quad$ w/o Readability ($sc_{read}$)  & 3.9 & 4.0 \\
\midrule
Phi   & Full Fitness Function                   & \textbf{4.3} & \textbf{4.3} \\
\cmidrule(lr){2-4}
      & $\quad$ w/o Length Score ($sc_{len}$)  & 3.8 & 3.7 \\
      & $\quad$ w/o Template Adherence ($sc_{temp}$) & 3.7 & 3.6 \\
      & $\quad$ w/o Readability ($sc_{read}$)  & 4.1 & 4.0 \\
      & $\quad$ w/o Forbidden Term Penalty ($pen_{forb}$) & 4.2 & 4.2 \\
\bottomrule
\end{tabular}}
\caption{Ablation study on the various rewards within the fitness function in the feedback generation approach, reporting Constructiveness (Const.) and Relevance (Relev.) across models.}
\label{tab:fitness_ablation}
\end{table}

\begin{figure*}
\centering

\begin{tcolorbox}[
    width=0.95\textwidth,
    colback=gray!10,
    colframe=black,
    title=Example Feedback: Extra Experiments,
    boxrule=0.4mm,
    arc=1.5mm,
    fontupper=\scriptsize,
    fonttitle=\small\bfseries,
    boxsep=2pt,
    left=2pt,
    right=2pt,
    top=2pt,
    bottom=2pt,
]

\textbf{Issue Detected:} The authors could also do [extra experiment X]

\textbf{Reviewer Comment:} "...no check for false negatives\ldots\ only one language, one domain, one genre\ldots\ no significance testing"

\textbf{Feedback:} Your comment highlights suggestions for additional experiments and concerns about generalization. While extra experiments (e.g., multilingual evaluation or statistical tests) are possible, the paper provides sufficient evidence to support its claims. The contribution introduces the XXX benchmark and a temporal clustering method that improves retrieval-augmented generation for emerging mentions. These contributions are demonstrated convincingly within the chosen scope. Monolingual studies (English, social media genre) are still valuable practically and theoretically. If you believe significance testing or cross-domain validation is essential for validity, please clearly justify why; otherwise, these points are better framed as suggestions rather than reasons for rejection.

\end{tcolorbox}

\caption{Feedback generated using our approach with Phi on an instance of our dataset, \textsc{LazyReviewPlus}.}
\label{fig:example}

\end{figure*}

\subsection{Effectiveness of  \textbf{feedback generation}   }
We present the results of our \textbf{automated evaluation} in Table~\ref{tab:feedback_models_metrics}.\footnote{Full results in Table~\ref{tab:feedback_grouped_model}} along with the feedback generated by our system in Fig~\ref{fig:example}.  We find that our approach consistently outperforms all baselines, producing targeted, diverse feedback with the same number of candidates. While Best-of-N (\texttt{BoN}) boosts diversity, it lacks structure; Self-refinement (\texttt{Self-refine}) improves quality but still drifts off-task without external rewards~\cite{lee2025evolvingdeeperllmthinking}.%We find that our approach consistently outperforms all baselines, generating targeted and diverse feedback with the same number of candidates generated as the baselines. We observe that Best-of-N (\texttt{BoN}) increases diversity but lacks structure, while Self-refinement (\texttt{Self-refine}) progressively improves quality yet still deviates from the task without external rewards~\cite{lee2025evolvingdeeperllmthinking}. 

%We also find that template-based prompting (\texttt{Temp}) boosts performance—Phi improves from 2.1 to 3.1 in \textit{Constructiveness} and 2.2 to 3.2 in \textit{Relevance}~\cite{xu2024jamplate, kang2025template}. The Plan-then-Generate paradigm yields modest gains (e.g., \textit{Constructiveness} 3.1 → 3.3), suggesting that structured planning helps, though template scaffolding drives most of the improvement. Overall, we find Phi to be the strongest performer, likely due to its scale and diverse pretraining~\cite{abdin2024phi4technicalreport}. For \textbf{human evaluation}, we achieve substantial inter-annotator agreement in terms of Krippendorff's $\alpha$ across all metrics ( \textit{Constructiveness}: 0.65, \textit{Relevance}: 0.68, \textit{Specificity}: 0.72, \textit{Conciseness}: 0.72). All agreement values are statistically significant: Constructiveness CI = [0.603, 0.697], Relevance CI = [0.635, 0.725], and both Specificity and Conciseness CI = [0.678, 0.762], indicating the results are robust and likely generalizable to a larger population. The results for the \textit{human evaluation}~(cf. Table~\ref{tab:feedback_human}) for the best-performing model, Phi-4 reveals a similar pattern as our \textit{automated evaluation}, with our approach outperforming baselines across metrics and the rankings across baselines being the same. The human alignment with the evaluator, \texttt{Prometheus} is strong (Spearman $\rho$ = 0.85), validating the automated metrics used.\footnote{The alignment study is in \S\ref{sec:alignment}}

We also find that template-based prompting (\texttt{Temp}) consistently boosts performance over zero-shot generation. Phi improves from 2.1 to 3.1 in \textit{Constructiveness} and from 2.2 to 3.2 in \textit{Relevance}~\cite{xu2024jamplate, kang2025template}. The Plan-then-Generate paradigm yields more modest additional gains (e.g., \textit{Constructiveness} improving from 3.1 to 3.3), suggesting that structured planning contributes, but that template scaffolding drives most of the improvement. The Template-Plan-Best-of-N strategy improves performance across the board (e.g., Constructiveness improving from 3.1 to 3.4 for Qwen). However, without parent selection and crossover, the population it generates remains constrained by its own internal distribution and consequently lags behind our proposed algorithm. Across models, Phi is the strongest performer, consistent with its reported strength on reasoning-oriented tasks~\cite{abdin2024phi4technicalreport}.

\noindent For \textbf{human evaluation} on the best-performing model (Phi), we obtain substantial inter-annotator agreement across all four metrics (Krippendorff's $\alpha$: \textit{Constructiveness} 0.65, \textit{Relevance} 0.68, \textit{Specificity} 0.72, \textit{Conciseness} 0.72), with all confidence intervals lying within $[0.60, 0.77]$. The human rankings (Table~\ref{tab:feedback_human}) mirror the automated ones: our approach outperforms every baseline on every metric, and the baseline ordering is preserved. Human ratings align strongly with \texttt{Prometheus} (Spearman $\rho = 0.85$), supporting its use as the automated evaluator throughout this paper.\footnote{Full alignment analysis in \S\ref{sec:alignment}.}

\subsection{Ablation Study for Feedback Generation} We perform a series of ablation studies to establish the contribution of each aspect within our approach. \textit{\underline{i. Effect of components within our approach}}: We ablate our algorithm in Table~\ref{tab:ablation_short} for Phi and find that removing the Template Construction phase drops Constructiveness by $\sim$ 0.7 points for Phi, as templates provide structured scaffolds that reduce \textit{review extension}. Removing the fitness function lowers scores by $\sim$ 0.6 points and final candidate is often longer than its parents, revealing \textit{length bias}, a common reward hacking shortcoming in LLM outputs~\cite{epoch2025outputlength}. We obtain similar results for the other models in this work.\footnote{Full ablation for all models in Table~\ref{tab:ablation}.}  
 \textit{\underline{ii. Effect of individual rewards}}: We analyze each reward in our \textit{fitness function} (cf. Table~\ref{tab:fitness_ablation}) and find that removing the template adherence ($sc_{temp}$) reward drops Constructiveness by 0.6 points for Phi as outputs go off-topic. Removing the length-based reward ($sc_{len}$) reduces Conciseness by 0.6 points, reflecting how longer responses become convoluted and harder to read. Removing the cross-over component also leads to a decline in the constructiveness scores by 0.4 points pointing to the selection of less effective candidates.~\footnote{Full ablation table in Table~\ref{tab:fitness_ablation_long}} 
 \noindent \textit{\underline{iii. Effect of planner inputs.}} We ablate the three knowledge sources passed to the planner (paper abstract, reviewer-written summary, and reviewer-written strengths) in Table~\ref{tab:planner_ablation}. Removing any single source reduces Phi's \textit{Constructiveness} by 0.1--0.3 points, and removing two sources reduces it by up to 0.6 points, indicating that each input contributes complementary guidance rather than redundant signal. The effect is most pronounced for frequent, context-dependent issues such as ``extra experiments,'' where grounding feedback requires the abstract for scope, the summary for the reviewer's framing, and the strengths to avoid contradicting acknowledged contributions.\looseness=-1
\section{Expert Assessment of Review Quality Improvement Through Feedback} \label{sec:exp_rewrite}

\noindent \textbf{Setup.} To evaluate the quality of the generated feedback, we conduct a controlled study comparing the original reviews, rewrites based only on the detected \textit{lazy thinking} and \textit{specificity issues}, and rewrites that additionally incorporate the generated feedback. Following \citet{purkayastha-etal-2025-lazyreview}, we form two groups of two Ph.D. students each. One rewrites 50 reviews using only issues from our \textit{issue detection} module, emulating the setup in prior works~\cite{10.1371/journal.pone.0320444, purkayastha-etal-2025-lazyreview}, while the other uses both issues and targeted feedback from our \textit{feedback generation} approach, with each feedback item marked as \textit{actionable} or not. Both groups have access to the papers corresponding to the reviews. Rewrites are evaluated on \textit{Constructiveness}, \textit{Justified}, and \textit{Adherence}. Two senior PhD students who are not part of the reviewing pool act as judges to perform pairwise comparisons in a \textbf{blind annotation setup} between original reviews, issue-only rewrites, and issue-plus-feedback rewrites, splitting 50 reviews evenly and reserving 10 for agreement. We further ask the judges to annotate the issues in these reviews following our issue detection schema introduced in Sec~\S\ref{sec:background} and also provide a list of issues identified within each of these reviews.

\noindent \textbf{Results.}
We report the win-tie-loss results for the controlled experiment in Table~\ref{tab:orig-lazy-feedback}. Reviews rewritten with feedback (Issue det. \textit{w/} feed) outperform both reviews written with only \textit{issue detection} (Issue det.) and the original reviews (Orig), with 85\% and 95\% win rates on Constructiveness, respectively. Issue-only feedback improves over original reviews, consistent with \citet{purkayastha-etal-2025-lazyreview}. Reviewers in the feedback-based rewrite group incorporated 96\% of the feedback, marking it actionable. As measured by our automatic detector, feedback-based rewrites reduce identified issues by up to \textbf{92.4}\%, compared to 85.2\% for rewrites with only \textit{issue detection} (relative to original reviews), demonstrating the effectiveness of the \textit{feedback generation} module. Following~\citet{purkayastha-etal-2025-lazyreview}, a Bradley-Terry preference model~\cite{bradley1952rank} trained on adherence preferences yields scores of 1.2 for feedback-based rewrites, -1.0 for original reviews, and 0.6 for \textit{issue detection}-based rewrites. This corresponds to a 93.4\% win rate over original texts and 78.5\% over issue-detection rewrites, further supporting the effectiveness of the feedback-generation approach. Inter-annotator agreement (Krippendorff's $\alpha$) is 0.71, 0.73, and 0.77 for Constructiveness, Justification, and Adherence, respectively. Using the judges' gold annotations (Krippendorff's $\alpha$ = 0.67) instead, we observe an 86.7\% issue reduction between original and feedback-assisted reviews, compared to 74.1\% between original and issue-detected reviews. The detector in our approach and the gold annotations from the judges achieved a high correlation (Spearman $\rho$ = 0.74, $p<0.01$), outlining the reliability of our detector in identifying issue types consistent with human judgment.

Beyond issue reduction, we also examine rewrite quality more closely: the detector identified only 60.0\% of true issues, yet close agreement with judges suggests reviewers addressed issues beyond the detected subset. Rewrites rarely introduced new problems, only 8\% added a new specificity issue and largely preserved original concerns, with 96\% of those items retained (18\% of which were reorganized into the comments/suggestions section rather than removed). The issue-retention study had an Krippendorff's $\alpha$ = 0.94.

\begin{table}[!t]
\centering
\resizebox{!}{0.1\columnwidth}{%
\begin{tabular}{lcccc}
\hline
\textbf{Type} & \makecell{\textbf{Cons.} \\ (W/T/L)} & \makecell{\textbf{Just.} \\ (W/T/L)} & \makecell{\textbf{Adh.} \\ (W/T/L)} & \makecell{$\Delta$ \textbf{Issue} \\ ($\uparrow$)}  \\
\hline
Issue det. vs. Orig              & 85/5/10  & 80/5/15  & 85/5/10 & 85.2  \\
Issue det. \textit{w} feed vs Orig       & 95/5/0   & 95/5/0   & 90/5/5 & \textbf{92.4}  \\
Issue det. \textit{w} feed vs. Issu det.      & 85/10/5  & 90/5/5   & 85/10/5 & 78.5  \\
\hline
\end{tabular}%
}
\caption{
Comparison of Original, Issue-detected, and Issue-detected with feedback re-written reviews on Constructiveness, Justification, and Adherence using Win/Tie/Loss rates. $\Delta$\textbf{Issue} indicates issue reduction.}
\label{tab:orig-lazy-feedback}
\end{table}

\section{Related Work}
\noindent \textbf{Review Quality Assessment.} 
As scientific publications rise, reviewers face increasing burden, driving interest in assessing review quality~\cite{kuznetsov2024naturallanguageprocessingpeer}. Prior studies have measured tonality~\cite{bharti2024politepeer}, thoroughness~\cite{lu-etal-2025-identifying} and substantiation of review claims~\cite{guo-etal-2023-automatic}. \citet{sadallah2025goodbadconstructiveautomatically} evaluate reviews based on constructiveness for author revisions, while \citet{du-etal-2024-llms} examine deficiencies in both human and LLM-generated reviews. In contrast, we focus on detecting cognitive biases (\textit{lazy thinking}) and \textit{specificity} issues following ACL ARR guidelines~\cite{Rogers_Augenstein_2021}, and propose a targeted feedback mechanism to improve review quality prior to submission. %Prior work targets single-segment, single-issue detection~\cite{purkayastha-etal-2025-lazyreview}; we extend this by identifying issues with high precision followed by focused, diverse feedback for each issue type.

%As the number of scientific publications continues to rise steeply, overburdening reviewers, interest in assessing the quality of reviews has increasingly grown within the NLP community~\cite{kuznetsov2024naturallanguageprocessingpeer}. Several studies have evaluated review quality in terms of tonality~\cite{bharti2024politepeer}, thoroughness~\cite{lu2025identifyingaspectspeerreviews}, and constructiveness~\cite{sadallah2025goodbadconstructiveautomatically}. In this work, we focus on detecting and providing feedback for review comments reflecting cognitive biases—termed \textit{lazy thinking}—as well as \textit{stylistic} issues, in line with ACL ARR guidelines~\cite{Rogers_Augenstein_2021}. Prior work in this area has mostly addressed issue detection in a single-segment, single-issue setup, without generating targeted verbal feedback to mitigate such flaws~\cite{purkayastha-etal-2025-lazyreview}. We close this gap by generating focused and diverse feedback tailored to each issue outlined in the ACL ARR guidelines~\cite{Rogers_Augenstein_2021}. 

\noindent \textbf{LLM-generated feedback.} 
LLMs have been widely used for feedback, especially in education, supporting English learners~\cite{kasneci2023chatgpt, escalante2023ai}, coding instruction~\cite{misiejuk2024augmenting, azaiz2024feedback}, and TOEFL preparation~\cite{kasneci2023chatgpt}. Studies show LLM feedback matches human quality and drives positive outcomes. Building on this, \citet{thakkar2025llmfeedbackenhancereview} used Claude at ICLR 2025 to improve review specificity, correctness, and politeness. However, their method ignores guideline compliance and relies on a closed-source model. We address both by enforcing ACL ARR adherence and using open-weight LLMs, ensuring integrity and privacy.\looseness=-1

%LLMs have been widely explored for providing feedback, particularly in education, such as assisting English learners~\cite{kasneci2023chatgpt, escalante2023ai}, supporting teachers in coding content creation~\cite{misiejuk2024augmenting, azaiz2024feedback}, and helping students prepare for the TOEFL iBT exam~\cite{kasneci2023chatgpt}. These studies show that LLM-generated feedback is comparable to human feedback and effectively drives positive outcomes. Extending this, \citet{thakkar2025llmfeedbackenhancereview} conducted a controlled experiment at ICLR 2025 using a closed-source model (Claude) for \hlgreen{feedback generation} aimed at improving reviews by addressing misunderstandings, enhancing specificity, and promoting politeness. While effective, their approach does not ensure conference/journal guideline adherence and raises ethical concerns due to reliance on a closed-source model. Our approach addresses both issues via issue detection for ACL ARR compliance and deploying open-source LLMs, preserving integrity and privacy.

\section{Conclusion}
To enhance review quality, we introduce an open-weight, LLM-driven framework that identifies issues precisely and generates targeted feedback via iteratively refined templates. It outperforms zero-shot and fine-tuned LLM-based baselines, and explicit feedback further improves review quality. Additionally, we release \textsc{LazyReviewPlus}, an expert-annotated dataset of review sentences labeled for detecting \textit{lazy thinking} and lack of \textit{specificity}. We hope our work inspires further efforts to enhance the quality of peer review.

\section*{Limitations}
In this work, we propose a new dataset alongside a precision-oriented issue identification and feedback generation framework designed to enhance the quality of peer reviews. However, our approach comes with several limitations.  

First, we introduce \textsc{LazyReviewPlus}, a single-segment, multi-label classification dataset for detecting \textit{lazy thinking} and \textit{specificity} issues in peer reviews. The dataset currently covers only conferences hosted on ACL Rolling Review (ARR), due to their well-defined guidelines for \textit{lazy thinking} and \textit{specificity}. Extending this resource to other domains, conferences, or languages would require developing equally rigorous guidelines with precise definitions for these issues. 

Second, in terms of methodology, we conduct extensive experiments and analyses for both \textbf{issue identification} and \textbf{feedback generation}. However, these results may not be fully generalizable to other scientific or evaluative tasks, and thus the results should be interpreted with caution. Moreover, our feedback generation approach relies on predefined templates that are refined using an iterative, reranking-based generation algorithm. Over time, however, reviewers may become desensitized to recurring feedback styles. The long-term effectiveness and robustness of a template-driven approach therefore warrant further investigation, especially in repeated or large-scale deployment settings. Additionally, deploying our approach requires curating templates for each conference upon the introduction of new guidelines. Nevertheless, curating 25 templates for the ACL Rolling Review guidelines required only 250 minutes — a modest one-time overhead relative to the substantial resources involved in organizing a conference.  Moreover, while addressing \textit{lazy thinking} and improving \textit{specificity} is an attempt to enhance review quality, overall scientific outcomes depend on multiple confounding factors, including paper quality, author rebuttals, and meta-reviewer decisions. Evaluating the long-term impact of guideline-compliant reviews on acceptance decisions and scientific progress remains an important direction for future work. Additionally, our feedback generation algorithm is bounded by upstream detection recall: the detector identified 60.0\% of true issues confirmed by blind human annotation in our controlled study. Improving recall, e.g. via the tunable operating points as in Table~\ref{tab:model_thresholds}, remains an important direction for extending issue coverage in future work.

Third, our framework relies on large language models, which may carry stylistic or cultural biases inherited from their training data. Although peer reviewing is intended to be domain-neutral, such biases may still manifest subtly in phrasing or prioritization. Future work could incorporate bias detection and mitigation strategies to enhance fairness and neutrality in feedback generation.  

Finally, while automated feedback systems can assist reviewers and improve the clarity and constructiveness of reviews, they should complement—rather than replace—human judgment. The ultimate responsibility for fair, constructive, and contextually informed reviewing must remain with human reviewers. Our controlled study (§6) evaluates feedback quality following the design of \citet{purkayastha-etal-2025-lazyreview}; it does not measure voluntary adoption as is the case in real-world reviewing. We view this evaluation as a necessary first step toward establishing that the generated feedback provides value when adopted, which should precede future work as the ReVAS tool used in ARR to test this setup in real-world settings.\footnote{\url{https://revas.mbzuai.ac.ae/}}

\section*{Ethics Statement}
Before commencing the project, we obtained ethics approvals from the IRB of TU Darmstadt (Project No. EK 31/2025: `Uncertainty-Guided Evolving Thinking for Review Feedback Agent') and from Monash University (Project No. 45733: `Review Feedback Agent'), respectively. Our dataset, \textsc{LazyReviewPlus}, has been released at \url{https://github.com/UKPLab/emnlp2026-reviewfeedbackagent} under the CC-BY-NC-4.0 license. The underlying ARR reviews were collected from \textsc{NLPeer}~\cite{dycke-etal-2023-nlpeer}, which is licensed under the same terms. The analysis and automatic annotation processes do not involve any personal or sensitive information. The annotators participating in the user study and dataset annotation were compensated at a rate of \$25 per hour, ensuring fair remuneration.

\section*{Acknowledgements}
This work has been co-funded by the German Research Foundation (DFG) as part of the Research Training Group KRITIS No. GRK 2222, the European Union (ERC, InterText, 101054961) and the LOEWE Distinguished Chair ``Ubiquitous Knowledge Processing'', LOEWE initiative, Hesse, Germany (Grant Number: LOEWE/4a//519/05/00.002(0002)/81). Views and opinions expressed are however those of the author(s) only and do not necessarily reflect those of the European Union or the European Research Council. Neither the European Union nor the granting authority can be held responsible for them. We gratefully acknowledge the support of Microsoft with a grant for access to OpenAI GPT models via the Azure cloud (Accelerate Foundation Model Academic Research). The work of Anne Lauscher is
funded under the Excellence Strategy of the German Federal Government and the Federal States.
The authors acknowledge the support of Schloss
Dagstuhl – Leibniz Center for Informatics through
the Dagstuhl Seminar \textit{`24052: Reviewer No. 2: Old and New Problems in Peer Review'}.

We thank Zhuang Li, Tao Feng, Christina Kang, Yvette Liu, Shuo Huang, and Tianqiang Yan (Henry) for their help with dataset annotation and evaluation of the generated feedback; and Tim Baumgärtner and Sheng Lu for their feedback on a draft of the paper.

\bibliography{custom}

@article{ware2015stm,
  title={An overview of scientific and scholarly journal publishing},
  author={Ware, Mark and Mabe, Michael},
   journal={The {STM} {R}eport},
  year={2015},
  pages={1082:1083},
  note={\url{https://digitalcommons.unl.edu/scholcom/9/}}
}

@article{lee2025evolvingdeeperllmthinking,
      title={Evolving {D}eeper {LLM} {T}hinking}, 
      author={Kuang-Huei Lee and Ian Fischer and Yueh-Hua Wu and Dave Marwood and Shumeet Baluja and Dale Schuurmans and Xinyun Chen},
      year={2025},
      eprint={2501.09891},
      archivePrefix={arXiv},
      primaryClass={cs.AI},
      journal={Arxiv Preprint arXiv: 2501.09891},
      url={https://arxiv.org/abs/2501.09891}, 
}

@article{qwen2025qwen25technicalreport,
      title={Qwen2.5 Technical Report}, 
      author={An Yang and Baosong Yang and Beichen Zhang and Binyuan Hui and Bo Zheng and Bowen Yu and Chengyuan Li and Dayiheng Liu and Fei Huang and Haoran Wei and others},
      year={2025},
      eprint={2412.15115},
      journal={Arxiv preprint arXiv: 2412.15115} ,
      archivePrefix={arXiv},
      primaryClass={cs.CL},
      url={https://arxiv.org/abs/2412.15115}, 
}

@article{openai2024gpt4ocard,
      title={{GPT}-4o {S}ystem {C}ard}, 
      author={Aaron Hurst and Adam Lerer and Adam P. Goucher and Adam Perelman and Aditya Ramesh and Aidan Clark and AJ Ostrow and Akila Welihinda and Alan Hayes and Alec Radford and Aleksander Mądry and Alex Baker-Whitcomb and others},
      year={2024},
      journal={Arxiv preprint arXiv: 2410.21276},
      eprint={2410.21276},
      archivePrefix={arXiv},
      primaryClass={cs.CL},
      url={https://arxiv.org/abs/2410.21276}, 
}

@article{openai2025gptoss120bgptoss20bmodel,
      title={gpt-oss-120b \& gpt-oss-20b Model Card}, 
      author={Sandhini Agarwal and Lama Ahmad and Jason Ai and Sam Altman and Andy Applebaum and Edwin Arbus and Rahul K. Arora and Yu Bai and Bowen Baker and Haiming Bao and Boaz Barak and Ally Bennett and Tyler Bertao and Nivedita Brett and Eugene Brevdo and Greg Brockman and Sebastien Bubeck and Che Chang and Kai Chen and Mark Chen and others},
      year={2025},
      eprint={2508.10925},
      journal={Arxiv preprint arXiv: 2508.10925},
      archivePrefix={arXiv},
      primaryClass={cs.CL},
      url={https://arxiv.org/abs/2508.10925}, 
}

@article{abdin2024phi4technicalreport,
      title={Phi-4 Technical Report}, 
      author={Marah Abdin and Jyoti Aneja and Harkirat Behl and Sébastien Bubeck and Ronen Eldan and Suriya Gunasekar and Michael Harrison and Russell J. Hewett and Mojan Javaheripi and Piero Kauffmann and James R. Lee and others},
      year={2024},
      eprint={2412.08905},
    journal={ArXiv preprint arXiv: 2412.08905},
      archivePrefix={arXiv},
      primaryClass={cs.CL},
      url={https://arxiv.org/abs/2412.08905}, 
}

@article{deepseekai2024deepseekllmscalingopensource,
      title={Deep{S}eek {LLM}: {S}caling {O}pen-{S}ource {L}anguage {M}odels with {L}ongtermism}, 
      author={Xiao Bi and Deli Chen and Guanting Chen and Shanhuang Chen and Damai Dai and Chengqi Deng and Honghui Ding and Kai Dong and Qiushi Du and Zhe Fu and Huazuo Gao and Kaige Gao and Wenjun Gao and Ruiqi Ge and others},
      year={2024},
      eprint={2401.02954},
        journal={Arxiv preprint arXiv: 2401.02954},
      archivePrefix={arXiv},
      primaryClass={cs.CL},
      url={https://arxiv.org/abs/2401.02954}, 
}

@article{ai2025yiopenfoundationmodels,
      title={Yi: Open Foundation Models by 01.AI}, 
      author={Alex Young and Bei Chen and Chao Li and Chengen Huang and Ge Zhang and Guanwei Zhang and Guoyin Wang and Heng Li and Jiangcheng Zhu and Jianqun Chen and others},
      year={2025},
      eprint={2403.04652},
        journal={ArXiv preprint arXiv: 2403.04652},
      archivePrefix={arXiv},
      primaryClass={cs.CL},
      url={https://arxiv.org/abs/2403.04652}, 
}

@inproceedings{pichler-etal-2025-evaluating,
    title = "Evaluating {LLM}-Prompting for Sequence Labeling Tasks in Computational Literary Studies",
    author = "Pichler, Axel  and
      Pagel, Janis  and
      Reiter, Nils",
    editor = "Kazantseva, Anna  and
      Szpakowicz, Stan  and
      Degaetano-Ortlieb, Stefania  and
      Bizzoni, Yuri  and
      Pagel, Janis",
    booktitle = "Proceedings of the 9th Joint SIGHUM Workshop on Computational Linguistics for Cultural Heritage, Social Sciences, Humanities and Literature (LaTeCH-CLfL 2025)",
    month = may,
    year = "2025",
    address = "Albuquerque, New Mexico",
    publisher = "Association for Computational Linguistics",
    url = "https://aclanthology.org/2025.latechclfl-1.5/",
    doi = "10.18653/v1/2025.latechclfl-1.5",
    pages = "32--46",
    ISBN = "979-8-89176-241-1"
}

@inproceedings{lan-etal-2024-multi,
    title = "Multi-label Sequential Sentence Classification via Large Language Model",
    author = "Lan, Mengfei  and
      Zheng, Lecheng  and
      Ming, Shufan  and
      Kilicoglu, Halil",
    editor = "Al-Onaizan, Yaser  and
      Bansal, Mohit  and
      Chen, Yun-Nung",
    booktitle = "Findings of the Association for Computational Linguistics: EMNLP 2024",
    month = nov,
    year = "2024",
    address = "Miami, Florida, USA",
    publisher = "Association for Computational Linguistics",
    url = "https://aclanthology.org/2024.findings-emnlp.944/",
    doi = "10.18653/v1/2024.findings-emnlp.944",
    pages = "16086--16104"
}

@inproceedings{
wei2025ai,
title={Position: The {ML} Community Must Build an {AI}-Augmented Peer-Review Ecosystem},
author={Qiyao Wei and Samuel Holt and Jing Yang and Markus Wulfmeier and Mihaela van der Schaar},
booktitle={Proceedings of the Forty-third International Conference on Machine Learning Position Paper Track, ICML 2026, Seoul, South Korea, July 6-11, 2026},
year={2026},
url={https://openreview.net/forum?id=0eAsCehpyb},
publisher={OpenReview.net}
}

@inproceedings{kang2025template,
  title={Template-{D}riven {LLM}-{P}araphrased {F}ramework for {T}abular {M}ath {W}ord {P}roblem {G}eneration},
  author={Kang, Xiaoqiang and Wang, Zimu and Jin, Xiaobo and Wang, Wei and Huang, Kaizhu and Wang, Qiufeng},
  booktitle={Proceedings of the Thirty-Ninth AAAI Conference on Artificial Intelligence, February 29-March 4, 2025, Philadelphia, Pennsylvania, USA},
  pages={24303--24311},
  year={2025},
url={https://ojs.aaai.org/index.php/AAAI/article/view/34607},
  publisher={AAAI Press}
}

@inproceedings{huot-etal-2023-text,
    title = "Text-Blueprint: An Interactive Platform for Plan-based Conditional Generation",
    author = "Huot, Fantine  and
      Maynez, Joshua  and
      Narayan, Shashi  and
      Amplayo, Reinald Kim  and
      Ganchev, Kuzman  and
      Louis, Annie Priyadarshini  and
      Sandholm, Anders  and
      Das, Dipanjan  and
      Lapata, Mirella",
    editor = "Croce, Danilo  and
      Soldaini, Luca",
    booktitle = "Proceedings of the 17th Conference of the European Chapter of the Association for Computational Linguistics: System Demonstrations",
    month = may,
    year = "2023",
    address = "Dubrovnik, Croatia",
    publisher = "Association for Computational Linguistics",
    url = "https://aclanthology.org/2023.eacl-demo.13/",
    doi = "10.18653/v1/2023.eacl-demo.13",
    pages = "105--116"
}

@inproceedings{lu-etal-2025-identifying,
    title = "Identifying Aspects in Peer Reviews",
    author = "Lu, Sheng  and
      Kuznetsov, Ilia  and
      Gurevych, Iryna",
    editor = "Christodoulopoulos, Christos  and
      Chakraborty, Tanmoy  and
      Rose, Carolyn  and
      Peng, Violet",
    booktitle = "Findings of the Association for Computational Linguistics: EMNLP 2025",
    month = nov,
    year = "2025",
    address = "Suzhou, China",
    publisher = "Association for Computational Linguistics",
    url = "https://aclanthology.org/2025.findings-emnlp.326/",
    doi = "10.18653/v1/2025.findings-emnlp.326",
    pages = "6145--6167",
    ISBN = "979-8-89176-335-7"
}

@inproceedings{purkayastha-etal-2023-exploring,
    title = "Exploring Jiu-Jitsu Argumentation for Writing Peer Review Rebuttals",
    author = "Purkayastha, Sukannya  and
      Lauscher, Anne  and
      Gurevych, Iryna",
    editor = "Bouamor, Houda  and
      Pino, Juan  and
      Bali, Kalika",
    booktitle = "Proceedings of the 2023 Conference on Empirical Methods in Natural Language Processing",
    month = dec,
    year = "2023",
    address = "Singapore",
    publisher = "Association for Computational Linguistics",
    url = "https://aclanthology.org/2023.emnlp-main.894/",
    doi = "10.18653/v1/2023.emnlp-main.894",
    pages = "14479--14495"
}

@inproceedings{wolf-etal-2020-transformers,
    title = "Transformers: State-of-the-Art Natural Language Processing",
    author = "Wolf, Thomas  and
      Debut, Lysandre  and
      Sanh, Victor  and
      Chaumond, Julien  and
      others",
    editor = "Liu, Qun  and
      Schlangen, David",
    booktitle = "Proceedings of the 2020 Conference on Empirical Methods in Natural Language Processing: System Demonstrations",
    month = oct,
    year = "2020",
    address = "Online",
    publisher = "Association for Computational Linguistics",
    url = "https://aclanthology.org/2020.emnlp-demos.6/",
    doi = "10.18653/v1/2020.emnlp-demos.6",
    pages = "38--45"
}

@inproceedings{pham-etal-2025-surveypilot,
    title = "{S}urvey{P}ilot: an Agentic Framework for Automated Human Opinion Collection from Social Media",
    author = "Pham, Viet Thanh  and
      Qu, Lizhen  and
      Li, Zhuang  and
      Sharma, Suraj  and
      Haffari, Gholamreza",
    editor = "Che, Wanxiang  and
      Nabende, Joyce  and
      Shutova, Ekaterina  and
      Pilehvar, Mohammad Taher",
    booktitle = "Proceedings of the 63rd Annual Meeting of the Association for Computational Linguistics (Volume 1: Long Papers)",
    month = jul,
    year = "2025",
    address = "Vienna, Austria",
    publisher = "Association for Computational Linguistics",
    url = "https://aclanthology.org/2025.acl-long.221/",
    doi = "10.18653/v1/2025.acl-long.221",
    pages = "4397--4422",
    ISBN = "979-8-89176-251-0"
}

@article{epoch2025outputlength,
    title={{LLM} responses to benchmark questions are getting longer over time},
    author={Luke Emberson and Ben Cottier and Josh You and Tom Adamczewski and Jean-Stanislas Denain},
    year={2025},
journal={Blog Post},
    url={https://epoch.ai/data-insights/output-length}
  }

@article{bradley1952rank,
  title={Rank analysis of incomplete block designs: I. the method of paired comparisons},
  author={Bradley, Ralph Allan and Terry, Milton E},
  journal={Biometrika},
  volume={39},
  number={3/4},
  url={https://www.jstor.org/stable/2334029},
  pages={324--345},
  year={1952},
  publisher={JSTOR},
  note={\url{https://doi.org/10.1093/biomet/39.3-4.324}}
}

@article{taye2023understanding,
  title={Understanding of machine learning with deep learning: architectures, workflow, applications and future directions},
  url={https://www.mdpi.com/2073-431X/12/5/91},
  author={Taye, Mohammad Mustafa},
  journal={Computers},
  volume={12},
  number={5},
  pages={91},
  year={2023},
  publisher={Mdpi}
}

@article{
jeong2025llmselect,
title={{LLM}-Select: Feature Selection with Large Language Models},
author={Daniel P Jeong and Zachary Chase Lipton and Pradeep Kumar Ravikumar},
journal={Transactions on Machine Learning Research},
issn={2835-8856},
year={2025},
url={https://openreview.net/forum?id=16f7ea1N3p},
note={},
volume= "2025"
}

@inproceedings{manikandan2023language,
  title={Language models are weak learners},
  author={Manikandan, Hariharan and Jiang, Yiding and Kolter, J Zico},
  booktitle={Proceedings of the Thirty-seventh Conference on Neural Information Processing Systems, NeurIPS 2023, New Orleans, Louisiana, USA, Dec 10-16, 2023},
year={2023},
pages={},
url={https://openreview.net/forum?id=559NJBfN20},
publisher={Openreview.net}

}

@inproceedings{
madaan2023selfrefine,
title={Self-{R}efine: {I}terative {R}efinement with {S}elf-{F}eedback},
author={Aman Madaan and Niket Tandon and Prakhar Gupta and Skyler Hallinan and Luyu Gao and others},
booktitle={Proceedings of the Thirty-seventh Conference on Neural Information Processing Systems, NeurIPS 2023, New Orleans, Louisiana, USA, Dec 10-16, 2023},
year={2023},
pages={},
url={https://openreview.net/forum?id=S37hOerQLB},
publisher={Openreview.net}
}

@INPROCEEDINGS{yaacoub2025analyzingfeedbackmechanismsaigenerated,
  author={Yaacoub, Antoun and Assaghir, Zainab and Prevost, Lionel and Da-Rugna, Jérôme},
  booktitle={2025 9th International Conference on Computer, Software and Modeling (ICCSM)}, 
  title={Analyzing Feedback Mechanisms in Ai-Generated Mcqs: Insights Into Readability, Lexical Properties, and Levels of Challenge}, 
  year={2025},
  volume={},
  number={},
  pages={108-113},
  doi={10.1109/ICCSM66818.2025.00026}}

@article{brown2024largelanguagemonkeysscaling,
      title={Large Language Monkeys: Scaling Inference Compute with Repeated Sampling}, 
      author={Bradley Brown and Jordan Juravsky and Ryan Ehrlich and Ronald Clark and Quoc V. Le and Christopher Ré and Azalia Mirhoseini},
      year={2024},
      eprint={2407.21787},
    journal={Arxiv preprint arXiv: 2407.21787},
      archivePrefix={arXiv},
      primaryClass={cs.LG},
      url={https://arxiv.org/abs/2407.21787}, 
}

@inproceedings{azaiz2024feedback,
author = {Azaiz, Imen and Kiesler, Natalie and Strickroth, Sven},
title = {Feedback-{G}eneration for {P}rogramming {E}xercises With {GPT}-4},
year = {2024},
isbn = {9798400706004},
publisher = {Association for Computing Machinery},
address = {New York, NY, USA},
url = {https://doi.org/10.1145/3649217.3653594},
doi = {10.1145/3649217.3653594},
booktitle = {Proceedings of the 2024 on Innovation and Technology in Computer Science Education V. 1},
pages = {31–37},
numpages = {7},
location = {Milan, Italy},
series = {ITiCSE 2024}
}

@article{kasneci2023chatgpt,
  title={{ChatGPT} for good? {O}n opportunities and challenges of large language models for education},
  author={Kasneci, Enkelejda and Se{\ss}ler, Kathrin and K{\"u}chemann, Stefan and Bannert, Maria and Dementieva, Daryna and Fischer, Frank and Gasser, Urs and Groh, Georg and G{\"u}nnemann, Stephan and H{\"u}llermeier, Eyke and others},
  journal={Learning and individual differences},
  volume={103},
  pages={102274},
  year={2023},
url={https://www.sciencedirect.com/science/article/abs/pii/S1041608023000195},
  publisher={Elsevier},
  note={\url{https://doi.org/10.1016/j.lindif.2023.102274}}
}

@article{misiejuk2024augmenting,
  title={Augmenting assessment with {AI} coding of online student discourse: {A} question of reliability},
  journal = {Computers and Education: Artificial Intelligence},
volume = {6},
pages = {100216},
year = {2024},
issn = {2666-920X},
doi = {https://doi.org/10.1016/j.caeai.2024.100216},
url = {https://www.sciencedirect.com/science/article/pii/S2666920X24000171},
author = {Kamila Misiejuk and Rogers Kaliisa and Jennifer Scianna},
note={\url{https://doi.org/10.1016/j.caeai.2024.100216}}
}

@article{escalante2023ai,
  title={{AI}-generated feedback on writing: {I}nsights into efficacy and {ENL} student preference},
  author={Escalante, Juan and Pack, Austin and Barrett, Alex},
  journal={International Journal of Educational Technology in Higher Education},
  volume={20},
  number={1},
url={http://link.springer.com/article/10.1186/s41239-023-00425-2},
  pages={57},
  year={2023},
  publisher={Springer},
  note={\url{https://doi.org/10.1186/s41239-023-00425-2

}}
}

@article{bharti2024politepeer,
  title={Polite{PEER}: does peer review hurt? {A} dataset to gauge politeness intensity in the peer reviews},
  author={Bharti, Prabhat Kumar and Navlakha, Meith and Agarwal, Mayank and Ekbal, Asif},
  journal={Language Resources and Evaluation},
  volume={58},
  number={4},
  url={https://link.springer.com/article/10.1007/s10579-023-09662-3},
  pages={1291--1313},
  year={2024},
  publisher={Springer},
  note={\url{https://doi.org/10.1007/s10579-023-09662-3}}
}

@inproceedings{kennard-etal-2022-disapere,
    title = "{DISAPERE}: A Dataset for Discourse Structure in Peer Review Discussions",
    author = "Kennard, Neha Nayak  and
      O{'}Gorman, Tim  and
      Das, Rajarshi  and
      Sharma, Akshay  and
      Bagchi, Chhandak  and
      Clinton, Matthew  and
      Yelugam, Pranay Kumar  and
      Zamani, Hamed  and
      McCallum, Andrew",
    editor = "Carpuat, Marine  and
      de Marneffe, Marie-Catherine  and
      Meza Ruiz, Ivan Vladimir",
    booktitle = "Proceedings of the 2022 Conference of the North American Chapter of the Association for Computational Linguistics: Human Language Technologies",
    month = jul,
    year = "2022",
    address = "Seattle, United States",
    publisher = "Association for Computational Linguistics",
    url = "https://aclanthology.org/2022.naacl-main.89/",
    doi = "10.18653/v1/2022.naacl-main.89",
    pages = "1234--1249"
}

@article{feng-etal-2023-less,
    title = "Less is More: Mitigate Spurious Correlations for Open-Domain Dialogue Response Generation Models by Causal Discovery",
    author = "Feng, Tao  and
      Qu, Lizhen  and
      Haffari, Gholamreza",
    journal = "Transactions of the Association for Computational Linguistics",
    volume = "11",
    year = "2023",
    address = "Cambridge, MA",
    publisher = "MIT Press",
    url = "https://aclanthology.org/2023.tacl-1.30/",
    doi = "10.1162/tacl_a_00561",
    pages = "511--530"
}

@inproceedings{kim-etal-2024-prometheus,
    title = "Prometheus 2: An Open Source Language Model Specialized in Evaluating Other Language Models",
    author = "Kim, Seungone  and
      Suk, Juyoung  and
      Longpre, Shayne  and
      Lin, Bill Yuchen  and
      Shin, Jamin  and
      Welleck, Sean  and
      Neubig, Graham  and
      Lee, Moontae  and
      Lee, Kyungjae  and
      Seo, Minjoon",
    editor = "Al-Onaizan, Yaser  and
      Bansal, Mohit  and
      Chen, Yun-Nung",
    booktitle = "Proceedings of the 2024 Conference on Empirical Methods in Natural Language Processing",
    month = nov,
    year = "2024",
    address = "Miami, Florida, USA",
    publisher = "Association for Computational Linguistics",
    url = "https://aclanthology.org/2024.emnlp-main.248/",
    doi = "10.18653/v1/2024.emnlp-main.248",
    pages = "4334--4353"
}

@inproceedings{maurya-etal-2025-unifying,
    title = "Unifying {AI} Tutor Evaluation: An Evaluation Taxonomy for Pedagogical Ability Assessment of {LLM}-Powered {AI} Tutors",
    author = "Maurya, Kaushal Kumar  and
      Srivatsa, Kv Aditya  and
      Petukhova, Kseniia  and
      Kochmar, Ekaterina",
    editor = "Chiruzzo, Luis  and
      Ritter, Alan  and
      Wang, Lu",
    booktitle = "Proceedings of the 2025 Conference of the Nations of the Americas Chapter of the Association for Computational Linguistics: Human Language Technologies (Volume 1: Long Papers)",
    month = apr,
    year = "2025",
    address = "Albuquerque, New Mexico",
    publisher = "Association for Computational Linguistics",
    url = "https://aclanthology.org/2025.naacl-long.57/",
    doi = "10.18653/v1/2025.naacl-long.57",
    pages = "1234--1251",
    ISBN = "979-8-89176-189-6"
}

@article{sahnan2025llmsautomatefactcheckingarticle,
    title = "Can {LLM}s Automate Fact-Checking Article Writing?",
    author = "Sahnan, Dhruv  and
      Corney, David  and
      Larraz, Irene  and
      Zagni, Giovanni  and
      Miguez, Ruben  and
      Xie, Zhuohan  and
      Gurevych, Iryna  and
      Churchill, Elizabeth  and
      Chakraborty, Tanmoy  and
      Nakov, Preslav",
    journal = "Transactions of the Association for Computational Linguistics",
    volume = "14",
    year = "2026",
    address = "Cambridge, MA",
    publisher = "MIT Press",
    url = "https://aclanthology.org/2026.tacl-1.23/",
    doi = "10.1162/tacl.a.644",
    pages = "489--509"
}

@article{goldberg1990note,
  title={A note on Boltzmann tournament selection for genetic algorithms and population-oriented simulated annealing},
  author={Goldberg, David E},
  journal={Complex Systems},
  volume={4},
  pages={445--460},
  year={1990},
  publisher={Complex Systems Publications, Inc.},
  url={https://www.complex-systems.com/abstracts/v04_i04_a05/}
}

@article{flesch1948new,
  title={A new readability yardstick.},
  author={Flesch, Rudolph},
  journal={Journal of applied psychology},
  volume={32},
  number={3},
  pages={221-233},
  year={1948},
  url={https://psycnet.apa.org/record/1949-01274-001?doi=1},
  publisher={American Psychological Association},
  note={\url{https://doi.org/10.1037/h0057532}}
}

@article{hu2020makes, 
title={What {M}akes {A} {G}ood {S}tory? {D}esigning {C}omposite {R}ewards for {V}isual {S}torytelling}, 
volume={34}, 
url={https://ojs.aaai.org/index.php/AAAI/article/view/6305}, 
DOI={10.1609/aaai.v34i05.6305}, 
abstractNote={&lt;p&gt;Previous storytelling approaches mostly focused on optimizing traditional metrics such as BLEU, ROUGE and CIDEr. In this paper, we re-examine this problem from a different angle, by looking deep into what defines a natural and topically-coherent story. To this end, we propose three assessment criteria: &lt;em&gt;relevance&lt;/em&gt;, &lt;em&gt;coherence&lt;/em&gt; and &lt;em&gt;expressiveness&lt;/em&gt;, which we observe through empirical analysis could constitute a “high-quality” story to the human eye. We further propose a reinforcement learning framework, ReCo-RL, with reward functions designed to capture the essence of these quality criteria. Experiments on the Visual Storytelling Dataset (VIST) with both automatic and human evaluation demonstrate that our ReCo-RL model achieves better performance than state-of-the-art baselines on both traditional metrics and the proposed new criteria.&lt;/p&gt;}, 
number={05}, 
journal={Proceedings of the AAAI Conference on Artificial Intelligence}, 
author={Hu, Junjie and Cheng, Yu and Gan, Zhe and Liu, Jingjing and Gao, Jianfeng and Neubig, Graham}, 
year={2020}, 
month={Apr.}, 
pages={7969-7976} }

@article{narayan-etal-2023-conditional,
    title = "Conditional Generation with a Question-Answering Blueprint",
    author = "Narayan, Shashi  and
      Maynez, Joshua  and
      Amplayo, Reinald Kim  and
      Ganchev, Kuzman  and
      Louis, Annie  and
      Huot, Fantine  and
      Sandholm, Anders  and
      Das, Dipanjan  and
      Lapata, Mirella",
    journal = "Transactions of the Association for Computational Linguistics",
    volume = "11",
    year = "2023",
    address = "Cambridge, MA",
    publisher = "MIT Press",
    url = "https://aclanthology.org/2023.tacl-1.55/",
    doi = "10.1162/tacl_a_00583",
    pages = "974--996"
}

@inproceedings{xu2024jamplate,
  title={Jamplate: {E}xploring {LLM}-{E}nhanced {T}emplates for {I}dea {R}eflection},
  author = {Xu, Xiaotong (Tone) and Yin, Jiayu and Gu, Catherine and Mar, Jenny and Zhang, Sydney and E, Jane L. and Dow, Steven P.},
year = {2024},
isbn = {9798400705083},
publisher = {Association for Computing Machinery},
url = {https://doi.org/10.1145/3640543.3645196},
doi = {10.1145/3640543.3645196},
booktitle = {Proceedings of the 29th International Conference on Intelligent User Interfaces, IUI '24, March 18-21, 2024, Greenville, SC, USA},
pages={907–921},
numpages = {15},
location = {Greenville, SC, USA}
}

@article{krippendorff2004reliability,
  title={Reliability in content analysis: Some common misconceptions and recommendations},
  author={Krippendorff, Klaus},
  journal={Human communication research},
  volume={30},
  number={3},
  pages={411--433},
  year={2004},
  url={https://academic.oup.com/hcr/article-abstract/30/3/411/4331534},
  publisher={Wiley Online Library},
  note={\url{https://doi.org/10.1111/j.1468-2958.2004.tb00738.x}}
}

@inproceedings{sadallah2025goodbadconstructiveautomatically,
    title = "The Good, the Bad and the Constructive: Automatically Measuring Peer Review{'}s Utility for Authors",
    author = {Sadallah, Abdelrahman  and
      Baumg{\"a}rtner, Tim  and
      Gurevych, Iryna  and
      Briscoe, Ted},
    editor = "Christodoulopoulos, Christos  and
      Chakraborty, Tanmoy  and
      Rose, Carolyn  and
      Peng, Violet",
    booktitle = "Proceedings of the 2025 Conference on Empirical Methods in Natural Language Processing",
    month = nov,
    year = "2025",
    address = "Suzhou, China",
    publisher = "Association for Computational Linguistics",
    url = "https://aclanthology.org/2025.emnlp-main.1476/",
    doi = "10.18653/v1/2025.emnlp-main.1476",
    pages = "28979--29009",
    ISBN = "979-8-89176-332-6"
}

@Inbook{Ramshaw1999,
author="Ramshaw, L. A.
and Marcus, M. P.",
editor="Armstrong, Susan
and Church, Kenneth
and Isabelle, Pierre
and Manzi, Sandra
and Tzoukermann, Evelyne
and Yarowsky, David",
title="Text Chunking Using Transformation-Based Learning",
bookTitle="Natural Language Processing Using Very Large Corpora",
year="1999",
publisher="Springer Netherlands",
address="Dordrecht",
pages="157--176",
isbn="978-94-017-2390-9",
doi="10.1007/978-94-017-2390-9_10",
url="https://doi.org/10.1007/978-94-017-2390-9_10",
note={\url{https://doi.org/10.1007/978-94-017-2390-9_10}}
}

@inproceedings{dycke-etal-2022-yes,
    title = "Yes-Yes-Yes: Proactive Data Collection for {ACL} Rolling Review and Beyond",
    author = "Dycke, Nils  and
      Kuznetsov, Ilia  and
      Gurevych, Iryna",
    editor = "Goldberg, Yoav  and
      Kozareva, Zornitsa  and
      Zhang, Yue",
    booktitle = "Findings of the Association for Computational Linguistics: EMNLP 2022",
    month = dec,
    year = "2022",
    address = "Abu Dhabi, United Arab Emirates",
    publisher = "Association for Computational Linguistics",
    url = "https://aclanthology.org/2022.findings-emnlp.23/",
    doi = "10.18653/v1/2022.findings-emnlp.23",
    pages = "300--318"
}

@inproceedings{rogers-etal-2023-report,
    title = "Program Chairs' Report on Peer Review at ACL 2023",
    author = "Rogers, Anna  and
      Karpinska, Marzena  and
      Boyd-Graber, Jordan  and
      Okazaki, Naoaki",
    editor = "Rogers, Anna  and
      Boyd-Graber, Jordan  and
      Okazaki, Naoaki",
    booktitle = "Proceedings of the 61st Annual Meeting of the Association for Computational Linguistics (Volume 1: Long Papers)",
    month = jul,
    year = "2023",
    address = "Toronto, Canada",
    publisher = "Association for Computational Linguistics",
    url = "https://aclanthology.org/2023.acl-long.911/",
    pages = "xl-lxxv"
}

@article{aczel2021billion,
  title={A billion-dollar donation: estimating the cost of researchers’ time spent on peer review},
  author={Aczel, Balazs and Szaszi, Barnabas and Holcombe, Alex O},
  journal={Research integrity and peer review},
  volume={6},
  number={1},
  pages={1--8},
  year={2021},
  url={https://link.springer.com/article/10.1186/s41073-021-00118-2},
  publisher={Springer},
  note={\url{https://link.springer.com/article/10.1186/s41073-021-00118-2}}
}

@article{https://doi.org/10.1002/asi.22636,
author = {van Dalen, Hendrik P. and Henkens, Kène},
title = {Intended and unintended consequences of a publish-or-perish culture: A worldwide survey},
journal = {Journal of the American Society for Information Science and Technology},
volume = {63},
number = {7},
pages = {1282-1293},
doi = {https://doi.org/10.1002/asi.22636},
url = {https://asistdl.onlinelibrary.wiley.com/doi/abs/10.1002/asi.22636},
eprint = {https://asistdl.onlinelibrary.wiley.com/doi/pdf/10.1002/asi.22636},
year = {2012},
note={\url{https://doi.org/10.1002/asi.22636}}
}

@inproceedings{
liang2024mapping,
title={Mapping the Increasing Use of {LLM}s in Scientific Papers},
author={Weixin Liang and Yaohui Zhang and Zhengxuan Wu and Haley Lepp and Wenlong Ji and Xuandong Zhao and Hancheng Cao and Sheng Liu and Siyu He and Zhi Huang and Diyi Yang and Christopher Potts and Christopher D Manning and James Y. Zou},
booktitle={First Conference on Language Modeling, University of Pennsylvania, Philadelphia, PA, October 7-9, 2024},
year={2024},
url={https://openreview.net/forum?id=YX7QnhxESU},
publisher={Openreview.net}
}

@article{Publication_Ethics, 
title={{ACL} {P}olicy on {P}ublication {E}thics}, 
author      = {{ACL Publication Ethics Committee}},
  journal = {Association for Computational Linguistics},
  year        = {2024},
  url         = {https://www.aclweb.org/adminwiki/index.php/ACL_Policy_on_Publication_Ethics},
  publisher={aclweb.org}
}

@article{kuznetsov-etal-2022-revise,
    title = "Revise and Resubmit: An Intertextual Model of Text-based Collaboration in Peer Review",
    author = "Kuznetsov, Ilia  and
      Buchmann, Jan  and
      Eichler, Max  and
      Gurevych, Iryna",
    journal = "Computational Linguistics",
    volume = "48",
    number = "4",
    month = dec,
    year = "2022",
    address = "Cambridge, MA",
    publisher = "MIT Press",
    url = "https://aclanthology.org/2022.cl-4.16/",
    doi = "10.1162/coli_a_00455",
    pages = "949--986"
}

@article{10.1371/journal.pone.0320444,
    doi = {10.1371/journal.pone.0320444},
    author = {Goldberg, Alexander AND Stelmakh, Ivan AND Cho, Kyunghyun AND Oh, Alice AND Agarwal, Alekh AND Belgrave, Danielle AND Shah, Nihar B},
    journal = {PLOS ONE},
    publisher = {Public Library of Science},
    title = {Peer reviews of peer reviews: A randomized controlled trial and other experiments},
    year = {2025},
    month = {04},
    volume = {20},
    url = {https://doi.org/10.1371/journal.pone.0320444},
    pages = {1-18},
    number = {4},

}

@inproceedings{guo-etal-2023-automatic,
    title = "Automatic Analysis of Substantiation in Scientific Peer Reviews",
    author = "Guo, Yanzhu  and
      Shang, Guokan  and
      Rennard, Virgile  and
      Vazirgiannis, Michalis  and
      Clavel, Chlo{\'e}",
    editor = "Bouamor, Houda  and
      Pino, Juan  and
      Bali, Kalika",
    booktitle = "Findings of the Association for Computational Linguistics: EMNLP 2023",
    month = dec,
    year = "2023",
    address = "Singapore",
    publisher = "Association for Computational Linguistics",
    url = "https://aclanthology.org/2023.findings-emnlp.684/",
    doi = "10.18653/v1/2023.findings-emnlp.684",
    pages = "10198--10216"
}

@INPROCEEDINGS{9891914,
  author={Sevilla, Jaime and Heim, Lennart and Ho, Anson and Besiroglu, Tamay and Hobbhahn, Marius and Villalobos, Pablo},
  booktitle={2022 International Joint Conference on Neural Networks (IJCNN)}, 
  title={Compute Trends Across Three Eras of Machine Learning}, 
  year={2022},
  volume={},
  number={},
  pages={1-8},
  doi={10.1109/IJCNN55064.2022.9891914}}

@article{kuznetsov2024naturallanguageprocessingpeer,
      title={What {C}an {N}atural {L}anguage {P}rocessing {D}o for {P}eer {R}eview?}, 
      author={Ilia Kuznetsov and Osama Mohammed Afzal and Koen Dercksen and Nils Dycke and Alexander Goldberg and Tom Hope and Dirk Hovy and Jonathan K. Kummerfeld and Anne Lauscher and others},
      year={2024},
      eprint={2405.06563},
      archivePrefix={arXiv},
      primaryClass={cs.CL},
      journal={Arxiv preprint arXiv: 2405.06563},
      url={https://arxiv.org/abs/2405.06563}, 
}

@article{thakkar2025llmfeedbackenhancereview,
  title={A large-scale randomized study of large language model feedback in peer review},
  author={Thakkar, Nitya and Yuksekgonul, Mert and Silberg, Jake and Garg, Animesh and Peng, Nanyun and Sha, Fei and Yu, Rose and Vondrick, Carl and Zou, James},
  journal={Nature Machine Intelligence},
  volume={8},
  number={3},
  pages={326--336},
  url={https://www.nature.com/articles/s42256-026-01188-x},
  year={2026},
  publisher={Nature Publishing Group UK London}
}

@article{jansen2025constructive,
  title={Constructive feedback can function as a reward: Students’ emotional profiles in reaction to feedback perception mediate associations with task interest},
  author={Jansen, Thorben and H{\"o}ft, Lars and Bahr, J Luca and Kuklick, Livia and Meyer, Jennifer},
  journal={Learning and Instruction},
  volume={95},
  pages={102030},
  url={http://sciencedirect.com/science/article/pii/S0959475224001579},
  year={2025},
  publisher={Elsevier},
  note={\url{https://doi.org/10.1016/j.learninstruc.2024.102030}}
}

@inproceedings{dycke-etal-2023-nlpeer,
    title = "{NLP}eer: A Unified Resource for the Computational Study of Peer Review",
    author = "Dycke, Nils  and
      Kuznetsov, Ilia  and
      Gurevych, Iryna",
    editor = "Rogers, Anna  and
      Boyd-Graber, Jordan  and
      Okazaki, Naoaki",
    booktitle = "Proceedings of the 61st Annual Meeting of the Association for Computational Linguistics (Volume 1: Long Papers)",
    month = jul,
    year = "2023",
    address = "Toronto, Canada",
    publisher = "Association for Computational Linguistics",
    url = "https://aclanthology.org/2023.acl-long.277/",
    doi = "10.18653/v1/2023.acl-long.277",
    pages = "5049--5073"
}

@article{Rogers_Augenstein_2021, 
title={How to review for {ACL} {R}olling {R}eview}, 
url={https://aclrollingreview.org/reviewertutorial#6-check-for-lazy-thinking}, 
journal={ACL Rolling Review}, 
author={Rogers, Anna and Augenstein, Isabelle}, 
year={2024}, 
month={Nov}}

@inproceedings{purkayastha-etal-2025-lazyreview,
    title = "{L}azy{R}eview: A Dataset for Uncovering Lazy Thinking in {NLP} Peer Reviews",
    author = "Purkayastha, Sukannya  and
      Li, Zhuang  and
      Lauscher, Anne  and
      Qu, Lizhen  and
      Gurevych, Iryna",
    editor = "Che, Wanxiang  and
      Nabende, Joyce  and
      Shutova, Ekaterina  and
      Pilehvar, Mohammad Taher",
    booktitle = "Proceedings of the 63rd Annual Meeting of the Association for Computational Linguistics (Volume 1: Long Papers)",
    month = jul,
    year = "2025",
    address = "Vienna, Austria",
    publisher = "Association for Computational Linguistics",
    url = "https://aclanthology.org/2025.acl-long.165/",
    doi = "10.18653/v1/2025.acl-long.165",
    pages = "3280--3308",
    ISBN = "979-8-89176-251-0"
}

@article{liu2019robertarobustlyoptimizedbert,
      title={Ro{BERT}a: {A} {R}obustly {O}ptimized {BERT} {P}retraining {A}pproach}, 
      author={Yinhan Liu and Myle Ott and Naman Goyal and Jingfei Du and Mandar Joshi and Danqi Chen and Omer Levy and Mike Lewis and Luke Zettlemoyer and Veselin Stoyanov},
      year={2019},
      journal={Arxiv preprint arXiv: 1907.11692},
      eprint={1907.11692},
      archivePrefix={arXiv},
      primaryClass={cs.CL},
      url={https://arxiv.org/abs/1907.11692}, 
}

@inproceedings{du-etal-2024-llms,
    title = "{LLM}s Assist {NLP} Researchers: Critique Paper (Meta-)Reviewing",
    author = "Du, Jiangshu  and
      Wang, Yibo  and
      Zhao, Wenting  and
      Deng, Zhongfen  and
      Liu, Shuaiqi  and
      Lou, Renze  and
      Zou, Henry Peng  and
      Narayanan Venkit, Pranav  and
      Zhang, Nan  and
      Srinath, Mukund  and
      Zhang, Haoran Ranran  and
      Gupta, Vipul  and
      Li, Yinghui  and
      Li, Tao  and
      Wang, Fei  and
      Liu, Qin  and
      Liu, Tianlin  and
      Gao, Pengzhi  and
      Xia, Congying  and
      Xing, Chen  and
      Jiayang, Cheng  and
      Wang, Zhaowei  and
      Su, Ying  and
      Shah, Raj Sanjay  and
      Guo, Ruohao  and
      Gu, Jing  and
      Li, Haoran  and
      Wei, Kangda  and
      Wang, Zihao  and
      Cheng, Lu  and
      Ranathunga, Surangika  and
      Fang, Meng  and
      Fu, Jie  and
      Liu, Fei  and
      Huang, Ruihong  and
      Blanco, Eduardo  and
      Cao, Yixin  and
      Zhang, Rui  and
      Yu, Philip S.  and
      Yin, Wenpeng",
    editor = "Al-Onaizan, Yaser  and
      Bansal, Mohit  and
      Chen, Yun-Nung",
    booktitle = "Proceedings of the 2024 Conference on Empirical Methods in Natural Language Processing",
    month = nov,
    year = "2024",
    address = "Miami, Florida, USA",
    publisher = "Association for Computational Linguistics",
    url = "https://aclanthology.org/2024.emnlp-main.292/",
    doi = "10.18653/v1/2024.emnlp-main.292",
    pages = "5081--5099"
}

\appendix

\section{Appendix}

\subsection{Choice of LLMs and implementation details} \label{sec:choice}
Following~\citet{purkayastha-etal-2025-lazyreview}, we select models that are privacy-preserving in accordance with ARR publication ethics guidelines~\cite{Publication_Ethics} and of manageable sizes for practical deployment. All the LLMs are implemented using the vLLM library and huggingface transformers~\cite{wolf-etal-2020-transformers}.\footnote{\url{https://docs.vllm.ai/en/latest/}}. For calculating readability, we use the \texttt{textstat} package.\footnote{\url{https://pypi.org/project/textstat/}} The experiments are run on a single A100 GPU and none of the experiments consumed more than 36 hours.

\subsection{Annotation Guidelines} \label{sec:annotation}
\subsubsection{Task Description} 
As the volume of scientific publications grows, maintaining high-quality peer review becomes increasingly important. One common issue in reviews is the presence of non-specific reviews, which makes it difficult for authors and program chairs to address concerns effectively. Your task is to identify such non-specific issues in paper reviews and classify them into predefined categories.

Due to the high reviewing load, the reviewers are often prone to different kinds of bias, which inherently contribute to lower reviewing quality and hampers scientific progress in the field. The ACL peer reviewing guidelines characterize some of such reviewing bias in the table below, known as \textit{lazy thinking}. The table mentions the heuristic and the reason for the statement being problematic. In this task, we aim to annotate these biased sentences in peer-reviews from scientific conferences.

\subsubsection{Typical \textit{Lazy Thinking} and \textit{Specificity Issues}}
Some specificity and \textit{lazy thinking} issues are shown in Table~\ref{tab:issues}. The full \textit{specificity} issues table is in Table~\ref{tab:issues_rewrite} and \textit{lazy thinking} in Table~\ref{tab:heuristics_clean} respectively.
\begin{table}[!t]
    \centering
    \resizebox{!}{0.22\textwidth}{\begin{tabularx}{0.5\textwidth}{p{0.6cm}p{2cm}X}
    \toprule
    & \textbf{Issue} & \textbf{Meaning / Phrasing} \\ \midrule
    
    \multirow{2}{*}{\makebox[0.7cm][c]{\rotatebox{90}{\textit{Lazy Thinking}}}} 
        & Results are not novel & If the paper claims e.g., a novel method, and you think you've seen this before, you need to provide a reference. \\ \cmidrule(lr){2-3}
        & The topic is too niche & A main track paper may well make a big contribution to a narrow subfield. \\ \midrule
    
   \multirow{2}{*}{\makebox[0.7cm][c]{\rotatebox{90}{\textit{Specificity}}}} 
        & The contribution is not novel & Highly similar work X and Y has been published 3+ months prior to the submission deadline. \\ \cmidrule(lr){2-3}
        & X is not clear & Y and Z are missing from the description of X. \\ 
    
    \bottomrule
    \end{tabularx}}
    \caption{Examples of peer review issues categorized as \textit{lazy thinking} and \textit{specificity} as per ACL ARR guidelines~\cite{Rogers_Augenstein_2021}.}
    \label{tab:issues}
\end{table}

\subsubsection{Outlining differences within Summary of Weaknesses and Comments, Suggestions, and Typos}

\textbf{Summary of Weaknesses:} This field is often naively interpreted as the reasons to reject. ARR review guidelines provide the following definition:

\begin{quote}
What are the concerns that you have about the paper that would cause you to favor prioritizing other high-quality papers that are also under consideration for publication? These could include concerns about correctness of the results or argumentation, limited perceived impact of the methods or findings (note that impact can be significant both in broad or in narrow sub-fields), lack of clarity in exposition, or any other reason why interested readers of \textit{ACL} papers may gain less from this paper than they would from other papers under consideration. Where possible, please number your concerns so authors may respond to them individually.
\end{quote}

This field should contain reasons that make the paper not ready for publication at the current stage.

\medskip

\textbf{Comments, Suggestions, and Typos:} These are additional suggestions that reviewers can provide to authors to improve the manuscript. ARR review guidelines define this field as:

\begin{quote}
If you have any comments to the authors about how they may improve their paper, other than addressing the concerns above (weakness), please list them here.
\end{quote}

\medskip

The major difference between these two fields is the \textbf{severity of the issue}.

%\subsection{Annotation Format}
\subsubsection{Annotation Instructions for Review Sentences}

\begin{table}[!htbp]
\centering
\resizebox{!}{0.18\textwidth}{%
\begin{tabular}{l l l l l}  % 5 columns: Method Type, Method, Precision, Recall, F1
\toprule
\textbf{Method} & \textbf{Model} & \textbf{Prec.} & \textbf{Rec.} & \textbf{F1} \\
\midrule
\multirow{5}{*}{Sequential} & Yi  & 0.72  & 0.71  & 0.72 \\ \cmidrule(lr){2-5}
                            & Phi  & 0.73  & 0.72  & 0.72 \\ \cmidrule(lr){2-5}
                            & Qwen  & 0.67  & 0.61  & 0.64 \\ \cmidrule(lr){2-5}
                            & Deep.  & 0.62  & 0.62  & 0.62 \\ \cmidrule(lr){2-5}
                            & Oss. & 0.71 & 0.68 & 0.69 \\
\specialrule{2.5pt}{1pt}{1pt}
\multirow{5}{*}{Standalone}  & Yi  & 0.78  & 0.76  & 0.77 \\ \cmidrule(lr){2-5}
                            & Phi  & \textbf{0.81}  & \textbf{0.77}  & \textbf{0.79} \\ \cmidrule(lr){2-5}
                            & Qwen  & 0.73  & 0.67 & 0.70 \\ \cmidrule(lr){2-5}
                            & Deep.  & 0.72  & 0.66  & 0.69 \\ \cmidrule(lr){2-5}
                            & Oss. & 0.75  & 0.73  & 0.73 \\
\bottomrule
\end{tabular}}
\caption{Performance comparison of various models across different methods on the segment detection task, evaluated in terms of Precision (Prec.), Recall (Rec.), and F1.}
\label{tab:segment_iden_full}
\end{table}
To identify review segments, we first determine the boundary of each sentence using BIO tags. Given a review sentence, we classify it as the Beginning (B), Inside (I) of a review segment, or Other (O). We consider the preceding sentence: if it relates to a different topic, the current sentence is marked as B; if it continues the previous sentence, it is marked as I; and if it is not relevant (e.g., punctuation or non-substantial content), it is marked as O.

Next, we identify the section of the review sentence. Each sentence is categorized as belonging to the Summary of Weaknesses, Comments and Suggestions, or Summary of Strengths. According to ARR guidelines, major criticisms that influence acceptance decisions are placed in the Summary of Weaknesses, while other suggestions or minor criticisms belong in Comments and Suggestions.

Finally, we identify \textit{lazy thinking} or non-\textit{specific} writing using multi-label classification. For each sentence, we assign appropriate categories of \textit{lazy thinking} (e.g., ``the results are not novel'') and non-specificity (e.g., ``X was done in the way Y''). The entire segment, formed by all sentences tagged with B/I together, is considered before assigning labels. For example, the sentences ``I do not understand the setup'' (B) and ``There is no mention of epochs or hyperparameters in the dataset'' (I) together form a segment: 

\begin{quote}
``I do not understand the setup. There is no mention of epochs or hyperparameters in the dataset.''
\end{quote}

The segment collectively contributes to the label, e.g., \textbf{None}.
\begin{table*}[t]
    \centering
    \small{
    \begin{tabularx}{0.8\textwidth}{XXp{2.5cm}}%
    \toprule
    %\rowcolor{gray}
    \textbf{Review Segment} & \textbf{\textit{Lazy Thinking}} & \textbf{\textit{Specificity}} \\ \midrule
    More experiments on datasets are needed, and I would also like to see some other LLMs, like the Gemini family, also be evaluated to make the paper more comprehensive and generalizable. & Authors could also do extra experiment; Authors
should compare to a  closed model & N/A \\ \midrule 
Although the paper is interesting and strongly supported by experimental results, understanding the writing is tough. & The paper has language errors & X is not clear \\ \midrule 
Results in Table 1 are lower than SOTA results, casting doubt on the usefulness and novelty of this approach. & Results do not surpass the latest SOTA; Results are not novel & The contribution is not novel \\ \midrule 
    \end{tabularx}}
    \caption{Instances from our dataset, \textsc{LazyReviewPlus} where each segment may have multiple \textit{lazy thinking} and \textit{specificity} issues as per the ACL ARR guidelines~\cite{Rogers_Augenstein_2021} }
    \label{tab:new_dataset}
\end{table*}

\subsection{Analysis of the dataset, \textsc{LazyReviewPlus}}
\begin{figure*}
    \centering
    % Row 1
    \begin{subfigure}[b]{0.45\textwidth}
        \centering
        \includegraphics[width=\textwidth]{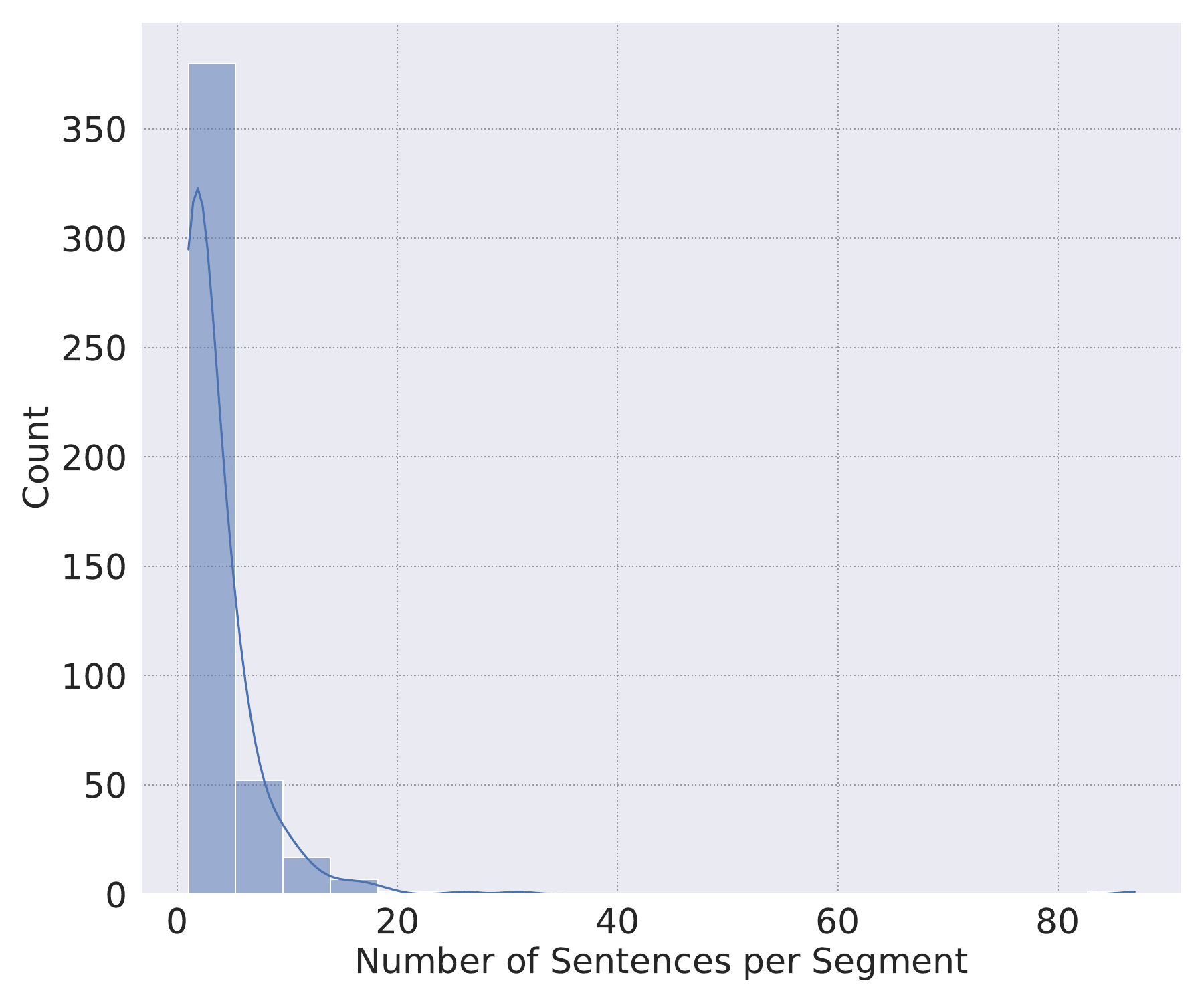}
        \caption{Sentences within a segment}
        \label{fig:seg_length_distribution}
    \end{subfigure}
    \hfill
    \begin{subfigure}[b]{0.45\textwidth}
        \centering
        \includegraphics[width=\textwidth]{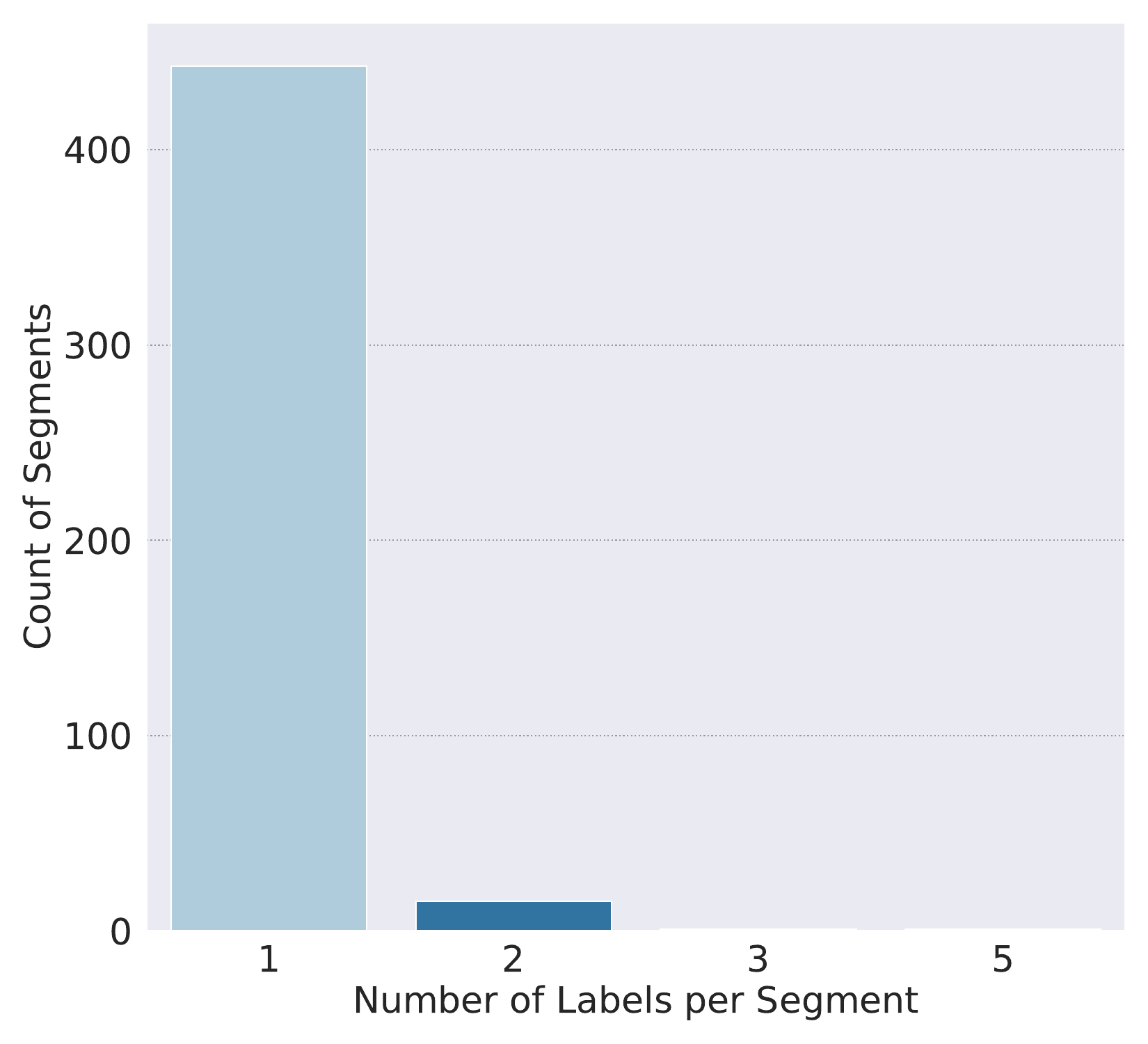}
        \caption{Distribution of labels within segments}
        \label{fig:label_to_segment}
    \end{subfigure}

    \caption{Overview of sentence and label distributions in our proposed dataset, \textsc{LazyReviewPlus}. }
\end{figure*}

We show instances from our dataset in Table~\ref{tab:new_dataset}. We show the distribution of sentences per segment in Fig~\ref{fig:seg_length_distribution}.The distribution of labels within each segment in Fig~\ref{fig:label_to_segment}. The distribution of \textit{specificity} and \textit{lazy thinking} issues in Figures \ref{fig:specificity_iss} and \ref{fig:lazy} respectively.

\begin{table*}[h!]
\centering
\resizebox{!}{0.2\textwidth}{\begin{tabular}{p{6cm} p{8cm}}
\hline
\textbf{Non-Specific Issues } & \textbf{Specific Rewrite } \\
\hline
The paper is missing relevant references & The paper is missing references XYZ \\
X is not clear & Y and Z are missing from the description of X. \\
The formulation of X is wrong & The formulation of X misses the factor Y \\
The contribution is not novel & Highly similar work X and Y has been published 3+ months prior to submission deadline \\
The paper is missing recent baselines & The proposed method should be compared against recent methods X, Y and Z (see H14 below for requesting comparisons to 'closed' systems) \\
X was done in the way Y & X was done in the way Y which has the disadvantage Z \\
The algorithm's interaction with dataset is problematic & It's possible that when using the decoding (line 723) on the dataset 3 (line 512), there might not be enough training data to rely on the n-best list.[If reasonably well-known entities are discussed] \\
\hline
\end{tabular}}
\caption{Examples of \textit{specificity} issues and their corresponding specific rewrites.}
\label{tab:issues_rewrite}
\end{table*}

\begin{table*}[h!]
\centering
\resizebox{!}{0.7\textwidth}{\begin{tabular}{p{3cm} p{10cm}}
\hline
\textbf{Heuristic} & \textbf{Why this is problematic} \\
\hline
H1. The results are not surprising & Many findings seem obvious in retrospect, but this does not mean that the community is already aware of them and can use them as building blocks for future work. Some findings may seem intuitive but haven’t previously been tested empirically. \\ \hline
H2. The results contradict what I would expect & You may be a victim of confirmation bias, and be unwilling to accept data contradicting your prior beliefs. \\ \hline
H3. The results are not novel & If the paper claims e.g. a novel method, and you think you've seen this before - you need to provide a reference (note the policy on what counts as concurrent work). If you don't think that the paper is novel due to its contribution type (e.g. reproduction, reimplementation, analysis) — please note that they are in scope of the CFP and deserve a fair hearing. \\ \hline
H4. This has no precedent in the existing literature & Papers that are more novel tend to be harder to publish. Reviewers may be unnecessarily conservative. \\ \hline
H5. The results do not surpass the latest SOTA & SOTA results are neither necessary nor sufficient for a scientific contribution. An engineering paper could also offer improvements on other dimensions (efficiency, generalizability, interpretability, fairness, etc.) If the authors do not claim that their contribution achieves SOTA status, the lack thereof is not an issue. \\ \hline
H6. The results are negative & The bias towards publishing only positive results is a known problem in many fields, and contributes to hype and overclaiming. If something systematically does not work where it could be expected, the community does need to know about it. \\ \hline
H7. This method is too simple & The goal is to solve the problem, not to solve it in a complex way. Simpler solutions are preferable, as they are less brittle and easier to deploy in real-world settings. \\ \hline
H8. The paper doesn't use [my preferred methodology] & NLP is an interdisciplinary field, relying on many kinds of contributions: models, resource, survey, data/linguistic/social analysis, position, and theory. \\ \hline
H9. The topic is too niche & A main track paper may well make a big contribution to a narrow subfield. \\ \hline
H10. The approach is tested only on [not English] & The same is true of NLP research that tests only on English. Monolingual work on any language is important practically (methods/resources) and theoretically (deeper understanding of language). \\ \hline
H11. The paper has language errors & As long as the writing is clear enough, better scientific content should be more valuable than better journalistic skills. \\ \hline
H12. The paper is missing the [reference X] & Per ACL policy, missing references to prior highly relevant work is a problem if such work was published 3+ months before the submission deadline. Otherwise, missing references belong in the "suggestions" section. \\ \hline
H13. The authors could also do [extra experiment X] & It is always possible to come up with extra experiments and follow-up work. But a paper only needs to present sufficient evidence for the claim that the authors are making. Extra experiments belong in the ``nice-to-have" category rather than ``reasons to reject." \\ \hline
H14. The authors should compare to a 'closed' model X & Requesting comparisons to closed-source models is only reasonable if it directly bears on the claim the authors are making. Many important questions require more openness than closed models allow. \\ \hline
H15. The authors should have done [X] instead & “I would have written this paper differently.” This criticism applies only if the authors’ choices prevent them from answering their research question or their framing is misleading. Otherwise, it is just a suggestion, not a weakness. \\ \hline
H16. Limitations != weaknesses & No paper is perfect, and most CL venues now require a Limitations section. A good review should not list limitations as reasons to reject unless appropriately motivated. \\ \hline
\end{tabular}}
\caption{\textit{Lazy Thinking} issues as per ARR Reviewer guidelines~\cite{Rogers_Augenstein_2021}.}
\label{tab:heuristics_clean}
\end{table*}
\clearpage

\begin{table*}[t]
\centering
\small
\resizebox{!}{0.6\textwidth}{\begin{tabular}{|p{5cm}|p{15cm}|}
\hline
\textbf{Issue / Heuristic} & \textbf{Template} \\
\hline
H1. The results are not surprising & Your comment --- ``[insert reviewer comment here]'' --- suggests that the result is not surprising. While this may reflect your expertise, we encourage you to provide evidence or references (e.g., [references similar to title of the paper]) if this outcome is already known. What feels intuitive may not have been previously established, and empirical confirmation can still be a valuable contribution. \\
\hline
H2. The results contradict what I would expect & Your comment --- ``[insert reviewer comment here]'' --- indicates that the results contradict your expectations. Please consider that this may reflect confirmation bias, where prior beliefs influence interpretation. It’s important to evaluate empirical findings objectively, even if they challenge existing assumptions. We encourage you to focus on the evidence presented and discuss how these findings advance understanding, rather than dismissing them based on expectation. \\
\hline
H3. The results are not novel & Your comment --- ``[insert reviewer comment here]'' --- raises concerns about novelty. If you believe the work duplicates prior work, please provide specific references ( [references similar to the paper]) to support this claim (keeping in mind policies on concurrent work). If the paper’s contribution is a reproduction, reimplementation, or analysis, please note that these are within the scope of the call and deserve fair consideration. Clear justification will help ensure the review is constructive and fair. \\
\hline
H4. This has no precedent in the existing literature & Your comment --- ``[insert reviewer comment here]'' --- indicates that the work has no clear precedent. While novelty is valuable, please be aware that highly novel contributions can be challenging to publish and may initially seem unfamiliar. Reviewers are encouraged to carefully consider the potential impact of innovative ideas rather than dismissing them due to lack of precedent. Providing constructive suggestions on how the authors can better frame or validate their novel contribution would be helpful. \\
\hline
H5. The results do not surpass the latest SOTA & Your comment --- ``[insert reviewer comment here]'' --- highlights that the results do not surpass the latest state-of-the-art (SOTA). Please keep in mind that achieving SOTA is neither necessary nor sufficient for a scientific contribution. Contributions improving other important aspects like efficiency, generalizability, interpretability, or fairness are also valuable. If the authors do not claim SOTA status, the lack of it should not be considered a weakness. \\
\hline
H6. The results are negative & Your comment --- ``[insert reviewer comment here]'' --- suggests the results are negative. Please consider that the bias toward publishing only positive results is a known issue across many fields. Negative or null results that show where methods systematically do not work are important for the community to understand limitations and avoid hype or overclaiming. Acknowledging the value of such findings can help make your review more balanced and constructive. \\
\hline
H7. This method is too simple & Your comment --- ``[insert reviewer comment here]'' --- suggests the method is too simple. However, the goal is to solve the problem effectively, not to add unnecessary complexity. Simpler methods often have advantages such as greater robustness, easier deployment, and better interpretability. Encouraging recognition of these benefits will help make your review more balanced and useful. \\
\hline
H8. The paper doesn't use [my preferred methodology] & Your comment --- ``[insert reviewer comment here]'' --- suggests a preference for a specific methodology (e.g., deep learning). Please note that NLP is an interdisciplinary field that values diverse contributions, including models, resources, surveys, data/linguistic/social analyses, theoretical work, and more. Limiting evaluation only to one methodology may overlook important advances. Recognizing this diversity will help make your review more comprehensive and fair. \\
\hline
H9. The topic is too niche & Your comment --- ``[insert reviewer comment here]'' --- suggests the topic is too niche. While the focus may be narrow, contributions to specific subfields can have significant impact and serve as important building blocks for the broader community. Recognizing the value of focused work helps ensure fair and balanced reviews. \\
\hline
H10. The approach is tested only on [not English] & Your comment --- ``[insert reviewer comment here]'' --- notes that the approach is tested only on [not English]. Please consider that monolingual research on any language is important both practically (developing methods and resources for that language) and theoretically (contributing to broader linguistic understanding). The NLP community values such focused contributions. Recognizing this helps ensure fair evaluation of research beyond English. \\
\hline
H11. The paper has language errors & Your comment --- ``[insert reviewer comment here]'' --- points out language errors. While clear writing is important, please prioritize the scientific content and contributions over minor language issues. Many papers can be improved with editorial help, but this should not overshadow the paper’s value. Focusing on the core scientific merits will make your review more balanced and constructive. \\
\hline
H12. The paper is missing the [reference X] & Your comment --- ``[insert reviewer comment here]'' --- points out missing references. Please note ACL policy requires missing references to published highly relevant prior work that appeared at least three months before the submission deadline to be addressed seriously. Missing references to unpublished or very recent preprints should be treated as suggestions rather than critical flaws. Clarifying this distinction will help maintain fair and policy-compliant reviews. \\
\hline
H13. The authors could also do [extra experiment X] & Your comment --- ``[insert reviewer comment here]'' --- suggests additional experiments. While extra experiments can always be proposed, a paper only needs to provide sufficient evidence to support its claims. Additional experiments are generally considered “nice-to-have” and fit better in the suggestions section rather than as reasons for rejection. If you believe an extra experiment is essential for validity, please clearly justify why in your review. \\
\hline
H14. The authors should compare to a 'closed' model X & Your comment --- ``[insert reviewer comment here]'' --- requests comparisons to closed-source models (e.g., ChatGPT). Please consider that such comparisons are only meaningful if they directly support the paper’s claims. Issues like test contamination and lack of transparency often limit their usefulness. Scientific progress often relies on openness and reproducibility, which closed models do not readily provide. Encouraging openness helps maintain rigorous evaluation standards. \\
\hline
H15. The authors should have done [X] instead & Your comment --- ``[insert reviewer comment here]'' --- suggests the authors should have done [X] instead. While multiple valid approaches often exist, this criticism is only warranted if the authors’ choices prevent them from answering their research question, if their framing is misleading, or if the question itself lacks merit. Otherwise, such remarks are best framed as suggestions rather than weaknesses. Clarifying this distinction helps make your review fairer and more constructive. \\
\hline
H16. Limitations != weaknesses & Your comment --- ``[insert reviewer comment here]'' --- treats the paper’s acknowledged limitations as weaknesses. While it’s important to recognize limitations, they do not by themselves justify rejection. Most CL venues expect a Limitations section to promote transparency. If you believe a limitation seriously undermines the work, please clearly explain why. Distinguishing limitations from fatal flaws will improve the fairness and usefulness of your review. \\
\hline
The paper is missing relevant references & Your comment --- ``[insert reviewer comment here]'' --- needs to be rewritten by mentioning the exact references the authors are missing. \\
\hline
X is not clear & Your comment --- ``[insert reviewer comment here]'' --- needs to be rewritten by specifying the details that are exactly missing such as Y and Z are missing from the description of X. \\
\hline
The formulation of X is wrong & Your comment --- ``[insert reviewer comment here]'' --- needs to be rephrased as the formulation of X misses the component Y. \\
\hline
The contribution is not novel & Your comment --- ``[insert reviewer comment here]'' --- needs to be rephrased as highly similar work X and Y has been published 3+ months prior to submission deadline. \\
\hline
The paper is missing recent baselines & Your comment --- ``[insert reviewer comment here]'' --- needs to be rephrased as the proposed method should be compared against recent methods X, Y and Z. \\
\hline
X was done in the way Y & Your comment --- ``[insert reviewer comment here]'' --- needs to be rephrased as X was done in the way Y which has the disadvantage Z. \\
\hline
\end{tabular}}
\caption{Templates corresponding to heuristics and common reviewer issue types.}
\label{tab:templates}
\end{table*} 

\clearpage

\begin{figure*}[t]
\begin{tcolorbox}[
colback=gray!5,
colframe=black,
title={Standalone Segment Detection Prompt},
boxrule=0.5mm,
arc=2mm,
fontupper=\small
]

\textbf{Instruction:}

You are a Segment Tagger. Your task is to tag the sentence with one of these labels in the context of the review. A segment consists of multiple sentences. The first sentence should be tagged with ``B'' and the remaining sentences with ``I''. Carefully examine where the sentence appears in the review before assigning a label.

\vspace{0.5em}

\textbf{Labels:}

B: Beginning of a segment.\\
I: Continuation of a segment. Always follows a B.
O: Outside any segment

\vspace{0.5em}

\textbf{Review:} \texttt{\{review\}}\\

\textbf{Sentence:} \texttt{\{sentence\}}

\vspace{0.5em}

\textbf{Output:}

Answer with \emph{only one label}: B, I or O.

\end{tcolorbox}
\caption{Prompt used for standalone segment detection.}
\label{fig:standalone_prompt}
\end{figure*}

\begin{figure*}[t]
\begin{tcolorbox}[
colback=gray!5,
colframe=black,
title={Sequential Segment Detection Prompt},
boxrule=0.5mm,
arc=2mm,
fontupper=\small
]

\textbf{Instruction:}

You are a Segment Tagger. Your task is to tag the sentence with one of these labels in the context of the review. A segment consists of multiple sentences. The first sentence should be tagged with ``B'' and the remaining sentences with ``I''. Carefully examine where the sentence appears in the review before assigning a label.

\vspace{0.5em}

\textbf{Labels:}

B: Beginning of a segment.\\
I: Continuation of a segment. Always follows a B.\\
O: Outside any segment.

\vspace{0.5em}

\textbf{Review:} \texttt{\{review\}}\\

\textbf{Previous sentence tag:} \texttt{\{previous\_prediction\}}\\

\textbf{Sentence:} \texttt{\{sentence\}}

\vspace{0.5em}

\textbf{Output:}

Answer with \emph{only one label}: B, I, or O.

\end{tcolorbox}
\caption{Prompt used for sequential segment detection.}
\label{fig:sequential_prompt}
\end{figure*}

\subsection{Segment Identification} \label{sec:segment_id_full}
%For a review $R = (s_1, \dots, s_n)$, each sentence $s_i$ is assigned a tag $y_i \in \{B, I, O\}$, where $B$ marks the beginning of a segment, $I$ a sentence inside a segment, and $O$ a sentence outside any segment. Let $f(s_i) = y_i$ be the tagging function; then the full review is tagged as $Y = (y_1, \dots, y_n) = f(R)$, with $y_i = B$ if $s_i$ starts a segment, $y_i = I$ if it continues a segment, and $y_i = O$ otherwise.\vspace{-2mm}

\noindent \textbf{Approach}
We perform segment detection using two approaches: \textbf{1. Standalone}: For each sentence $s_i$ in review $R$, predict $\hat{y}i = f\theta(s_i, R)$ to indicate beginning ($B$), inside ($I$), or outside ($O$) a segment. This relies solely on $s_i$ contextualized within the full review. \textbf{2. Sequential}: To capture inter-sentence dependencies, the LLM conditions on prior predictions $\hat{Y}{<i} = (\hat{y}1, \dots, \hat{y}{i-1})$, predicting $\hat{y}i = f\theta(s_i, R, \hat{Y}{<i})$. This provides temporary memory of past tagging decisions.

\noindent \textbf{Prompts.} The prompt used for standalone and sequential segment identification is provided in Fig~\ref{fig:standalone_prompt} and Fig~\ref{fig:sequential_prompt} respectively.

\noindent \textbf{Overall Results.} We present the performance of all models across different approaches in Table~\ref{tab:segment_iden_full}. The \textbf{standalone} strategy, which contextualizes each review segment within the full review, emerges as the most effective. In contrast, the \textbf{sequential} strategy suffers from error propagation, as reliance on prior predictions leads to cascading misclassifications, and sporadically may also introduce spurious correlations, consistent with prior findings~\cite{feng-etal-2023-less, purkayastha-etal-2025-lazyreview}. Among the models, Phi consistently outperforms the others, which can be attributed to its specialized training on textbook-quality corpora combined with a balanced mix of synthetic and real-world data~\cite{abdin2024phi4technicalreport}.

\begin{figure*}[h!]
    \centering
    % Row 1: 3 images side by side
    \begin{subfigure}{0.32\textwidth}
        \centering
        \includegraphics[width=\linewidth]{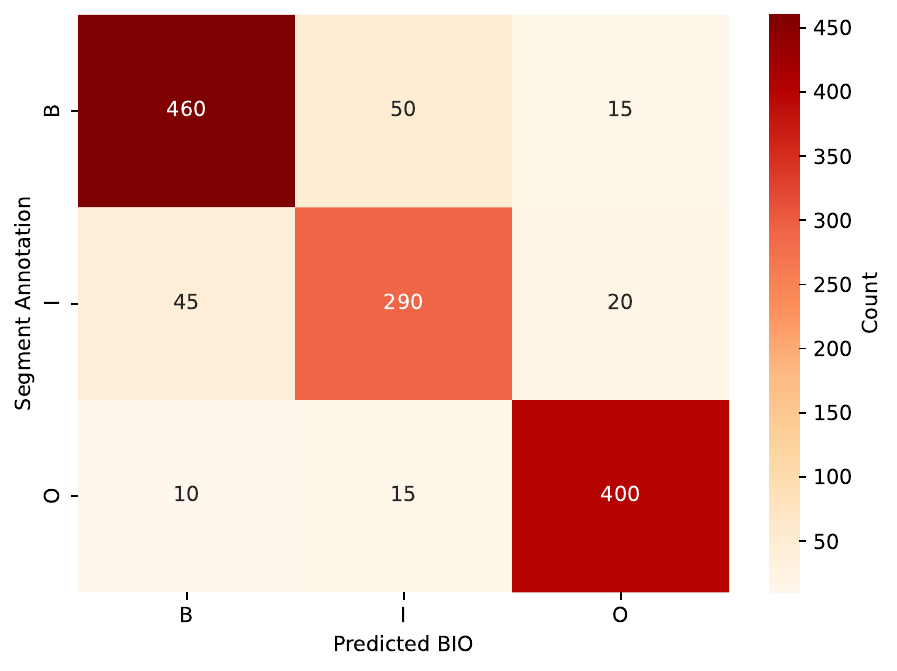}
        \caption{Yi}
    \end{subfigure}%
    \begin{subfigure}{0.32\textwidth}
        \centering
        \includegraphics[width=\linewidth]{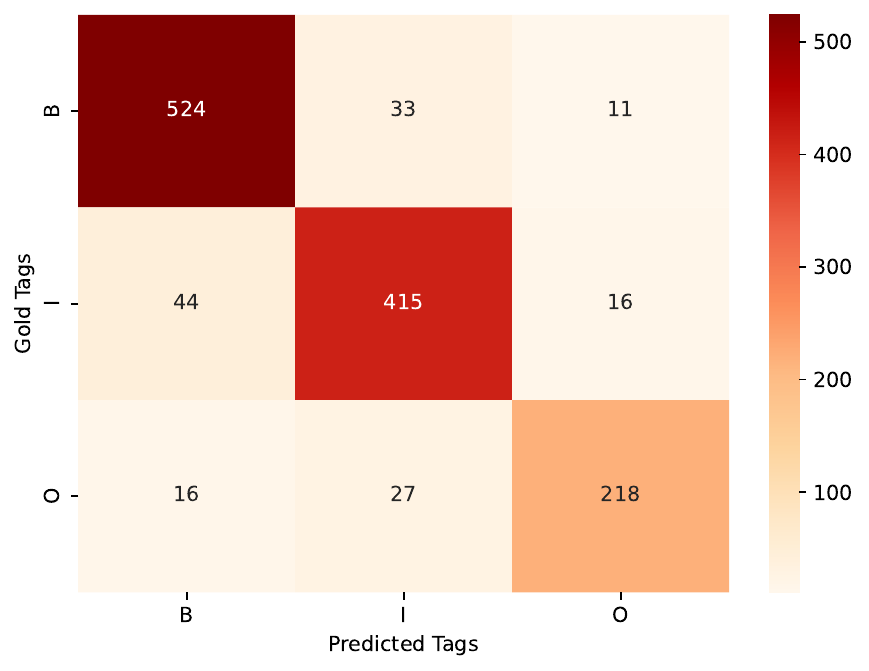}
        \caption{Phi}
    \end{subfigure}%
    \begin{subfigure}{0.32\textwidth}
        \centering
        \includegraphics[width=\linewidth]{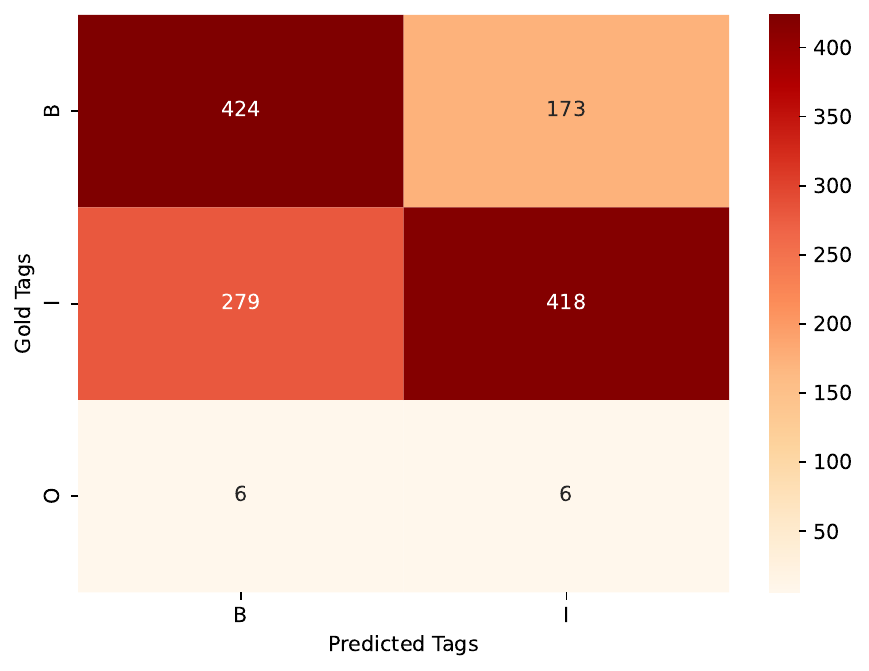}
        \caption{Qwen}
    \end{subfigure}\\[1em]
    
    % Row 2: 2 images side by side (centered)
    \hspace{5em}
    \begin{subfigure}{0.35\textwidth}
        \centering
        \includegraphics[width=\linewidth]{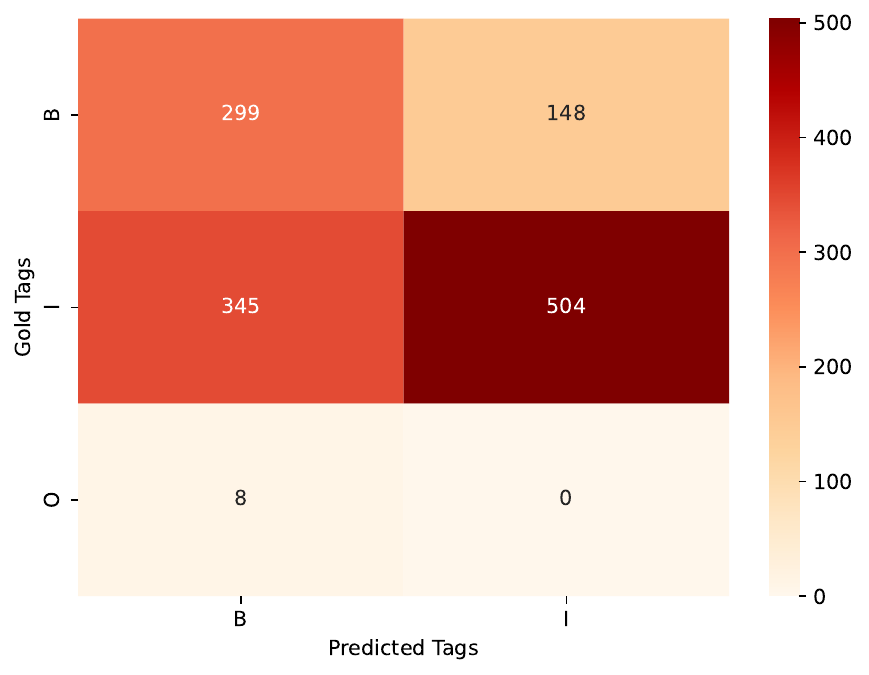}
        \caption{Deepseek}
    \end{subfigure}% 
    \begin{subfigure}{0.35\textwidth}
        \centering
        \includegraphics[width=\linewidth]{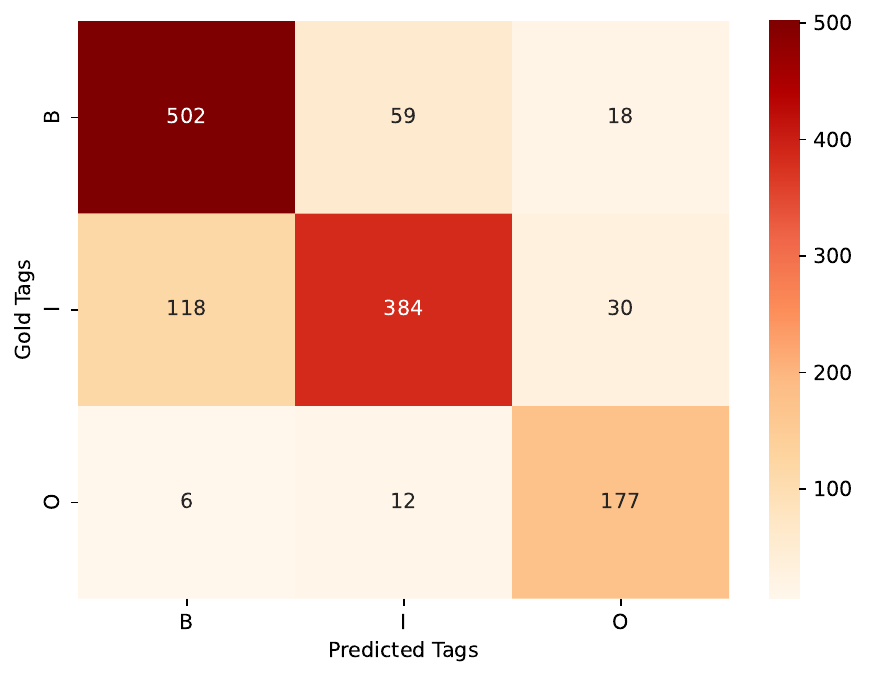}
        \caption{OSS}
    \end{subfigure}%
    \hfill
    
    \caption{Confusion matrices for all the models on the \textbf{segment detection} task.}
    \label{fig:five_images_grid}
\end{figure*}

\noindent \textbf{Error Analysis.} We analyze the performance of all models using the best-performing approach, standalone. We observe that the top model, Phi, occasionally confuses B tags with I tags. This pattern of confusion is consistent across the other models in our study (cf. Fig~\ref{fig:five_images_grid}). This behavior is intuitive, as some review comments can be interpreted as standalone statements instead of occurring within the broader context. Nevertheless, the presence of extra `B' tags does not substantially hinder \textbf{issue detection}, since our segment-level annotations ensure that each segment still corresponds to annotator detected issues within the dataset.

\subsection{Issue Identification} \label{sec:Issue_additional}

\subsubsection{Zero-shot Issue Identification} 
\begin{figure*}[t]
    \centering
    \includegraphics[width=\textwidth]{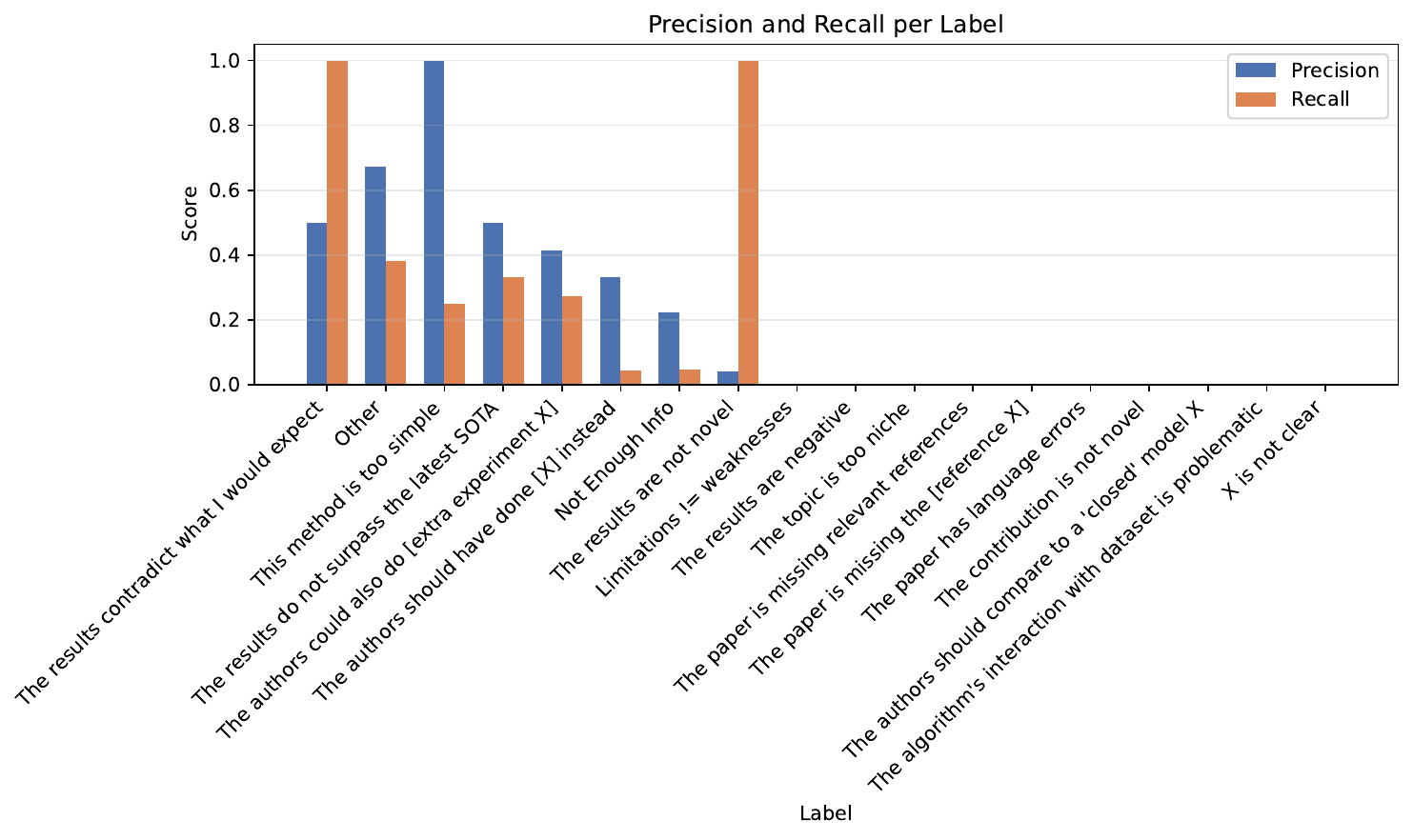}
    
    \caption{Zero shot results for \textbf{issue detection} with GPT OSS 20B. $8$ out of $18$ classes remain undetected yielding a miss rate of $8/18=0.444 \sim 44\%$.} 
    \vspace{4mm}
    \begin{minipage}{\textwidth}
        \small
        
    \end{minipage}
    \label{fig:issue_identification_LLM_challenge}
\end{figure*}

The zero-shot results from the best-performing LLM, GPT-OSS, fail to capture several classes that are sufficiently frequent in our dataset (cf. Fig.~\ref{fig:issue_identification_LLM_challenge}). While this type of classification achieves high precision for some classes, it introduces substantially more errors than a controlled approach.

\subsubsection{Baselines} For \textbf{issue detection}, we employ the following baselines: (i) Zero-shot (\texttt{LLM\textsubscript{Zero}}): the LLM predicts $l_i \subseteq \mathcal{L}$ given $s_i$; (ii) Finetuning (\texttt{LLM\textsubscript{Fine}}): we finetune the LLM on the training split and evaluate with cross-validation; (iii) QA-based classification (\texttt{LLM\textsubscript{QA}} / \texttt{LLM\textsubscript{QAFine}}): the pretrained LLM (\texttt{LLM\textsubscript{QA}}) or its finetuned variant (\texttt{LLM\textsubscript{QAFine}}) classifies $s_i$ using the same feature-specific questions and answers as in our approach. (iv) Encoder-Only Baseline: We use \texttt{RoBERTa}~\cite{liu2019robertarobustlyoptimizedbert} as a baseline, trained on the same data as the \texttt{LLM\textsubscript{Fine}} model, to evaluate the impact of task-specific finetuning using smaller models.
\subsubsection{Prompts for Issue Identification}
\begin{figure*}[t]
    \centering
    \includegraphics[width=\textwidth]{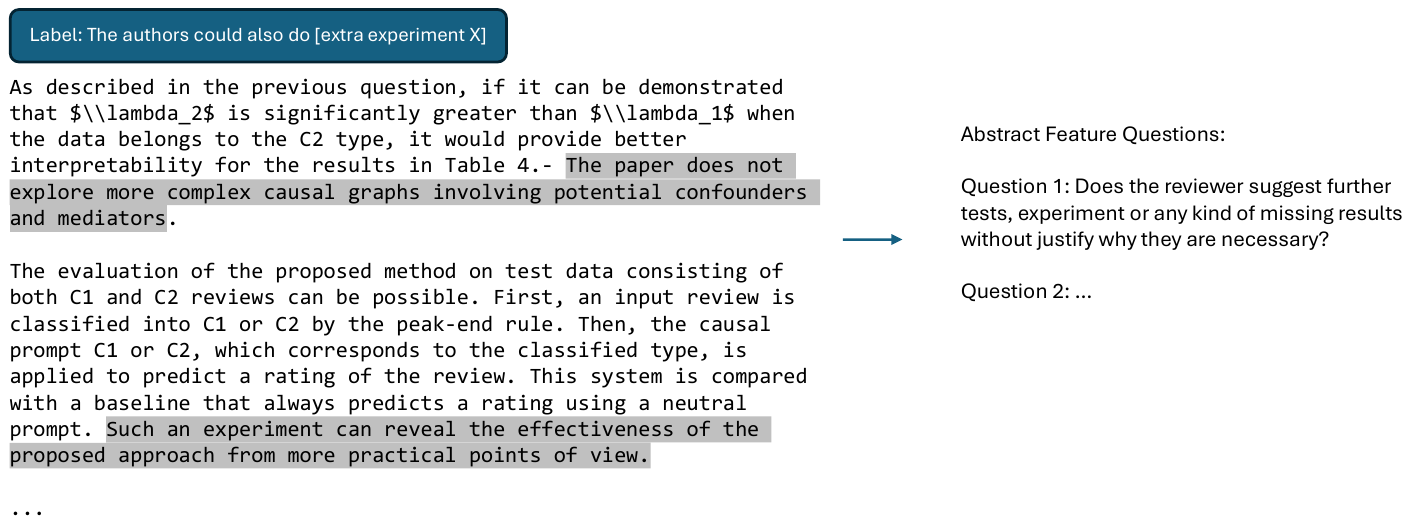}
    
    \caption{Example of abstract feature questions generated by our \textbf{issue detection} approach.}
    \label{fig:running example}
    \label{fig:issue_identification_question}
\end{figure*}

We present the \textbf{issue identification} prompt in Fig~\ref{fig:Issue_Identification_Prompt}. The feature extraction prompt extracts abstract-level feature and converts it into the feature question for the LLM to answer using the Feature QA prompt. The response from the Feature QA prompt is mapped to the discrete values: Yes $\rightarrow$ 1, No $\rightarrow$ -1, and Other $\rightarrow$ 0. We show a running example of a segment and abstract feature questions in Fig~\ref{fig:running example}. 

\subsubsection{Data Split for Issue Identification} \label{sec:data_split} 
\begin{figure*}[ht]
    \centering
    \title{50:50 split of our dataset comparing with distribution aware review level split}\includegraphics[width=1\textwidth,height=0.25\textheight]{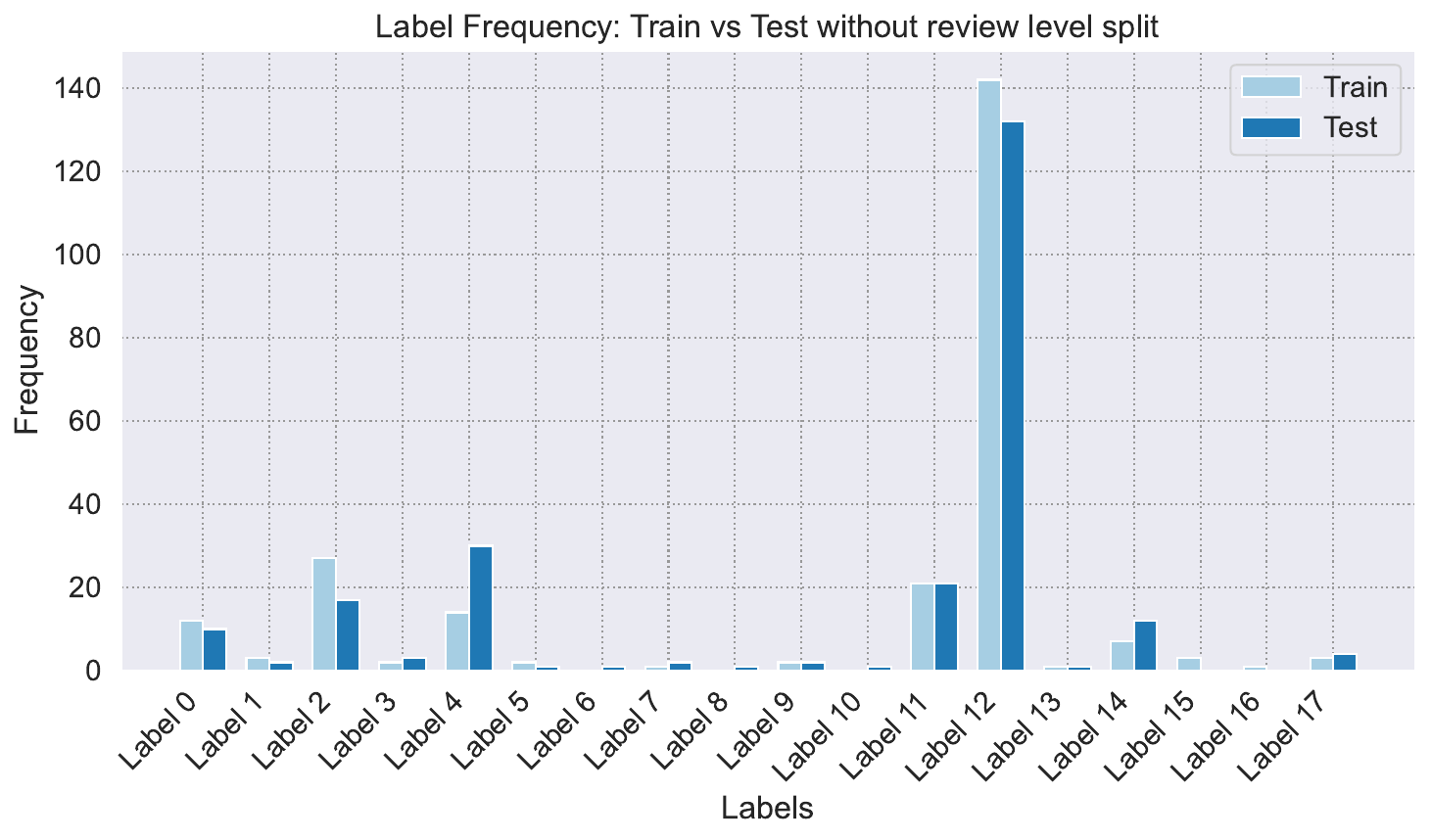}\\[1ex]
    \includegraphics[width=1\textwidth,height=0.25\textheight]{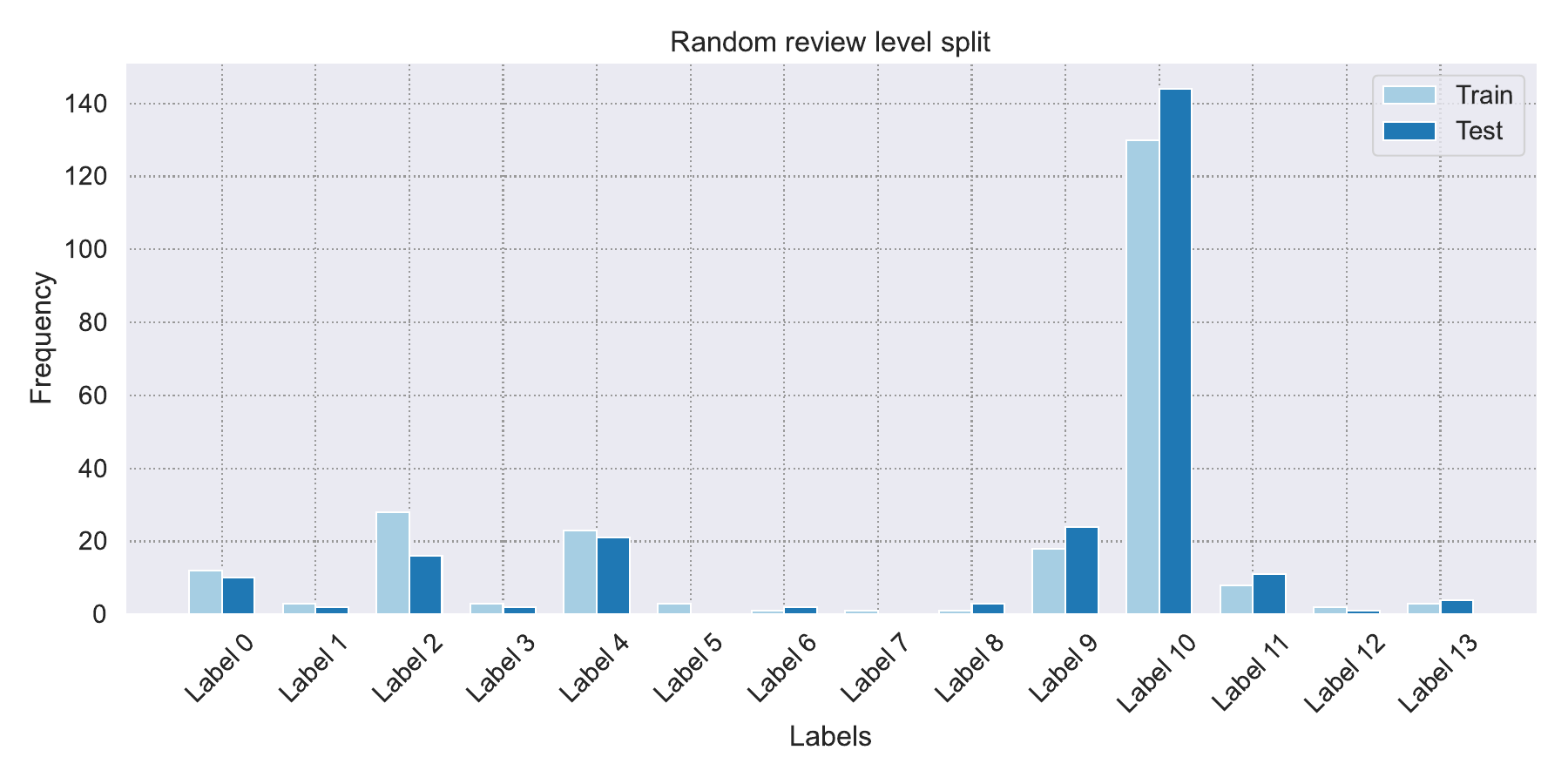}\\[1ex]
    \includegraphics[width=1\textwidth,height=0.25\textheight]{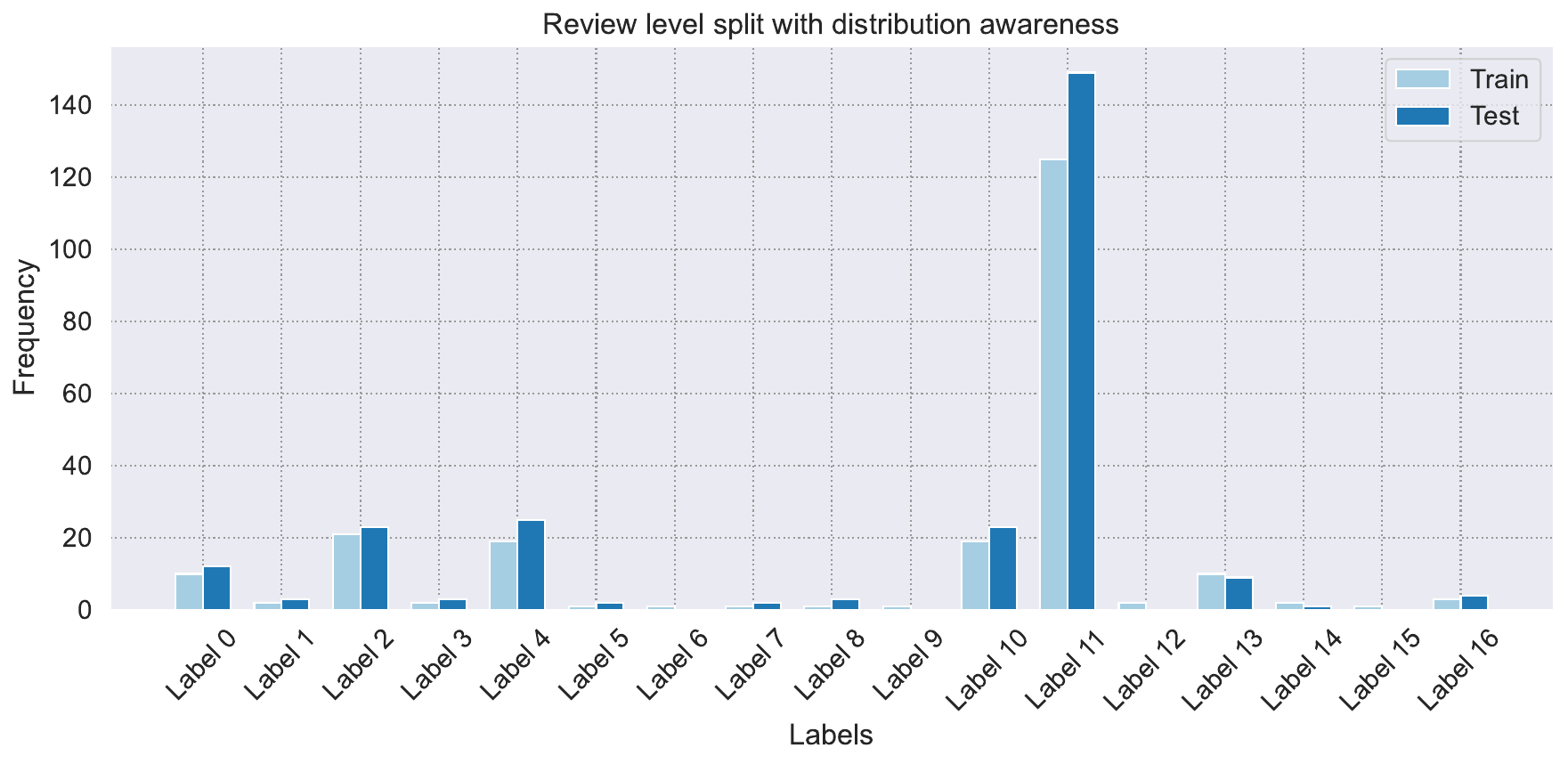}
    \caption{Comparing the label distribution using various split methods.\centering}
    \vspace{4mm}
    \label{fig:review_level_split}
\end{figure*}

We show the distributions of labels within train and test by employing various methods in Fig~\ref{fig:review_level_split}. Random review-level splits result in poor label balance, whereas our approach achieves nearly the same label balance as the \texttt{sklearn} train-test split. However, the \texttt{sklearn} split does not account for real-world scenarios, where full reviews are used as input for issue identification. In contrast, our distribution-aware split maintains label balance for effective cross-validation while also reflecting realistic use cases. We also show the distance across various split methods in Table~\ref{tab:distance}.

\subsubsection{Results}
\noindent \textbf{Overall results.}
We report overall performances in Table~\ref{tab:methods_comparison}. Our neurosymbolic approach, combining LLM-extracted features with a classical ML classifier, achieves the best results across the board, surpassing zero-shot performance by at least \textbf{0.39} points. Notably, \texttt{RoBERTa} outperforms all LLM-based baselines, reaffirming that task-specific finetuning with smaller models can be more effective than LLMs with limited data. Fine-tuning LLMs on our dataset yields minimum gains (0.02–0.10 in Yi), highlighting the value of a specialized dataset, yet still falls short of our approach due to LLMs' high data requirements for task adaptation. \texttt{LLM\textsubscript{QA}} improves over most of zero-shot variants, confirming the value of feature-aligned questioning, while \texttt{LLM\textsubscript{QAFine}} underperforms, likely due to data-constrained segment-level finetuning. Overall, Extra Trees paired with Yi- or GPT-OSS-generated feature delivers the strongest \textbf{issue detection}, benefiting from structured neurosymbolic representations and tree-based non-linear modeling. Low variance (below 0.01) further confirms dataset robustness (cf. Table~\ref{tab:crossval_results}). %Additionally, smaller models such as DeepSeek and Phi achieve better performance than larger LLMs which demonstrates that our framework is not critically dependent on a single architecture or scale.

\subsubsection{Ablation Study} 
\begin{figure}
    \centering
    \includegraphics[width=1.0\linewidth]{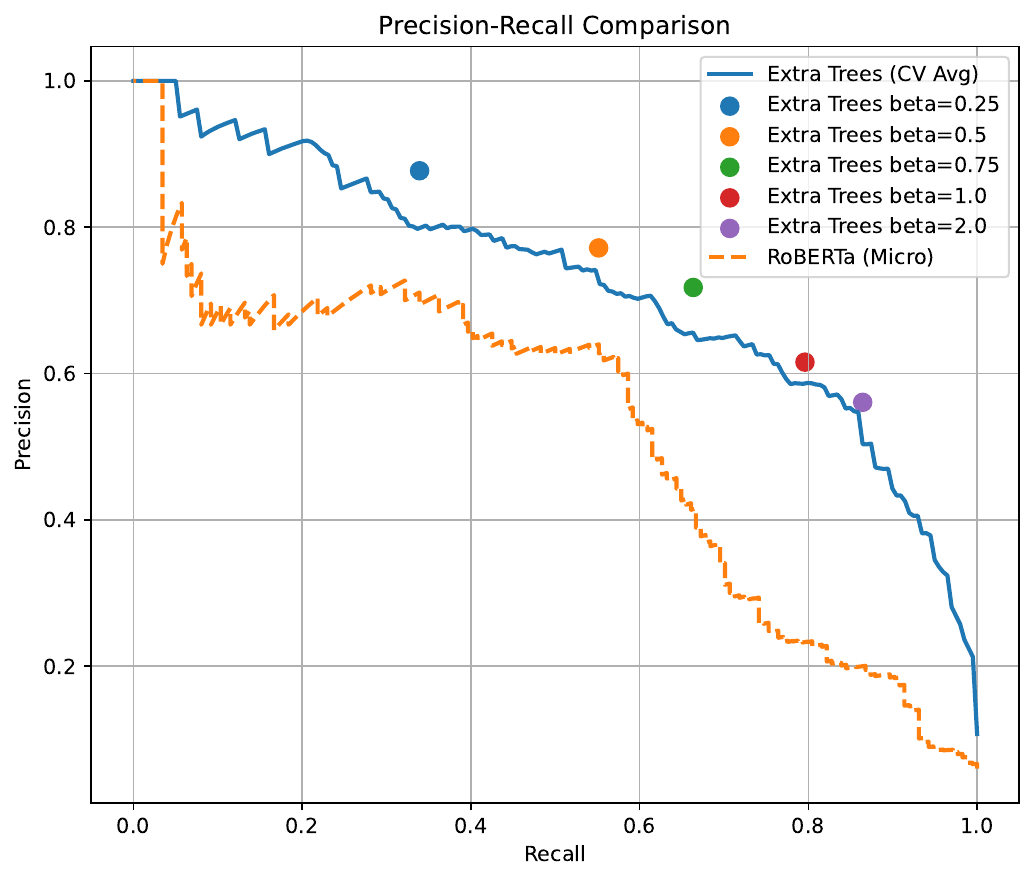}
    \caption{Precision-Recall Curve for \texttt{RoBERTa} and our approach operating at different thresholds.}
    \label{fig:threshold_plot}
\end{figure}

We conduct a series of ablation experiments to evaluate the effectiveness of our approach:
\textbf{1. Representation Effectiveness:} We employed a sentence embedding model to encode review segments into vector representations and evaluated their effectiveness for issue classification. Table~\ref{tab:embedding} reports the performance of different classifiers on the issue detection task using embeddings from the \texttt{all-MiniLM-L6-v2} sentence transformer. Table~\ref{tab:5_features} presents the performance of classifiers using feature vectors constructed from five questions on GPT. Precision and recall results for the baselines discussed in \S\ref{sec:exp} are provided in Table~\ref{tab:LLM_baseline_RP}. Overall, even the best-performing embedding-based classifier performs approximately \textbf{0.09} macro \(F_{0.5}\) points worse than our proposed feature-vector-based approach. \textbf{2. Feature question size}: We evaluate the impact of reducing the feature vector size from 10 to 5 using our best-performing model, gpt, due to its superior feature extraction capabilities. We find that with reducing the feature vector score \textbf{Extra Tree model} performs stably well in cross-validation\footnote{Details in Table \ref{tab:5_features}.}. \textbf{3. Effect of various thresholds}: We evaluate our approach against the strongest baseline, \texttt{RoBERTa}, across thresholds $\{0.25, 0.5, 0.75, 1.0, 2.0\}$ (Table~\ref{tab:model_thresholds}). At $\beta=0.25$, Extra Trees with GPT features achieves high precision (0.83) but low recall (0.44). Increasing $\beta$ (0.5–2.0) boosts recall at some precision loss (e.g., 0.89 at $\beta=2.0$), showing tunable precision–recall trade-offs. Fig.~\ref{fig:threshold_plot} shows our approach consistently outperforms this strongest baseline, providing reliable signals while balancing precision and recall. \noindent \textbf{4. Human-written questions}. We also experimented with manually crafted questions for each issue type. Designing 10 questions per issue required approximately 45 minutes, resulting in a total effort of over 20 hours for all 27 issue types. The corresponding results are reported in Table~\ref{tab:human_written_question_results}. Using human-written questions, Extra Trees with Phi features achieves an F0.5 of 0.57, compared to 0.65 with LLM-generated questions under the same configuration (cf. Table 1). Human-written questions thus underperform LLM-generated ones while requiring over 20 hours of manual effort. Moreover, review guidelines frequently evolve, making manual question engineering costly to maintain, whereas our LLM-based question generation readily adapts to changing review standards. Automated question generation is therefore preferable on both performance and practical grounds.\looseness=-1
%Please add the following packages if necessary:
%\usepackage{booktabs, multirow} % for borders and merged ranges
%\usepackage{soul}% for underlines
%\usepackage{xcolor,colortbl} % for cell colors
%\usepackage{changepage,threeparttable} % for wide tables
%If the table is too wide, replace \begin{table}[!htp]...\end{table} with
%\begin{adjustwidth}{-2.5 cm}{-2.5 cm}\centering\begin{threeparttable}[!htb]...\end{threeparttable}\end{adjustwidth}

%Please add the following packages if necessary:
%\usepackage{booktabs, multirow} % for borders and merged ranges
%\usepackage{soul}% for underlines
%\usepackage{xcolor,colortbl} % for cell colors
%\usepackage{changepage,threeparttable} % for wide tables
%If the table is too wide, replace \begin{table}[!htp]...\end{table} with
%\begin{adjustwidth}{-2.5 cm}{-2.5 cm}\centering\begin{threeparttable}[!htb]...\end{threeparttable}\end{adjustwidth}
\begin{table*}[!htp]\centering

%\resizebox{ extwidth}{!}{ % use this if the table is too large
\begin{tabular}{lrrrr}\toprule
&\textbf{Review-level-random-split} & \textbf{Review-level-split} & \textbf{sklearn-split} \\\midrule
Distance &0.036 &0.014 &0.015 \\
\bottomrule
\end{tabular}
\caption{Distance value for the three different split methods using evenly split 50\% of data \centering}\label{tab:distance}
\end{table*}

%Please add the following packages if necessary:
%\usepackage{booktabs, multirow} % for borders and merged ranges
%\usepackage{soul}% for underlines
%\usepackage{xcolor,colortbl} % for cell colors
%\usepackage{changepage,threeparttable} % for wide tables
%If the table is too wide, replace \begin{table}[!htp]...\end{table} with
%\begin{adjustwidth}{-2.5 cm}{-2.5 cm}\centering\begin{threeparttable}[!htb]...\end{threeparttable}\end{adjustwidth}
\begin{table*}[ht]
\centering
\resizebox{\textwidth}{!}{%
\begin{tabular}{l rr rr rr rr rr}
\toprule
\textbf{Method} 
& \multicolumn{2}{c}{\textbf{Yi}} 
& \multicolumn{2}{c}{\textbf{Phi}} 
& \multicolumn{2}{c}{\textbf{Qwen}} 
& \multicolumn{2}{c}{\textbf{Deep.}} 
& \multicolumn{2}{c}{\textbf{Oss.}} \\
\cmidrule(lr){2-3} \cmidrule(lr){4-5} \cmidrule(lr){6-7} \cmidrule(lr){8-9} \cmidrule(lr){10-11}
& Precision & Recall & Precision & Recall & Precision & Recall & Precision & Recall & Precision & Recall \\
\midrule
KNN                      & 0.62             & 0.53         & 0.65             & 0.63          & 0.65              & 0.53          & 0.41              & 0.18           & 0.67              & 0.65          \\
Logistic Regression (L2) & 0.58             & 0.48         & 0.61             & 0.50          & 0.60              & 0.45          & 0.39              & 0.29           & 0.65              & 0.52          \\
Logistic Regression (L1) & 0.58             & 0.44         & 0.63             & 0.46          & 0.63              & 0.47          & 0.42              & 0.31           & 0.69              & 0.52          \\
Random Forest            & 0.67             & 0.63         & 0.67             & 0.63          & 0.69              & 0.62          & 0.66              & 0.15           & 0.70              & 0.60          \\
Decision Tree            & 0.51             & 0.50         & 0.48             & 0.50          & 0.55              & 0.57          & 0.25              & 0.26           & 0.55              & 0.56          \\
SVM (RBF)                & 0.67             & 0.33         & 0.69             & 0.42          & 0.80              & 0.39          & 0.77              & 0.13           & 0.77              & 0.42          \\
SVM (Linear)             & 0.55             & 0.46         & 0.55             & 0.46          & 0.52              & 0.42          & 0.36              & 0.33           & 0.64              & 0.53          \\
Gradient Boosting        & 0.66             & 0.44         & 0.63             & 0.49          & 0.69              & 0.52          & 0.41              & 0.23           & 0.69              & 0.56          \\
AdaBoost                 & 0.43             & 0.02         & 0.32             & 0.02          & 0.76              & 0.04          & 0.39              & 0.29           & 0.45              & 0.03          \\
Extra Trees              & 0.69             & 0.63         & 0.65             & 0.63          & 0.68              & 0.61          & 0.68              & 0.17           & 0.69              & 0.61          \\
MLP                      & 0.64             & 0.54         & 0.64             & 0.57          & 0.64              & 0.53          & 0.49              & 0.28           & 0.70              & 0.61          \\ \hline
\bottomrule
\end{tabular}%
}
\caption{Precision and Recall of different LLMs using our approach for \textbf{issue detection}.}
\label{tab:precision_recall}
\end{table*}

\begin{figure*}[t]  % or [h!] depending on placement
\centering
\begin{tcolorbox}[colback=gray!10, colframe=black, title=Issue Identification Prompt Library]

\textbf{Feature Extraction Prompt} \\ \vspace{4mm}
\noindent\fbox{%
  \parbox{\textwidth}{%
You are given:
- An issue label: {issue}
- Why this issue is problematic: {problem}
- Review segments (from NLP paper reviews): {segments}
- Desired number of questions: 10

Your goal:
Extract abstract, high level features that determine whether each review segment reflects the specified issue. Express each feature as an objective Yes/No question.

Rules you MUST follow:
1) Use ONLY the provided segments. Do NOT rely on external knowledge or assumptions.
2) Work at the ABSTRACT level (features that could generalize across wording), not surface phrasing.
3) Every question must be answerable by inspecting a single segment in isolation.
4) Questions must be neutral (no leading language), atomic (one idea per question), and non-redundant.
5) Avoid double negatives, multi-part questions, subjective terms (“clearly”, “obviously”), or jargon unless it appears in the segments.
6) Prefer beginnings like “Does…”, “Is…”, “Are…”, “Has…”, “Do…”, “Can…”.
7) Keep each question concise (ideally 18 words).
8) Produce EXACTLY N questions. If N is not given, produce 10.
9) OUTPUT FORMAT: Return ONLY a Python list of strings, with double quotes, no code fences, no comments, no trailing commas.
10) Do NOT include any analysis, notes, or explanations in the output — only the list.

Step-by-step procedure:
A) Parse and inventory:
   - Scan the segments to note common patterns about claims, evidence, specificity, comparisons, citations, quantification, assumptions, scope/coverage, and consistency.
   - Identify cues that would indicate presence/absence of the issue {issue}.

B) Abstract feature mining:
   - Convert recurring patterns into abstract properties that can reflect the similarity between the review segments you are given
   - Ensure each property directly supports diagnosing {issue} as problematic because {problem}.

C) Draft discriminative Yes/No tests:
   - For each property, write a Yes/No question that a reviewer could answer from a single segment.
   - Ensure questions collectively cover: evidence/citations, specificity/locating, correctness/faithfulness, scope, quantification, reproducibility/actionability, internal consistency, and fairness/balance (as applicable).

D) Prune and refine:
   - Remove overlaps; keep the most general, segment-checkable forms.
   - Rewrite to be atomic, neutral, and concise.
   - Ensure each question distinguishes “issue present” vs “issue absent”.

E) Final checks (must pass all):
   - [ ] Exactly N questions.
   - [ ] All are Yes/No answerable from a single segment.
   - [ ] No duplicates or near-duplicates.
   - [ ] No restating {issue} verbatim; focus on testable properties.
   - [ ] Python list of strings ONLY, no extra text.

Now produce the final output.
  }%
}

\textbf{Feature QA prompt} \\ \vspace{2mm}\label{app:QA_prompt}
\noindent\fbox{%
  \parbox{\textwidth}{%

You will be given a review segment. Your task is to evaluate its quality by answering a series of Yes/No questions.

Review Segment: {review}

Question: {question}

Respond strictly with either:

[[Yes]] if the answer is Yes

[[No]] if the answer is No

[[Other]] if the question is irrelevant

  }%
}

\end{tcolorbox}
\caption{Issue Identification Prompt}
\label{fig:Issue_Identification_Prompt}
\end{figure*}

\begin{table*}[!t]
\centering
\resizebox{!}{0.5\textwidth}{%
\begin{tabular}{l l c c c c}
\toprule
\textbf{Method} & \textbf{Model} & \textbf{Const.} & \textbf{Relev.} & \textbf{Spec.} & \textbf{Conc.} \\
\midrule
\multirow{5}{*}{\texttt{1-pass}} 
& Yi       & 1.9 & 2.0 & 1.8 & 1.9 \\ \cmidrule(lr){2-6}
& Phi      & \underline{2.1} & \underline{2.2} & \underline{2.0} & \underline{2.1} \\ \cmidrule(lr){2-6}
& Qwen     & 1.8 & 1.9 & 1.7 & 1.8 \\ \cmidrule(lr){2-6}
& Deep.    & 1.7 & 1.8 & 1.6 & 1.7 \\ \cmidrule(lr){2-6}
& Oss.     & 1.9 & 2.0 & 1.8 & 1.9 \\
\specialrule{1.5pt}{1pt}{1pt}
\multirow{5}{*}{\texttt{BoN}} 
& Yi       & 2.0 & 2.2 & 2.4 & 2.3 \\ \cmidrule(lr){2-6}
& Phi      & 2.2 & 2.3 & 2.1 & 2.2 \\ \cmidrule(lr){2-6}
& Qwen     & 2.0 & 2.1 & 2.0 & 2.0 \\ \cmidrule(lr){2-6}
& Deep.    & 1.8 & 1.9 & 1.7 & 1.8 \\ \cmidrule(lr){2-6}
& Oss.     & 2.0 & 2.1 & 2.0 & 2.1 \\ 
\specialrule{1.5pt}{1pt}{1pt}
\multirow{5}{*}{\texttt{Self-Ref.}}  
& Yi       & 2.3 & 2.4 & 2.4 & 2.5 \\ \cmidrule(lr){2-6}
& Phi      & 2.4 & 2.6 & 2.8 & 2.9 \\ \cmidrule(lr){2-6}
& Qwen     & 2.2 & 2.4 & 2.4 & 2.7 \\ \cmidrule(lr){2-6}
& Deep.    & 2.4 & 2.4 & 2.5 & 2.3 \\ \cmidrule(lr){2-6}
& Oss.     & 3.0 & 3.1 & 2.9 & 3.0 \\ 
\specialrule{1.5pt}{1pt}{1pt}
\multirow{5}{*}{\texttt{Temp}}  
& Yi       & 2.8 & 2.9 & 2.7 & 2.8 \\ \cmidrule(lr){2-6}
& Phi      & \underline{3.1} & \underline{3.2} & \underline{3.0} & \underline{3.3} \\ \cmidrule(lr){2-6}
& Qwen     & 2.9 & 3.0 & 2.8 & 2.9 \\ \cmidrule(lr){2-6}
& Deep.    & 2.7 & 2.8 & 2.6 & 2.8 \\ \cmidrule(lr){2-6}
& Oss.     & 2.8 & 2.9 & 2.7 & 2.8 \\ 
\specialrule{1.5pt}{1pt}{1pt}
\multirow{5}{*}{\texttt{Plan}}  
& Yi       & 3.0 & 3.1 & 2.9 & 3.0 \\ \cmidrule(lr){2-6}
& Phi      & \underline{3.3} & \underline{3.4} & \underline{3.2} & \underline{3.5} \\ \cmidrule(lr){2-6}
& Qwen     & 3.1 & 3.2 & 3.0 & 3.1 \\ \cmidrule(lr){2-6}
& Deep.    & 2.9 & 3.0 & 2.8 & 3.0 \\ \cmidrule(lr){2-6}
& Oss.     & 3.0 & 3.1 & 2.9 & 3.0 \\ 
\specialrule{1.5pt}{1pt}{1pt}
\multirow{5}{*}{\texttt{Temp + Plan}}  
& Yi       & 3.6 & 3.5 & 3.1  & 3.1  \\ \cmidrule(lr){2-6}
& Phi      & \underline{3.4} & \underline{3.5} & 3.4 & 3.6 \\ \cmidrule(lr){2-6}
& Qwen     & 3.4 & 3.4 & 3.4 & 3.4  \\ \cmidrule(lr){2-6}
& Deep.    & 3.2 & 3.3 & 3.1  & 3.2  \\ \cmidrule(lr){2-6}
& Oss.     & 3.2 & 3.4 & 3.2 & 3.2  \\ 
\specialrule{1.5pt}{1pt}{1pt}
\multirow{5}{*}{\texttt{Ours}}       
& Yi       & 3.9 & 3.8 & 3.8 & 3.8 \\ \cmidrule(lr){2-6}
& Phi      & \underline{\textbf{4.3}} & \underline{\textbf{4.3}} & \underline{\textbf{4.2}} & \underline{\textbf{4.3}} \\ \cmidrule(lr){2-6}
& Qwen     & 3.8 & 3.8 & 3.7 & 3.7 \\ \cmidrule(lr){2-6}
& Deep.    & 3.5 & 3.6 & 3.6 & 3.5 \\ \cmidrule(lr){2-6}
& Oss.     & 4.0 & 4.1 & 4.0 & 3.8 \\ 
\bottomrule
\end{tabular}}
\caption{Performance comparison of various models on the \textbf{feedback generation} task using four customized \textbf{automated metrics} from \texttt{Prometheus} V2: Constructiveness (Const.), Relevance (Relev.), Specificity (Spec.), and Conciseness (Conc.). Overall best results are \textbf{bolded} and method-wise best results are \underline{underlined}.}
\label{tab:feedback_grouped_model}
\end{table*}

\begin{table*}[!t]
\centering
\begin{minipage}{0.48\textwidth}
\centering
\resizebox{0.6\textwidth}{!}{\begin{tabular}{l r}
\toprule
\textbf{Classifier} & \textbf{F0.5} \\
\midrule
KNN & \textbf0.59 \\
Logistic Regression (L2) & 0.49 \\
Logistic Regression (L1) & 0.48 \\
Random Forest & 0.56 \\
Decision Tree & 0.37 \\
SVM (RBF) & \textbf{0.60} \\
SVM (Linear) & 0.45 \\

Gradient Boosting & 0.47 \\
AdaBoost & 0.51 \\
Extra Trees & 0.59 \\
Multi-layer Perceptron (MLP) & 0.56 \\

%Stochastic Gradient Descent (SGD) & 0.30 \\
\bottomrule
\end{tabular}}
\caption{Performance of various classifiers on the \textbf{issue detection} task using representations obtained from \texttt{all-MiniLM-L6-v2} sentence transformer model. \centering}
\label{tab:embedding}
\end{minipage}
\hfill
\begin{minipage}{0.48\textwidth}
\centering
\resizebox{0.8\textwidth}{!}{\begin{tabular}{l l r r}
\toprule
\textbf{Classifier} & \textbf{F0.5} & \textbf{Precision} & \textbf{Recall} \\
\midrule
KNN & 0.67 & 0.68 & 0.66 \\
Logistic Regression (L2) & 0.64 & 0.69 & 0.50 \\
Logistic Regression (L1) & 0.64 & 0.70 & 0.48 \\
Random Forest & 0.67 & 0.70 & 0.48 \\
Decision Tree & 0.58 & 0.57 & 0.60 \\
SVM (RBF) & \textbf{0.70} & 0.79 & 0.50 \\
SVM (Linear) & 0.62 & 0.67 & 0.50 \\
AdaBoost & 0.17 & 0.72 & 0.04 \\
GBoost & 0.66 & 0.70 & 0.56 \\
Extra Trees & \textbf{0.67} & \textbf{0.67} & \textbf{0.65} \\
MLP & 0.65 & 0.66 & 0.61 \\
%0.36 & 0.39 & 0.28 \\
\bottomrule
\end{tabular}}
\caption{Performance of various classifiers using GPT oss generated 5 question feature vector for the \textbf{issue detection} task.}
\label{tab:5_features}
\end{minipage}
\end{table*}

%Please add the following packages if necessary:
%\usepackage{booktabs, multirow} % for borders and merged ranges
%\usepackage{soul}% for underlines
%\usepackage{xcolor,colortbl} % for cell colors
%\usepackage{changepage,threeparttable} % for wide tables
%If the table is too wide, replace \begin{table}[!htp]...\end{table} with
%\begin{adjustwidth}{-2.5 cm}{-2.5 cm}\centering\begin{threeparttable}[!htb]...\end{threeparttable}\end{adjustwidth}
% \begin{table*}[!ht]
% \centering

% \resizebox{0.8\textwidth}{!}{%
% \begin{tabular}{lrrrrr}\toprule
% \textbf{Methods} & \textbf{Ours. train} & \textbf{Ours. test} & \textbf{Emb. train} & \textbf{Emb. test}\\\midrule
% KNN &0.36 &0.37 &0.40 &0.19  \\
% Logistic Regression (L2) &0.42 &0.43 &0.47 &0.18  \\
% Logistic Regression (L1) &0.38 &0.41 &0.42 &0.18  \\
% Random Forest &0.40 &0.45 &0.40 &0.22  \\
% Decision Tree &0.34 &0.37 &0.28 &0.19  \\
% SVM (RBF) &0.25 &0.39 &0.46 &0.19  \\
% SVM (Linear) &0.36 &0.40 &0.45 &0.17  \\
% SVM (Poly) &0.02 &0.09 &0.34 &0.32  \\
% Gradient Boosting &0.36 &0.41 &0.33 &0.21  \\
% AdaBoost &0.42 &0.39 &0.33 &0.17  \\
% Extra Trees &0.37 &0.44 &0.46 &0.28  \\
% Multi-layer Perceptron (MLP) &0.39 &0.40 &0.43 &0.18  \\
% Gaussian Naive Bayes &0.30 &0.31 &0.42 &0.18  \\
% Stochastic Gradient Descent (SGD) &0.35 &0.43 &0.38 &0.20  \\
% \bottomrule
% \end{tabular}%
% }
% \caption{F0.5 score on even split of training and test with 50\% each for \textbf{issue detection}. Embedding (Emb.) corresponds to representations obtained from \texttt{all-MiniLM-L6-v2} sentence transformer model.}
% \label{tab:overfitting}
% \end{table*}
\begin{table*}[!t]
\centering

\resizebox{0.8\textwidth}{!}{\begin{tabular}{l|ccc|ccc|ccc|ccc|ccc}
\toprule
\textbf{Model} & \multicolumn{3}{c|}{\textbf{0.25}} & \multicolumn{3}{c|}{\textbf{0.5}} & \multicolumn{3}{c|}{\textbf{0.75}} & \multicolumn{3}{c|}{\textbf{1.0}} & \multicolumn{3}{c}{\textbf{2.0}} \\
\cmidrule(lr){2-4}\cmidrule(lr){5-7}\cmidrule(lr){8-10}\cmidrule(lr){11-13}\cmidrule(lr){14-16}
 & F$_{0.25}$ & P & R & F$_{0.50}$ & P & R & F$_{0.75}$ & P & R & F$_{1.0}$ & P & R & F$_{2.0}$ & P & R \\
\midrule
KNN                          & 0.67        & 0.71            & 0.43         & 0.63       & 0.66           & 0.55        & 0.63        & 0.61            & 0.70         & 0.65       & 0.56           & 0.79        & 0.74       & 0.50           & 0.86        \\
Logistic Regression (L2)     & 0.74        & 0.79            & 0.44         & 0.68       & 0.70           & 0.64        & 0.68        & 0.68            & 0.68         & 0.69       & 0.63           & 0.77        & 0.78       & 0.50           & 0.92        \\
Logistic Regression (L1)     & 0.74        & 0.83            & 0.35         & 0.67       & 0.68           & 0.64        & 0.68        & 0.64            & 0.77         & 0.70       & 0.62           & 0.82        & 0.79       & 0.54           & 0.90        \\
Random Forest                & 0.74        & 0.81            & 0.41         & 0.69       & 0.71           & 0.66        & 0.70        & 0.67            & 0.74         & 0.70       & 0.66           & 0.76        & 0.78       & 0.54           & 0.89        \\
Decision Tree                & 0.53        & 0.53            & 0.54         & 0.53       & 0.53           & 0.54        & 0.53        & 0.53            & 0.54         & 0.53       & 0.53           & 0.54        & 0.54       & 0.53           & 0.54        \\
SVM (RBF)                    & 0.76        & 0.82            & 0.36         & 0.71       & 0.73           & 0.67        & 0.72        & 0.69            & 0.79         & 0.73       & 0.69           & 0.79        & 0.78       & 0.64           & 0.83        \\
SVM (Linear)                 & 0.71        & 0.79            & 0.40         & 0.68       & 0.67           & 0.72        & 0.69        & 0.66            & 0.75         & 0.71       & 0.63           & 0.80        & 0.77       & 0.62           & 0.83        \\
Gradient Boosting            & 0.68        & 0.71            & 0.49         & 0.66       & 0.66           & 0.67        & 0.67        & 0.63            & 0.75         & 0.69       & 0.62           & 0.78        & 0.76       & 0.55           & 0.85        \\
AdaBoost                     & 0.74        & 0.85            & 0.24         & 0.67       & 0.67           & 0.68        & 0.68        & 0.64            & 0.76         & 0.70       & 0.62           & 0.80        & 0.79       & 0.52           & 0.91        \\
Multi-layer Perceptron (MLP) & 0.76        & 0.80            & 0.42         & 0.69       & 0.73           & 0.58        & 0.67        & 0.67            & 0.71         & 0.68       & 0.67           & 0.71        & 0.75       & 0.54           & 0.85        \\
Extra Trees                  & 0.72        & 0.83            & 0.44         & 0.70       & 0.71           & 0.67        & 0.70        & 0.69            & 0.71         & 0.71       & 0.65           & 0.79        & 0.78       & 0.55           & 0.89       \\
\bottomrule
\end{tabular}}
\caption{Performance of different models across thresholds 0.25, 0.5, 0.75, 1.0, and 2.0. Metrics shown are  F\textsubscript{threshold} (e.g., F$_{0.25}$ for threshold 0.25), P (Precision), and R (Recall).}
\label{tab:model_thresholds}
\end{table*}

\begin{table*}[ht]
\centering

\resizebox{0.6\textwidth}{!}{\begin{tabular}{lcccccccccc}
\toprule
\multirow{2}{*}{\textbf{Baselines}} & \multicolumn{2}{c}{\textbf{Yi}} & \multicolumn{2}{c}{\textbf{Phi}} & \multicolumn{2}{c}{\textbf{Qwen}} & \multicolumn{2}{c}{\textbf{Deep.}} & \multicolumn{2}{c}{\textbf{Oss.}} \\
\cmidrule(lr){2-3} \cmidrule(lr){4-5} \cmidrule(lr){6-7} \cmidrule(lr){8-9} \cmidrule(lr){10-11}
 & P & R & P & R & P & R & P & R & P & R \\
\midrule
\texttt{LLM\textsubscript{zero}}      & 0.22 & 0.01 & 0.09 & 0.08 & 0.18 & 0.17 & 0.03 & 0.30 & 0.26 & 0.24 \\
\texttt{LLM\textsubscript{Fine}}     & 0.23 & 0.23 & 0.11 & 0.18 & 0.13 & 0.17 & 0.10 & 0.19 & 0.19 & 0.21 \\
\texttt{LLM\textsubscript{QA}}         & 0.15 & 0.09 & 0.18 & 0.04 & 0.12 & 0.13 & 0.19 & 0.26 & 0.10 & 0.40 \\
\texttt{LLM\textsubscript{QAFine}}    & 0.10 & 0.20 & 0.09 & 0.27 & 0.05 & 0.12 & 0.07 & 0.19 & 0.10 & 0.17 \\
\bottomrule
\end{tabular}}
\caption{Precision (P) and Recall (R) of various LLMs for the \textbf{issue detection} task.}
\label{tab:LLM_baseline_RP}
\end{table*}

\begin{table}[!t]
\centering
\resizebox{0.4\textwidth}{!}{\begin{tabular}{lccccc c}
\toprule
\textbf{Metric} & \textbf{Fold 1} & \textbf{Fold 2} & \textbf{Fold 3} & \textbf{Fold 4} & \textbf{Fold 5} & \textbf{Variance} \\
\midrule
F$_{0.5}$     & 0.55 & 0.65 & 0.69 & 0.66 & 0.57 & 0.004 \\
Precision     & 0.55 & 0.53 & 0.70 & 0.71 & 0.63 & 0.006 \\
Recall        & 0.46 & 0.50 & 0.48 & 0.54 & 0.45 & 0.001 \\
\bottomrule
\end{tabular}}
\caption{Cross-validation results over 5 folds showing F$_{0.5}$, Precision, Recall, and their variances.}
\label{tab:crossval_results}
\end{table}

\begin{table}[!t]
\centering
\resizebox{0.4\textwidth}{!}{\begin{tabular}{lccc}
\toprule
\textbf{Model} & \textbf{Mean F$_{0.5}$} & \textbf{Mean Precision} & \textbf{Mean Recall} \\
\midrule
KNN & 0.544 & 0.667 & 0.314 \\
Logistic Regression (L2) & 0.514 & 0.576 & 0.365 \\
Logistic Regression (L1) & 0.456 & 0.477 & 0.391 \\
Random Forest & 0.554 & 0.761 & 0.270 \\
SVM (RBF) & 0.463 & 0.948 & 0.153 \\
Extra Trees & 0.575 & 0.783 & 0.281 \\
\bottomrule
\end{tabular}}
\caption{Mean F$_{0.5}$, precision, and recall across different models using human-written questions as features.}
\label{tab:human_written_question_results}
\end{table}

\subsection{Additional details on feedback generation}
\label{sec:details_feedback_generation}
\subsubsection{Alignment Study with \texttt{Prometheus} and other evaluators} \label{sec:alignment}
To assess the reliability of our reference-free automated metric, we conducted an alignment study comparing its scores with those provided by \texttt{Prometheus} and human annotators. The results, summarized in Table~\ref{tab:model_high_correlation_metrics}, show an average Spearman correlation of \textbf{0.85}, demonstrating the strong effectiveness of the metric. We further compare the performance of \texttt{Prometheus} v1, \texttt{Prometheus} v2 and GPT 4o in Table~\ref{tab:comp_metrics}. \texttt{Prometheus} V2 outperforms all the other methods across the board in terms of Kendall $\tau$.  

\begin{table*}
\centering
\resizebox{2.0\columnwidth}{!}{
\begin{tabular}{lcccccccc}
\toprule
\multirow{2}{*}{\textbf{Model}} 
& \multicolumn{2}{c}{\textbf{Constructiveness}} 
& \multicolumn{2}{c}{\textbf{Relevance}} 
& \multicolumn{2}{c}{\textbf{Specificity}} 
& \multicolumn{2}{c}{\textbf{Conciseness}} \\
\cmidrule(lr){2-3}\cmidrule(lr){4-5}\cmidrule(lr){6-7}\cmidrule(lr){8-9}
& Kendall $\tau$ & Spearman $\rho$ 
& Kendall $\tau$ & Spearman $\rho$ 
& Kendall $\tau$ & Spearman $\rho$ 
& Kendall $\tau$ & Spearman $\rho$ \\
\midrule
Yi        & 0.75 & 0.83 & 0.76 & 0.84 & 0.75 & 0.83 & 0.76 & 0.85 \\
Phi       & \underline{0.78} & \underline{0.88} & \underline{0.77} & \underline{0.87} & \underline{0.76} & \underline{0.86} & \underline{0.78} & \underline{0.88} \\
Qwen      & 0.76 & 0.84 & 0.75 & 0.85 & 0.76 & 0.84 & 0.75 & 0.84 \\
Deep.     & 0.75 & 0.83 & 0.75 & 0.84 & 0.75 & 0.83 & 0.75 & 0.83 \\
Oss.      & 0.77 & 0.85 & 0.76 & 0.85 & 0.76 & 0.85 & 0.77 & 0.86 \\
\midrule
\textbf{Average} & 0.76 & 0.85 & 0.76 & 0.85 & 0.76 & 0.84 & 0.76 & 0.85 \\
\bottomrule
\end{tabular}}
\caption{Kendall $\tau$ and Spearman $\rho$ correlations with \texttt{Prometheus} for different models across four metrics: Constructiveness, Relevance, Specificity, and Conciseness. Method-wise best correlations are \underline{underlined}. The last row shows the average correlation across all models and metrics.}
\label{tab:model_high_correlation_metrics}
\end{table*}
\begin{table*}[t]
\centering
\small
\begin{tabular}{lcccccccccccc}
\toprule
& \multicolumn{3}{c}{Constr.} 
& \multicolumn{3}{c}{Rel.} 
& \multicolumn{3}{c}{Spec.} 
& \multicolumn{3}{c}{Conc.} \\
\cmidrule(lr){2-4} \cmidrule(lr){5-7} \cmidrule(lr){8-10} \cmidrule(lr){11-13}
Model & Pr-v1 & GPT-4o & Pr-v2 & Pr-v1 & GPT-4o & Pr-v2 & Pr-v1 & GPT-4o & Pr-v2 & Pr-v1 & GPT-4o & Pr-v2 \\
\midrule
Yi    & 0.73 & 0.76 & \textbf{0.75} & 0.74 & 0.75 & \textbf{0.76} & 0.73 & 0.74 & \textbf{0.75} & 0.74 & 0.75 & \textbf{0.76} \\
Phi   & 0.77 & 0.78 & \textbf{0.78} & 0.76 & 0.77 & \textbf{0.77} & 0.75 & 0.76 & \textbf{0.76} & 0.77 & 0.77 & \textbf{0.78} \\
Qwen  & 0.75 & 0.75 & \textbf{0.76} & 0.74 & 0.75 & \textbf{0.75} & 0.75 & 0.75 & \textbf{0.76} & 0.74 & 0.75 & \textbf{0.75} \\
Deep. & 0.74 & 0.74 & \textbf{0.75} & 0.74 & 0.74 & \textbf{0.75} & 0.74 & 0.74 & \textbf{0.75} & 0.74 & 0.74 & \textbf{0.75} \\
Oss.  & 0.76 & 0.76 & \textbf{0.77} & 0.75 & 0.75 & \textbf{0.76} & 0.75 & 0.75 & \textbf{0.76} & 0.76 & 0.76 & \textbf{0.77} \\
\midrule
Avg   & 0.75 & 0.76 & \textbf{0.76} & 0.75 & 0.76 & \textbf{0.76} & 0.74 & 0.75 & \textbf{0.76} & 0.75 & 0.76 & \textbf{0.76} \\
\bottomrule
\end{tabular}
\caption{Kendall $\tau$ correlations across review quality dimensions for \texttt{Prometheus} v1 (Pr-v1), v2 (Pr-v2) and GPT-4o respectively.}
\label{tab:comp_metrics}
\end{table*}

\subsubsection{Experimental Setup and Analysis for review rewrites and role extension} \label{sec:exp_role}
To identify occurrences of review extension and role separation in LLM-generated feedback, we use GPT-OSS 120B~\cite{openai2025gptoss120bgptoss20bmodel} as a judge, prompting it to detect such issues within zero-shot generations using the instructions in Table~\ref{tab:review_issues}. The alignment between GPT-OSS judgments and human evaluations yields a Spearman correlation of 0.85. Table~\ref{tab:review_issues} presents the results, illustrating the prevalence of role separation’ and review extension’ in most of LLM-generated feedback. Example cases of role separation and review extension are shown in Fig.~\ref{fig:role_sep} and Fig.~\ref{fig:review_ext}, respectively.
\begin{table}[t]
\centering
\small{\begin{tabular}{lcc}
\toprule
\textbf{Model} & \textbf{Review Ext. (\%)} & \textbf{Role Sep. (\%)} \\
\midrule
Yi & 42 & 23 \\
DeepSeek & 47 & 30 \\
OSS & 43 & 28 \\
Qwen & 45 & 25 \\
Phi & \textbf{41} & \textbf{23} \\
\bottomrule
\end{tabular}}
\caption{Comparison of review generation issues across models. Review Extension refers to instances where the LLM generates full reviews instead of targeted feedback, while Role Separation indicates cases where feedback is misdirected to authors instead of reviewers. The model having least amount of the issues is highlighted.}
\label{tab:review_issues}
\end{table}

\begin{figure*}[!t]
\begin{tcolorbox}[
  colback=gray!5,
  colframe=black,
  title={Review Extension},
  boxrule=0.5mm,
  arc=2mm,
  fontupper=\small
]

\textbf{Original Comment:} “...no check for false negatives… only one language, one domain, one genre… no significance testing.”
\textbf{Issue Detected}: \textbf{The authors could also do extra experiment, X}

\textbf{Generated feedback:}

You can rewrite your comment as the follows: While the paper presents a promising benchmark and methodology, the evaluation appears limited in scope. The experiments focus solely on English data from a single domain and genre, which may restrict the generalizability of the results. Including experiments across multiple languages or domains could strengthen the empirical claims. Additionally, incorporating significance testing would help confirm that the observed improvements are statistically meaningful. Overall, expanding the evaluation would improve both the rigor and the impact of the work.
\end{tcolorbox}
\caption{Review Extension issues of the models: Generating full reviews instead of feedback}
\label{fig:review_ext}
\end{figure*}

\begin{figure*}
\begin{tcolorbox}[
  colback=gray!5,
  colframe=black,
  title={Role Separation (Author-Facing Feedback)},
  boxrule=0.5mm,
  arc=2mm,
  fontupper=\small
]

\textbf{Original Comment:} “...no check for false negatives… only one language, one domain, one genre… no significance testing.”

\textbf{Issue Detected}: \textbf{The authors could also do extra experiment, X}

\textbf{Generated feedback:}

The current experiments demonstrate the benchmark and methodology effectively, but they are limited to English, a single domain, and one genre. To strengthen the claims, consider including additional languages, domains, or genres where feasible. Also, performing significance testing would provide stronger evidence for the reported improvements. While these are suggestions to enhance the work, the main contribution remains valid and valuable within the presented scope.

\end{tcolorbox}
\caption{Role Separation issue in our model: Generating feedback to authors instead of reviewers}
\label{fig:role_sep}
\end{figure*}
\subsubsection{List of forbidden terms in the fitness function}
To ensure that feedback remains professional, actionable, and relevant, we identify greetings and off-topic expressions that should be avoided in review comments. Such terms do not contribute to the evaluation of the work and may distract or confuse the author. In our zero-shot feedback generations, we found that approximately 82\% of the divergent feedback contained one or more of these terms. Consequently, we compiled them into a list of forbidden terms to promote professionalism and maintain focus in reviewer feedback. \texttt{forbidden\_terms = ["Hi", "Hello", "Hey", "Dear Author", "To whom it may concern", "Greetings", "Good morning", "Good afternoon", "Good evening", "Haha", "Hehe", "Lmao", "OMG", "Wow", "FYI", "Cheers", "Best regards", "Sincerely", "I think"]}

\subsubsection{Prompts for feedback generation}
\paragraph{Prompt for \texttt{Prometheus}} The Conciseness prompt is in Fig~\ref{fig:conciseness}, Relevance is in Fig~\ref{fig:relevance}, Specificity is in Fig~\ref{fig:specificity} and Constructiveness is in Fig~\ref{fig:constructiveness} respectively. 
\begin{figure*}[t]
\centering
\begin{tcolorbox}[colback=gray!10, colframe=black, title=Conciseness Evaluation Prompt]

\textbf{Task Description:} 
You are given an instruction, a response to evaluate, and a score rubric representing an evaluation criterion. 

Your goal: 
1. Write detailed feedback that assesses the response strictly based on the rubric.
2. Assign a score between 1 and 5, referring to the rubric.
3. Output in the following format: 
\begin{quote}
Feedback: (write a feedback for criteria) [RESULT] (an integer number between 1 and 5)
\end{quote}
4. Do not include any opening, closing, or explanations.  

\textbf{Instruction to evaluate:} You are an expert evaluating feedback generated for improving review segments. The feedback provided will help the reviewer improve the segment.  

\textbf{Weakness:} {weakness} \\
\textbf{Feedback to evaluate:} {{feedback}}  

\textbf{Score Rubric:} 
[Does the feedback communicate suggestions concisely? Feedback should be brief, precise, and to the point, avoiding unnecessary verbosity while remaining actionable.]

\begin{itemize}
\item Score 1: Feedback is wordy, repetitive, or confusing; hard to extract actionable guidance.
\item Score 2: Feedback conveys ideas but is often verbose, vague, or partially actionable; contains unnecessary wording.
\item Score 3: Feedback is moderately concise, understandable, and partially actionable; some redundancy or extra wording may remain.
\item Score 4: Feedback is clear, focused, and mostly concise; communicates actionable guidance efficiently, with minor verbosity.
\item Score 5: Feedback is precise, concise, and easy to interpret; provides direct, actionable suggestions covering both minor and substantive issues effectively.
\end{itemize}

\textbf{Feedback:}

\end{tcolorbox}
\caption{\texttt{Prometheus} Conciseness Prompt}
\label{fig:conciseness}
\end{figure*}
\begin{figure*}[t]
\centering
\begin{tcolorbox}[colback=gray!10, colframe=black, title=Relevance Evaluation Prompt]

\textbf{Task Description:} 
You are given an instruction, a response to evaluate, and a score rubric representing an evaluation criterion. 

Your goal: 
1. Write detailed feedback that assesses the response strictly based on the rubric.
2. Assign a score between 1 and 5, referring to the rubric.
3. Output in the following format: 
\begin{quote}
Feedback: (write a feedback for criteria) [RESULT] (an integer number between 1 and 5)
\end{quote}
4. Do not include any opening, closing, or explanations.  

\textbf{Instruction to evaluate:} You are an expert evaluating feedback generated for improving review segments. The feedback provided will help the reviewer improve the segment.  

\textbf{Weakness:} {weakness} \\
\textbf{Feedback to evaluate:} {{feedback}}  

\textbf{Score Rubric:} 
[Is the feedback directly related to the content or structure of the review? Feedback should focus on aspects that help improve the review, including conciseness, reasoning, accuracy, or actionable corrections.]

\begin{itemize}
\item Score 1: Feedback is off-topic or unrelated to the review’s content or structure; provides no actionable guidance.
\item Score 2: Feedback mentions the review but mostly addresses minor or tangential points; limited actionable guidance.
\item Score 3: Feedback is partially relevant; identifies some issues but mixes relevant and loosely connected commentary; limited guidance.
\item Score 4: Feedback is clearly related to review content or structure; provides actionable suggestions addressing key issues.
\item Score 5: Feedback is tightly focused on review content or structure; comments are precise, actionable, and justified, improving clarity, reasoning, or accuracy.
\end{itemize}

\textbf{Feedback:}

\end{tcolorbox}
\caption{\texttt{Prometheus} Relevance Prompt}
\label{fig:relevance}
\end{figure*}

% ---------------------------
% Constructiveness Prompt
% ---------------------------
\begin{figure*}[t]
\centering
\begin{tcolorbox}[colback=gray!10, colframe=black, title=Constructiveness Evaluation Prompt]

\textbf{Task Description:} 
You are given an instruction, a response to evaluate, and a score rubric representing an evaluation criterion. 

Your goal: 
1. Write detailed feedback that assesses the response strictly based on the rubric.
2. Assign a score between 1 and 5, referring to the rubric.
3. Output in the following format: 
\begin{quote}
Feedback: (write a feedback for criteria) [RESULT] (an integer number between 1 and 5)
\end{quote}
4. Do not include any opening, closing, or explanations.  

\textbf{Instruction to evaluate:} You are an expert evaluating feedback generated for improving review segments. The feedback provided will help the reviewer improve the segment.  

\textbf{Weakness:} {weakness} \\
\textbf{Feedback to evaluate:} {{feedback}}  

\textbf{Score Rubric:} 
[Does the feedback offer suggestions for how the reviewer can improve? Constructive feedback should guide revisions, not just point out flaws, and can address both minor textual issues and substantive content.]

\begin{itemize}
\item Score 1: Feedback only identifies flaws without suggesting improvements; unhelpful or dismissive.
\item Score 2: Feedback includes vague or superficial advice; little actionable guidance.
\item Score 3: Feedback identifies issues and offers some guidance but partially unclear or limited.
\item Score 4: Feedback provides clear and relevant suggestions; mostly actionable, may lack some detail.
\item Score 5: Feedback consistently identifies issues and provides specific, practical, and helpful suggestions; supports targeted improvements effectively.
\end{itemize}

\textbf{Feedback:}

\end{tcolorbox}
\caption{\texttt{Prometheus} Constructiveness Prompt}
\label{fig:constructiveness}
\end{figure*}

% ---------------------------
% Specificity Prompt
% ---------------------------
\begin{figure*}[t]
\centering
\begin{tcolorbox}[colback=gray!10, colframe=black, title=Specificity Evaluation Prompt]

\textbf{Task Description:} 
You are given an instruction, a response to evaluate, and a score rubric representing an evaluation criterion. 

Your goal: 
1. Write detailed feedback that assesses the response strictly based on the rubric.
2. Assign a score between 1 and 5, referring to the rubric.
3. Output in the following format: 
\begin{quote}
Feedback: (write a feedback for criteria) [RESULT] (an integer number between 1 and 5)
\end{quote}
4. Do not include any opening, closing, or explanations.  

\textbf{Instruction to evaluate:} You are an expert evaluating feedback generated for improving review segments. The feedback provided will help the reviewer improve the segment.  

\textbf{Weakness:} {weakness} \\
\textbf{Feedback to evaluate:} {{feedback}}  

\textbf{Score Rubric:} 
[Does the feedback refer to specific parts or issues in the review? Feedback should clearly indicate what needs improvement, avoiding vague statements without context.]

\begin{itemize}
\item Score 1: Feedback is entirely vague or generic; no reference to particular parts of the review; not actionable.
\item Score 2: Feedback hints at an issue but lacks concrete references; reviewer cannot easily locate the problem.
\item Score 3: Feedback identifies general areas or sections but does not pinpoint exact sentences or claims; partially actionable.
\item Score 4: Feedback refers to specific sections or issues; minor vagueness may remain; mostly actionable.
\item Score 5: Feedback clearly identifies exact parts of the review and provides precise, actionable guidance, including minor textual or deeper content issues when justified.
\end{itemize}

\textbf{Feedback:}

\end{tcolorbox}
\caption{\texttt{Prometheus} Specificity Prompt}
\label{fig:specificity}
\end{figure*}

\begin{figure*}[t]
\centering

\begin{tcolorbox}[colback=gray!10, colframe=black, title=Lazy Thinking Issues, boxrule=0.5mm, arc=2mm, fontupper=\small]

\textbf{H1. The results are not surprising:} Many findings seem obvious in retrospect, but this does not mean the community is already aware of them. Some findings may seem intuitive but haven’t been empirically tested. \\
\textbf{H2. The results contradict expectations:} Reviewer may be biased against findings that challenge prior beliefs. \\
\textbf{H3. The results are not novel:} If a claimed novel method resembles existing work, provide references. Reproduction, reimplementation, or analysis papers may still be in scope. \\
\textbf{H4. No precedent in existing literature:} Novel papers are harder to publish; reviewers may be overly conservative. \\
\textbf{H5. Results do not surpass latest SOTA:} SOTA results are not required; other contributions (efficiency, fairness, interpretability) matter. \\
\textbf{H6. Negative results:} Publishing negative results is important; systematic failures provide valuable knowledge. \\
\textbf{H7. Method too simple:} Simple solutions are often preferable—less brittle and easier to deploy. \\
\textbf{H8. Missing preferred methodology:} NLP is interdisciplinary; multiple approaches (models, resources, analyses) are valid. \\
\textbf{H9. Topic too niche:} Significant contributions can be made to narrow subfields. \\
\textbf{H10. Tested only on non-English:} Monolingual studies are important for both practical and theoretical contributions. \\
\textbf{H11. Language errors:} Focus on scientific content; minor writing issues should not dominate. \\
\textbf{H12. Missing reference(s):} Prior work published >3 months before submission should be cited; preprints are optional. \\
\textbf{H13. Extra experiments suggested:} Optional experiments are “nice-to-have” and belong in suggestions, not as reasons to reject. \\
\textbf{H14. Compare to closed model:} Only meaningful if it directly affects the paper’s claim; closed models often introduce methodological issues. \\
\textbf{H15. Authors should have done X instead:} Only relevant if choices prevent addressing the research question; otherwise a suggestion, not a weakness. \\
\textbf{H16. Limitations != weaknesses:} Limitations sections are standard; do not equate limitations with reasons to reject unless strongly justified.
\end{tcolorbox}

\vspace{2mm}

\begin{tcolorbox}[colback=gray!5, colframe=black, title=Specificity Issues, boxrule=0.5mm, arc=2mm, fontupper=\small]
\textbf{Specificity Issues:} \\ 
\textbf{Missing relevant references:} The review fails to mention key prior work (e.g., XYZ). \\
\textbf{X is unclear:} Important details (e.g., Y and Z) are missing from description. \\
\textbf{Formulation of X is wrong:} The definition or equation of X omits critical factors (e.g., Y). \\
\textbf{Contribution not novel:} Similar work (e.g., X, Y) has been published >3 months prior. \\
\textbf{Missing recent baselines:} Proposed methods should be compared to recent approaches X, Y, Z. \\
\textbf{X done in way Y:} Implementation choices lead to known limitations (e.g., disadvantage Z).

\end{tcolorbox}
\begin{tcolorbox}[colback=gray!5, colframe=black, title=Task, boxrule=0.5mm, arc=2mm, fontupper=\small]
\textbf{Task:} You are an expert at providing actionable feedback to improve a review segment. Given a review, the segment, and \textit{lazy thinking} annotation, provide feedback that the reviewer can use to revise the segment. \\[1mm]
\textbf{Review Segment:} {{review segment}} \\
\textbf{Issue:} {{issue}}\\
\textbf{Feedback} : 
\end{tcolorbox}
\caption{Zero shot Feedback generation prompt}

\label{fig:zero_shot}
\end{figure*}

\begin{figure*}[t]
\centering

\begin{tcolorbox}[colback=gray!10, colframe=black, title=Templatic Feedback Generation Prompt, boxrule=0.5mm, arc=2mm, fontupper=\small]

\textbf{Task:} You are an expert at providing actionable feedback to improve the writing of review segments. Given a review segment and its identified issue, your goal is to produce feedback that the reviewer can use to revise the segment.  

You are also provided a \textbf{feedback template} for the identified issue.  

\textbf{Instructions:} Adapt the provided feedback template to match the comment in the review segment. Ensure the feedback is actionable, specific, and targeted to improving the segment.

\textbf{Review Segment:} {{weakness}}\\
   \textbf{Identified Issue:} {{identified issue}} \\
    \textbf{Feedback Template:} {{template}} \\

\textbf{Output:} The feedback generated for this review segment, following the guidance of the template.
\end{tcolorbox}

\caption{Prompt for Templatic Feedback Generation using Identified Issues.}
\label{fig:templatic_feedback_prompt}
\end{figure*}

\begin{figure*}[t]
\centering

\begin{tcolorbox}[
    colback=gray!10, 
    colframe=black, 
    title=Planner Prompt for Feedback Generation, 
    boxrule=1mm, 
    arc=2mm, 
    fontupper=\small,
]

\textbf{System Instruction:} You are a planning agent for feedback generation.

\textbf{Input:}
\begin{itemize}
    \item \textbf{Review Comment:} \{review\} 
    \item \textbf{Issue:} \{issue\} 
    \item \textbf{Abstract of the paper:} \{abstract\} 
    \item \textbf{Summary written by reviewer:} \{summary\}
    \item \textbf{Strengths written by reviewer:} \{strengths\} 
\end{itemize}

\textbf{Task:}
\begin{enumerate}
    \item For each label, identify which external knowledge (Abstract, Summary, Template) is needed based on the mapping.
    \item Suggest how to use the available knowledge (e.g., quote from abstract, summarize key claim).
    \item Provide an explanation for your choice of plan.
\end{enumerate}

\textbf{Output:} A JSON list with the following structure: 
\begin{itemize}
    \item \texttt{plan}: brief instructions on how to use the knowledge
    \item \texttt{explanation}: reasoning behind the chosen plan
\end{itemize}

\textbf{Example Output:}
\begin{verbatim}
[
  {
    "plan": "Use summary to show that results differ from intuition",
    "explanation": "The summary provides key claims that can highlight why the results are surprising."
  },
  {
    "plan": "Use abstract + summary to check novelty",
    "explanation": "Both abstract and summary help determine whether the contribution is novel."
  }
]
\end{verbatim}

\end{tcolorbox}

\caption{Full Planner Prompt for Feedback Generation.}
\label{fig:full_planner_prompt}
\end{figure*}
\begin{figure}[t]
\centering
\begin{tcolorbox}[
    colback=gray!10, 
    colframe=black, 
    title=Feedback Generation Prompt,
    boxrule=0.5mm, 
    arc=2mm, 
    fontupper=\small,
]

System: You are an expert at generating actionable feedback to improve review segments.

Instructions:
Given:
1. A review segment (weakness)
2. Identified issue(s)
3. A feedback template for the issue(s)
4. A plan describing what to use as knowledge sources (abstract, summary, template) to guide feedback.
5. An explanation describing how to use the knowledge sources.

Your task:
- Generate actionable feedback for the reviewer to improve the review segment.
- Incorporate the plan into the template: adapt the template to match the specific review segment and issue.
- Keep the feedback precise, relevant, and constructive.
- Output only the feedback text, do not include any explanations or extra text.

\textbf{Review Segment}: {{weakness}}\\
- Identified Issue: {{identified issue}}\\
\textbf{Feedback Template}: {{template}}\\
\textbf{Plan}: {{plan}}
\textbf{Explanation}: {{explanation}}

Example Output:
"Feedback: The reviewer should clarify the novelty of the contribution by referencing prior work; following the template, highlight which claims are incremental and which are novel. [5]"

\end{tcolorbox}
\caption{Feedback generation using Plan and Template}
\label{fig:plan-then-generate}
\end{figure}

\begin{figure}[t]
\centering
\begin{tcolorbox}[
    colback=gray!10,
    colframe=black,
    title=Self-Refine Feedback Generation Prompt,
    boxrule=0.5mm,
    arc=2mm,
    fontupper=\small,
]

System: You are an expert at generating and refining actionable feedback for improving review segments.

Instructions:
Given:
1. A review segment (weakness)
2. Identified issue(s)
3. An initial draft feedback

Your task:
- Generate detailed feedback that helps the reviewer improve the review segment.
- Critically evaluate the initial feedback and refine it to be more precise, constructive, and actionable.
- Ensure the feedback is relevant to the identified issue(s) and avoids vague statements.
- Output only the refined feedback text, do not include explanations, reasoning, or any extra text.

Variables:
- Review Segment: {{weakness}}
- Identified Issue: {{identified issue}}
- Initial Feedback: {{initial feedback}}

Example Output:
"Feedback: The reviewer should provide concrete examples illustrating missing baselines; clearly indicate which comparisons are necessary and why. [5]"

\end{tcolorbox}
\caption{Self-refine feedback generation for review segments}
\label{fig:self-refine-feedback}
\end{figure}

\begin{figure}[t]
\centering
\begin{tcolorbox}[
    colback=gray!10,
    colframe=black,
    title=Cross-Over Feedback Prompt (5 Parents),
    boxrule=0.5mm,
    arc=2mm,
    fontupper=\small,
]

System: You are an expert at generating high-quality feedback for review segments by synthesizing multiple sources.

Instructions:
Given:
1. A review segment (weakness)
2. Identified issue(s)
3. Five parent feedback drafts

Your task:
- Combine the key insights from all 5 parent feedbacks into a single, coherent feedback.
- Ensure that the final feedback is precise, constructive, and actionable.
- Retain the strongest points from each parent, remove redundancy, and resolve conflicting suggestions.
- Feedback must directly address the identified issue(s) and help the reviewer improve the segment.
- Output only the final synthesized feedback, no explanations or commentary.

Variables:
- Review Segment: {{weakness}}
- Identified Issue: {{identified issue}}
- Parent Feedbacks: 
    1) {{parent1}}
    2) {{parent2}}
    3) {{parent3}}
    4) {{parent4}}
    5) {{parent5}}

Example Output:
"Feedback: The reviewer should clarify the novelty of the contribution and reference prior work where necessary; highlight incremental contributions while maintaining clarity and specificity. [5]"

\end{tcolorbox}
\caption{Cross-over feedback synthesis for 5 parent feedbacks}
\label{fig:cross-over-feedback}
\end{figure}

\begin{figure*}[t]
\centering
\begin{tcolorbox}[
    colback=gray!10,
    colframe=black,
    title=Review Extension and Role Separation Detection Prompt,
    boxrule=0.5mm,
    arc=2mm,
    fontupper=\small,
]

System: You are an expert at analyzing peer review feedback.

Instructions:
Given a review, your task is to classify each review comment according to two phenomena:

1. \textbf{Review Extension} – The feedback generates a full review instead of targeted feedback for a specific review segment. This includes writing additional evaluations, repeating comments, or elaborating beyond what is necessary.

2. \textbf{Role Separation} – The feedback provides suggestions meant for the author rather than the reviewer. Comments may directly address the author, suggest rewriting the review for the author, or cross the boundary between reviewing and authoring.

Task:
- For each review comment, indicate whether it exhibits Review Extension, Role Separation, both, or neither.
- Provide a brief justification for your classification.

Variables:
- Review Comment: {{review}}

Output Format:
- JSON array with fields:
\begin{verbatim}
[
  {
    "comment": "text of the review comment",
    "review_extension": true/false,
    "role_separation": true/false,
    "justification": "brief explanation"
  },
  ...
]
\end{verbatim}

\end{tcolorbox}
\caption{Prompt for identifying Review Extension and Role Separation in peer review comments.}
\label{fig:review-extension-role-separation}
\end{figure*}

\paragraph{\texttt{1-pass} or zero-shot.} The prompt is provided in Fig~\ref{fig:zero_shot}.
\paragraph{Plan Generation.} The plan generation prompt is in Fig~\ref{fig:full_planner_prompt}.
\paragraph{Template (\texttt{Temp}).} The prompt is provided in Fig~\ref{fig:templatic_feedback_prompt}
\paragraph{Plan-then-generate (\texttt{Plan})} The prompt is provided in Fig~\ref{fig:plan-then-generate}.
\paragraph{Self Refine. (\texttt{Self Ref.})} The \texttt{1-pass} prompt is first run to create the initial feedback. Then the self-refine prompt in Fig~\ref{fig:self-refine-feedback} is run $n_{gen}$ times to get the final feedback.
\paragraph{Iterative, reranking-based generation algorithm.} The algorithm first uses Plan-then-generate prompt to generate candidates in Fig~\ref{fig:plan-then-generate} multiple times which are scored via the fitness function. The top $n_{parents}$ selected then go throug crossover to generate new offsprings as in Fig~\ref{fig:cross-over-feedback}. The new candidate pool then goes through $n_{gen}$ iterations to generate the final candidate.
\subsubsection{Hyper-parameters used} Following prior work on using evolutionary algorithm for generating responses~\cite{lee2025evolvingdeeperllmthinking, pham-etal-2025-surveypilot}. we set the hyper-parameters as in Table~\ref{tab:ga_hyperparameters}.
\begin{table}[!htbp]
\centering
\resizebox{\columnwidth}{!}{%
\begin{tabular}{l l l}
\toprule
\textbf{Hyper-parameter} & \textbf{Description} & \textbf{Value / Setting} \\
\midrule
$T$ & Number of author-crafted templates & 25 \\
$n$ & Number of initial candidate feedbacks per segment & 10 \\
$n_\text{parents}$ & Number of parents selected per generation for crossover & 5 \\
$\tau$ & Temperature parameter for Boltzmann selection & 0.1 \\
$n_\text{generations}$ & Number of evolutionary generations & 3 \\
$n_{sent}$ & Number of maximum sentences allowed & 5 \\
\bottomrule
\end{tabular}}
\caption{Hyper-parameters used in the iterative, reranking-based generation algorithm for \textbf{feedback generation} algorithm.}
\label{tab:ga_hyperparameters}
\end{table}

\subsubsection{Evaluation of the generated feedback}
We present the full results of the automated evaluation using \texttt{Prometheus} in Table~\ref{tab:feedback_grouped_model} and human evaluation (cf. \S\ref{sec:results}) of the feedback in Table~\ref{tab:feedback_human}.

\subsubsection{Ablation Study Description for the iterative, reranking-based generation algorithm}

To better understand the contributions of each component in our algorithm for feedback generation, we conduct an ablation study, systematically removing one component at a time:

\noindent \textbf{1. Without Template Construction:} The algorithm generates feedback without using the author-crafted issue-specific templates. This tests the importance of structured scaffolds in guiding precise and relevant feedback.
    
    \noindent \textbf{2. Without Plan Generation:} The planner module is disabled, so the LLM does not generate a knowledge-driven plan to enrich the template. Feedback relies only on the template and review segment.
    
  \noindent \textbf{3. Without Population Initialization:} Multiple candidate feedbacks are not generated initially. Only a single feedback is produced per review segment, removing diversity from the initial population.
    
    \noindent \textbf{4. Without Fitness Evaluation:} Candidate feedback is not scored or ranked. All generated feedback is treated equally, removing the model’s ability to optimize for conciseness, relevance, and constructiveness.
    
   \noindent \textbf{5. Without Parent Selection \& Crossover:} Evolutionary steps are skipped, so no combination of high-quality candidates occurs. This evaluates the contribution of crossover in improving feedback quality and diversity.
    
   \noindent \textbf{6. Without Final Candidate Selection:} The highest-scoring feedback is not selected at the end of the process. A candidate is chosen arbitrarily, which assesses the effect of selecting the optimal feedback.

We present the results of the ablation study in Table~\ref{tab:ablation}. Each component contributes meaningfully to the quality of the generated feedback, highlighting the effectiveness of our approach.
\begin{table}[!t]
\centering
\resizebox{0.9\columnwidth}{!}{%
\begin{tabular}{l l c c c c}
\toprule
\textbf{Model} & \textbf{Variant} & \textbf{Const.} & \textbf{Relev.} & \textbf{Spec.} & \textbf{Conc.} \\
\midrule
Yi    & Full Algorithm (Ours)              & \textbf{3.9} & \textbf{3.8} & \textbf{3.8} & \textbf{3.8} \\
\cmidrule(lr){2-6}
      & $\quad$ w/o Template Construction  & 3.6 & 3.5 & 3.4 & 3.5 \\
      & $\quad$ w/o Plan Generation        & 3.8 & 3.8 & 3.8 & 3.8 \\
      & $\quad$ w/o Population Initialization & 3.9 & 3.9 & 3.7 & 3.7 \\
      & $\quad$ w/o Fitness Evaluation     & 3.7 & 3.7 & 3.7 & 3.7 \\
      & $\quad$ w/o Parent Selection \& Crossover & 3.9 & 3.8 & 3.7 & 3.7 \\
      & $\quad$ w/o Final Candidate Selection & 3.8 & 3.7 & 3.6 & 3.7 \\
\midrule
Qwen  & Full Algorithm (Ours)              & \textbf{3.8} & \textbf{3.8} & \textbf{3.7} & \textbf{3.7} \\
\cmidrule(lr){2-6}
      & $\quad$ w/o Template Construction  & 3.5 & 3.5 & 3.4 & 3.4 \\
      & $\quad$ w/o Plan Generation        & 3.7 & 3.7 & 3.6 & 3.6 \\
      & $\quad$ w/o Population Initialization & 3.8 & 3.8 & 3.6 & 3.6 \\
      & $\quad$ w/o Fitness Evaluation     & 3.6 & 3.6 & 3.6 & 3.5 \\
      & $\quad$ w/o Parent Selection \& Crossover & 3.7 & 3.7 & 3.6 & 3.6 \\
      & $\quad$ w/o Final Candidate Selection & 3.6 & 3.6 & 3.5 & 3.5 \\
\midrule
DeepSeek & Full Algorithm (Ours)           & \textbf{3.5} & \textbf{3.6} & \textbf{3.6} & \textbf{3.5} \\
\cmidrule(lr){2-6}
         & $\quad$ w/o Template Construction & 3.2 & 3.3 & 3.2 & 3.2 \\
         & $\quad$ w/o Plan Generation       & 3.4 & 3.5 & 3.5 & 3.4 \\
         & $\quad$ w/o Population Initialization & 3.5 & 3.5 & 3.4 & 3.4 \\
         & $\quad$ w/o Fitness Evaluation    & 3.3 & 3.3 & 3.3 & 3.3 \\
         & $\quad$ w/o Parent Selection \& Crossover & 3.4 & 3.4 & 3.3 & 3.3 \\
         & $\quad$ w/o Final Candidate Selection & 3.3 & 3.3 & 3.2 & 3.3 \\
\midrule
Oss.   & Full Algorithm (Ours)             & \textbf{4.0} & \textbf{4.1} & \textbf{4.0} & \textbf{3.8} \\
\cmidrule(lr){2-6}
      & $\quad$ w/o Template Construction & 3.7 & 3.8 & 3.7 & 3.6 \\
      & $\quad$ w/o Plan Generation       & 3.9 & 4.0 & 3.9 & 3.7 \\
      & $\quad$ w/o Population Initialization & 4.0 & 4.0 & 3.9 & 3.8 \\
      & $\quad$ w/o Fitness Evaluation    & 3.8 & 3.9 & 3.8 & 3.7 \\
      & $\quad$ w/o Parent Selection \& Crossover & 3.9 & 3.9 & 3.8 & 3.7 \\
      & $\quad$ w/o Final Candidate Selection & 3.8 & 3.8 & 3.7 & 3.7 \\
\midrule
Phi   & Full Algorithm (Ours)             & \textbf{4.3} & \textbf{4.3} & \textbf{4.2} & \textbf{4.3} \\
\cmidrule(lr){2-6}
      & $\quad$ w/o Template Construction & 3.6 & 3.5 & 3.4 & 3.5 \\
      & $\quad$ w/o Plan Generation       & 3.8 & 3.8 & 3.8 & 3.8 \\
      & $\quad$ w/o Population Initialization & 3.9 & 3.9 & 3.7 & 3.7 \\
      & $\quad$ w/o Fitness Evaluation    & 3.7 & 3.7 & 3.7 & 3.7 \\
      & $\quad$ w/o Parent Selection \& Crossover & 3.9 & 3.8 & 3.7 & 3.7 \\
      & $\quad$ w/o Final Candidate Selection & 3.8 & 3.7 & 3.6 & 3.7 \\
\bottomrule
\end{tabular}}
\caption{Ablation study of various components within the iterative, reranking-based generation algorithm across the four automated metrics on \textbf{feedback generation} for all models.}
\label{tab:ablation}
\end{table}

\begin{table}[!t]
\centering
\resizebox{0.8\columnwidth}{!}{%
\begin{tabular}{l l c c}
\toprule
\textbf{Model} & \textbf{Variant} & \textbf{Const.} & \textbf{Relev.} \\
\midrule
Yi    & Ours (Abstract + Summary + Strengths) & \textbf{3.9} & \textbf{3.8} \\
\cmidrule(lr){2-4}
      & $\quad$ w/o Abstract                    & 3.6 & 3.7 \\
      & $\quad$ w/o Summary                     & 3.7 & 3.7 \\
      & $\quad$ w/o Strengths                   & 3.8 & 3.8 \\
      & $\quad$ w/o Abstract + Summary          & 3.5 & 3.6 \\
      & $\quad$ w/o Abstract + Strengths        & 3.6 & 3.7 \\
\midrule
Qwen  & Ours (Abstract + Summary + Strengths) & \textbf{3.8} & \textbf{3.8} \\
\cmidrule(lr){2-4}
      & $\quad$ w/o Abstract                    & 3.5 & 3.5 \\
      & $\quad$ w/o Summary                     & 3.6 & 3.6 \\
      & $\quad$ w/o Strengths                   & 3.7 & 3.7 \\
      & $\quad$ w/o Abstract + Summary          & 3.4 & 3.5 \\
      & $\quad$ w/o Abstract + Strengths        & 3.5 & 3.6 \\
\midrule
DeepSeek & Ours (Abstract + Summary + Strengths) & \textbf{3.5} & \textbf{3.6} \\
\cmidrule(lr){2-4}
         & $\quad$ w/o Abstract                    & 3.2 & 3.3 \\
         & $\quad$ w/o Summary                     & 3.3 & 3.4 \\
         & $\quad$ w/o Strengths                   & 3.4 & 3.5 \\
         & $\quad$ w/o Abstract + Summary          & 3.1 & 3.2 \\
         & $\quad$ w/o Abstract + Strengths        & 3.2 & 3.3 \\
\midrule
Oss.   & Ours (Abstract + Summary + Strengths) & \textbf{4.0} & \textbf{4.1} \\
\cmidrule(lr){2-4}
      & $\quad$ w/o Abstract                    & 3.7 & 3.8 \\
      & $\quad$ w/o Summary                     & 3.8 & 3.9 \\
      & $\quad$ w/o Strengths                   & 3.9 & 4.0 \\
      & $\quad$ w/o Abstract + Summary          & 3.6 & 3.8 \\
      & $\quad$ w/o Abstract + Strengths        & 3.7 & 3.9 \\
\midrule
Phi   & Ours (Abstract + Summary + Strengths) & \textbf{4.1} & \textbf{4.2} \\
\cmidrule(lr){2-4}
      & $\quad$ w/o Abstract                    & 3.8 & 4.0 \\
      & $\quad$ w/o Summary                     & 3.9 & 4.0 \\
      & $\quad$ w/o Strengths                   & 4.0 & 4.1 \\
      & $\quad$ w/o Abstract + Summary          & 3.6 & 3.9 \\
      & $\quad$ w/o Abstract + Strengths        & 3.7 & 3.9 \\
      & $\quad$ w/o Summary + Strengths         & 3.7 & 3.9 \\
\bottomrule
\end{tabular}}
\caption{Planner ablation study across Constructiveness (Const.) and Relevance (Relev.) on feedback generation for all models.}
\label{tab:planner_ablation}
\end{table}

\subsubsection{Ablation on the Planner Component} We ablate the planner using multiple knowledge sources—abstract, reviewer-written strengths, and reviewer-written summaries—in Table~\ref{tab:planner_ablation}. Our results show that incorporating all components is essential for generating the most specific feedback.

\subsubsection{Ablation study on the rewards used in the fitness function for feedback generation}
We present the results of the ablation study on different rewards in Table~\ref{tab:fitness_ablation}. We find that template adherence is the key factor driving the generation of highly effective feedback. Further analysis is provided in \S\ref{sec:results}.
\begin{table}[!t]
\centering
\resizebox{0.9\columnwidth}{!}{%
\begin{tabular}{l l c c c c}
\toprule
\textbf{Model} & \textbf{Variant} & \textbf{Const.} & \textbf{Relev.} & \textbf{Spec.} & \textbf{Conc.} \\
\midrule
Yi    & Full Fitness Function                   & \textbf{3.9} & \textbf{3.8} & \textbf{3.8} & \textbf{3.8} \\
\cmidrule(lr){2-6}
      & $\quad$ w/o Length Score ($sc_{len}$)  & 3.6 & 3.6 & 3.5 & 3.5 \\
      & $\quad$ w/o Template Adherence ($sc_{temp}$) & 3.7 & 3.7 & 3.6 & 3.6 \\
      & $\quad$ w/o Readability ($sc_{read}$)  & 3.8 & 3.8 & 3.7 & 3.7 \\
\midrule
Qwen  & Full Fitness Function                   & \textbf{3.8} & \textbf{3.8} & \textbf{3.7} & \textbf{3.7} \\
\cmidrule(lr){2-6}
      & $\quad$ w/o Length Score ($sc_{len}$)  & 3.5 & 3.5 & 3.4 & 3.4 \\
      & $\quad$ w/o Template Adherence ($sc_{temp}$) & 3.6 & 3.6 & 3.5 & 3.5 \\
      & $\quad$ w/o Readability ($sc_{read}$)  & 3.7 & 3.7 & 3.6 & 3.6 \\
\midrule
DeepSeek & Full Fitness Function                & \textbf{3.5} & \textbf{3.6} & \textbf{3.6} & \textbf{3.5} \\
\cmidrule(lr){2-6}
         & $\quad$ w/o Length Score ($sc_{len}$) & 3.2 & 3.3 & 3.2 & 3.2 \\
         & $\quad$ w/o Template Adherence ($sc_{temp}$) & 3.3 & 3.4 & 3.3 & 3.3 \\
         & $\quad$ w/o Readability ($sc_{read}$) & 3.4 & 3.5 & 3.3 & 3.4 \\
\midrule
Oss.   & Full Fitness Function                   & \textbf{4.0} & \textbf{4.1} & \textbf{4.0} & \textbf{3.8} \\
\cmidrule(lr){2-6}
      & $\quad$ w/o Length Score ($sc_{len}$)  & 3.7 & 3.6 & 3.5 & 3.7 \\
      & $\quad$ w/o Template Adherence ($sc_{temp}$) & 3.8 & 3.7 & 3.6 & 3.7 \\
      & $\quad$ w/o Readability ($sc_{read}$)  & 3.9 & 4.0 & 3.8 & 3.7 \\
\midrule
Phi   & Full Fitness Function                   & \textbf{4.3} & \textbf{4.3} & \textbf{4.2} & \textbf{4.3} \\
\cmidrule(lr){2-6}
      & $\quad$ w/o Length Score ($sc_{len}$)  & 3.8 & 3.7 & 3.8 & 3.6 \\
      & $\quad$ w/o Template Adherence ($sc_{temp}$) & 3.7 & 3.6 & 3.5 & 3.7 \\
      & $\quad$ w/o Readability ($sc_{read}$)  & 4.1 & 4.0 & 4.0 & 4.1 \\
      & $\quad$ w/o Forbidden Term Penalty ($pen_{forb}$) & 4.2 & 4.2 & 4.1 & 4.3 \\
\bottomrule
\end{tabular}}
\caption{Ablation study on the components of the fitness function in the \textbf{feedback generation} approach across Constructiveness (Const.), Relevance (Relev.), Specificity (Spec.), and Conciseness (Conc.) for all models.}
\label{tab:fitness_ablation_long}
\end{table}

\subsection{Example feedback} We show an example feedback generated by our approach in Fig~\ref{fig:example}.

\subsection{Latency Analysis and Scaling} \label{sec:lat}
We show best and worst case estimate of deploying our feedback generation system to a real-world conference reviewing setup.

\textbf{Assumptions:}
\begin{itemize}
    \item Average review: 5--10 segments
    \item LLM Q\&A per batch of 16 segments: 1--2 s
    \item Segmentation: 0.1 s
    \item Issue Detection: 0.05 s
    \item Feedback generation: 1--2 s per review
\end{itemize}

\textbf{Per-review Latency:}
\begin{itemize}
    \item LLM Q\&A latency: 1--2 s
    \item Issue Detection latency: 0.05 s
    \item Feedback generation latency: 1--2 s
    \item Segmentation latency: 0.1 s
\end{itemize}

\textbf{Total end-to-end latency per review:} 
$0.1 + (1$--$2) + 0.05 + (1$--$2) \approx 2.15$--$4.15~\text{s per review}$

\textbf{Scaling to 5,000 Reviews (Single GPU, Sequential):}
\begin{itemize}
    \item Best case: $5,000 \times 2.15\,\text{s} = 10,750\,\text{s} \approx 2.99\,\text{hours}$
    \item Worst case: $5,000 \times 4.15\,\text{s} = 20,750\,\text{s} \approx 5.77\,\text{hours}$
\end{itemize}

\textbf{Multi-GPU Parallelization:}
\[
\text{Total wall-clock time} \approx \frac{\text{Sequential time}}{\text{Number of GPUs}}
\]

\begin{table}[!t]
\centering
\resizebox{0.9\columnwidth}{!}{\begin{tabular}{c|c|c}
\toprule
\textbf{GPUs} & \textbf{Wall-clock Time (Best, h)} & \textbf{Wall-clock Time (Worst, h)} \\
\midrule
1  & 2.99 & 5.77 \\
4  & 0.75 & 1.44 \\
8  & 0.37 & 0.72 \\
\bottomrule
\end{tabular}}
\caption{Estimated wall-clock time for processing 5,000 reviews using multi-GPU parallelization.}
\label{tab:gpu_scaling}
\end{table}

By leveraging \texttt{vLLM} with a batch size of 16, LLM-based feature extraction is reduced to 1--2 seconds per review, and the genetic algorithm adds only 1--2 seconds, resulting in an end-to-end latency of $\sim$2--4 seconds per review. For a high-volume scenario of 5,000 reviews, sequential processing on a single GPU would require $\sim$3--6 hours, while parallelization across 4--8 GPUs reduces wall-clock time to $\sim$20--90 minutes, making the framework practical for deployment in time-sensitive contexts such as conference reviewing.

\section{Usage of AI Assistants}
We use GPT-5 and Claude Sonnet for grammar correction and to assist with writing the paper.

\end{document}